\PassOptionsToPackage{table}{xcolor}
\documentclass[manuscript,screen]{acmart}

\usepackage{lineno,hyperref}
\modulolinenumbers[5]

\usepackage{titlesec}
\usepackage{tikz}
\usepackage{placeins}
\usetikzlibrary{mindmap}
\usepackage{ragged2e}
\usepackage{ragged2e}
\usepackage{rotating}
\usepackage{tabularx}
\usepackage{adjustbox}
\usepackage{graphicx}
\usepackage{subcaption}
\usepackage{multirow}
\usepackage{colortbl}
\usepackage{enumitem}
\setlist[enumerate]{itemsep=1pt,parsep=0pt,topsep=0pt,partopsep=0pt}
\usepackage{gensymb}
\usepackage{svg}
\usepackage{makecell}
\usepackage{soul}

\makeatletter
\global\let\oriCT@@do@color\CT@@do@color 
\usepackage{array,booktabs,arydshln}
	
\definecolor{LightCyan}{rgb}{0.88,1,1}

\titleformat{\paragraph}
{\normalfont\normalsize\bfseries}{\theparagraph}{1em}{}
\titlespacing*{\paragraph}
{0pt}{3.25ex plus 1ex minus .2ex}{1.5ex plus .2ex}

\AtBeginDocument{%
  \providecommand\BibTeX{{%
    \normalfont B\kern-0.5em{\scshape i\kern-0.25em b}\kern-0.8em\TeX}}}

\setcopyright{acmcopyright}
\copyrightyear{2023}
\acmYear{2023}
\acmDOI{XXXXXXX.XXXXXXX}

\usepackage{natbib}
\begin{document}

\title{Early Detection of Bark Beetle Attack Using Remote Sensing and Machine Learning: A Review}

\author{Seyed Mojtaba Marvasti-Zadeh}
\email{mojtaba.marvasti@ualberta.ca}
\orcid{0000-0003-0536-0796}
\affiliation{%
  \institution{Department of Renewable Resources, University of Alberta}
  \city{Edmonton}
  \state{Alberta}
  \country{Canada}
}

\author{Devin Goodsman}
\affiliation{%
  \institution{Canadian Forest Service, Natural Resources Canada}
  \city{Edmonton}
  \country{Canada}}
\email{devin.goodsman@nrcan-rncan.gc.ca}  

\author{Nilanjan Ray}
\affiliation{%
 \institution{Department of Computing Science, University of Alberta}
  \city{Edmonton}
  \country{Canada}}
\email{nray1@ualberta.ca}

\author{Nadir Erbilgin}
\affiliation{%
  \institution{Department of Renewable Resources, University of Alberta}
  \city{Edmonton}
  \country{Canada}}
\email{erbilgin@ualberta.ca}

\renewcommand{\shortauthors}{S.M. Marvasti-Zadeh, et al.}

\begin{abstract}
  Bark beetle outbreaks can have serious consequences on forest ecosystem processes, biodiversity, forest structure and function, and economies. Thus, accurate and timely detection of bark beetle infestations  {in the early stage} is crucial to mitigate the further impact, develop proactive forest management activities, and minimize economic losses. Incorporating \textit{remote sensing} (RS) data with \textit{machine learning} (ML) (or \textit{deep learning} (DL)) can provide a great alternative to the current approaches that primarily rely on aerial surveys and field surveys, which can be impractical over vast areas.  {Existing approaches that exploit RS and ML/DL exhibit substantial diversity due to the wide range of factors involved.}
  This paper provides a comprehensive review of past and current advances in the early detection of bark beetle-induced tree mortality from three primary perspectives: bark beetle \& host interactions, RS, and ML/DL.  {In contrast to prior efforts, this review encompasses all RS systems and emphasizes ML/DL methods to investigate their strengths and weaknesses.
  We parse existing literature based on multi- or hyper-spectral analyses and distill their knowledge based on: bark beetle species \& attack phases with a primary emphasis on early stages of attacks, host trees, study regions, RS platforms \& sensors, spectral/spatial/temporal resolutions, spectral signatures, \textit{spectral vegetation indices} (SVIs), ML approaches, learning schemes, task categories, models, algorithms, classes/clusters, features, and DL networks \& architectures.
  Although DL-based methods and the \textit{random forest} (RF) algorithm showed promising results, highlighting their potential to detect subtle changes across visible, thermal, and \textit{short-wave infrared} (SWIR) spectral regions, they still have limited effectiveness and high uncertainties. To inspire novel solutions to these shortcomings, we delve into the principal challenges \& opportunities from different perspectives, enabling a deeper understanding of the current state of research and guiding future research directions.} 

\end{abstract}


\maketitle

\section{Introduction} \label{sec:Intro}
\begin{figure}[!tbp]
\centering
\includegraphics[width=0.85\linewidth]{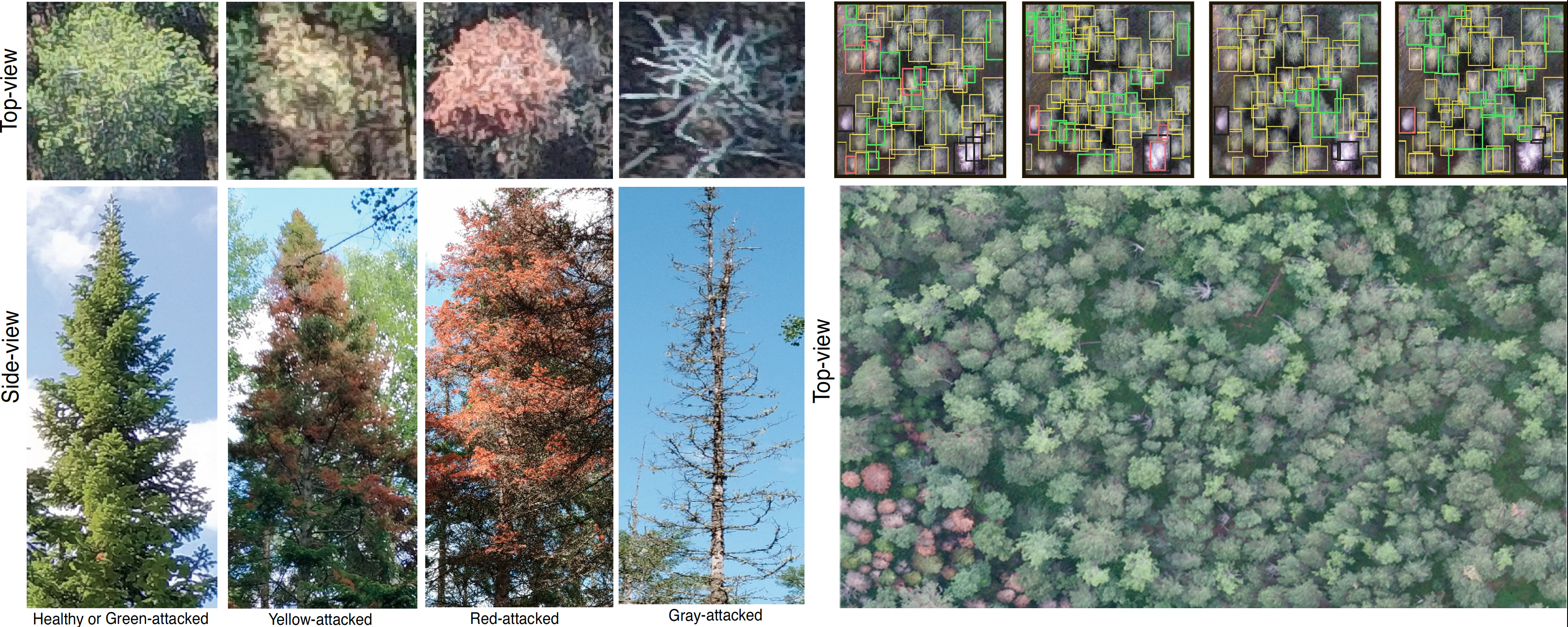}
\vspace{-0.2cm}
\caption{Detection of different stages of bark beetle attack. These stages are shown from images of the top-view (top-left row) and side-view (bottom-left row). Examples of challenging scenarios and promising results are shown in the bottom-right and top-right rows, respectively. Courtesy of \cite{BB_Fir_DL,Detection_UAV_YOLOs} }
\vspace{-0.3cm}
\label{fig:attack_stages} 
\end{figure}

Bark beetle outbreaks are among the most common natural disturbances in the world's coniferous forests, leading to landscape-level tree mortality and potentially having economic and environmental consequences \cite{BeetleOutbreak_US,Living_Bark_Beetles,BeetleImpacts}. Given their large impact at the landscape level, they can even exceed carbon emissions from forest fires \cite{Kurz_etal2008}. Generally, bark beetle-induced mortality in conifers progresses in phases that can be categorized into the \textit{green}, \textit{yellow}, \textit{red}, and \textit{gray attacks} (hereafter denoted as GAtt, YAtt, RAtt, and GRAtt respectively) based on the gradual discoloration of the crown of attacked trees as seen from the air (and to human eyes) and the time since the attack \cite{Book_MPB_Alberta_Gov} (see Fig.~\ref{fig:attack_stages}). {The GAtt is a pre-visual phase that occurs right after successful mass colonization in which bark beetles overcome host tree defenses and start mating and establishing in the host tissues. Colonization eventually leads to tree mortality through the combination of mass attacks and symbiotic pathogenic fungi carried by the beetles. Collectively they disrupt water and nutrient conductance in the xylem and phloem \mbox{\cite{Stimulation_tree_defenses}}. To date, reliable detection of GAtt can only be done from the ground and relies on the presence of RAtt trees (see below). Early detection of GAtt independent of RAtt would substantially increase the implementation of control measures by tackling directly the first generation of beetles instead of the second. However, GAtt detection from above the forest canopy remains a challenge because there are no visually (to human eyes) discernible spectral differences between healthy and GAtt trees}. In the following late spring or early summer, the crown of GAtt trees fades to yellow, becoming YAtt. By the time trees exhibit YAtt symptoms, it is too late to initiate proactive bark beetle management because brood beetles have already left the parental tree. Therefore, the beetle progeny from YAtt trees attack new hosts, leading to more GAtt trees.
The RAtt phase starts when the tree canopy becomes red, and visible signs of leaf color change are evident. The needles will then gradually drop off the host until only the branches without needles remain, which is termed GRAtt due to the greyish color of the dead trees. The duration of each phase is highly variable and depends on the availability of soil moisture, the species of both bark beetles (i.e., those with one vs. multiple generations per year) and host tree species, beetle attack density, and the prior host tree conditions (healthy vs. stressed). This phase can range from a few months for \textit{Ips typographus} on \textit{Picea abies} to over two years for \textit{Dendroctonus ponderosae} on \textit{Pinus contorta} \cite{Book_BarkBeetles_Biology_Ecology_Vega}. \\

\begin{figure}[!tbp]
\centering
\includegraphics[width=0.85\linewidth]{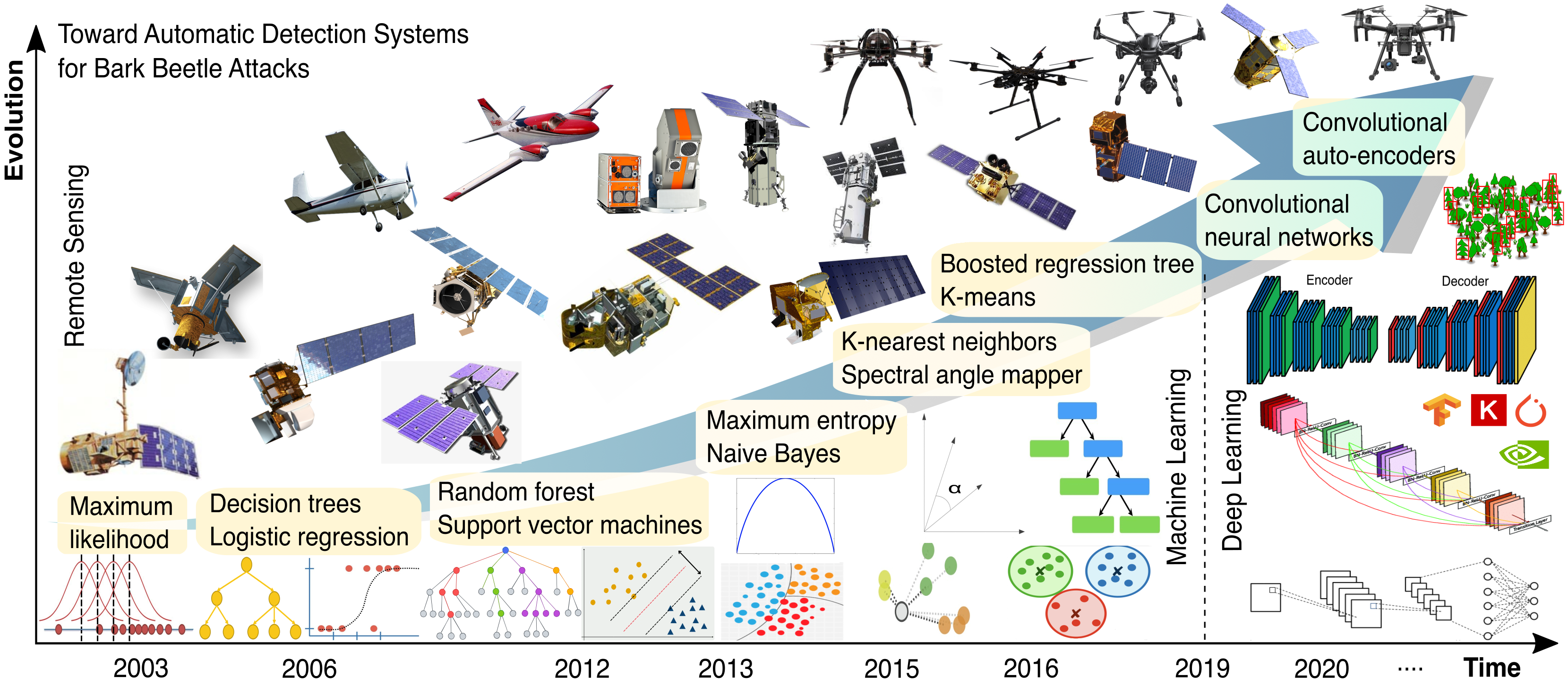}
\vspace{-0.2cm}
\caption{{Evolution of remote sensing and machine learning for bark beetle attack detection.}}
\vspace{-0.3cm}
\label{fig:overview} 
\end{figure}
\indent Monitoring bark beetle infestations typically entails both field and aerial surveys. Aerial surveys are typically conducted with helicopters or fixed-wing aircrafts. Field surveys focus on both monitoring bark beetle populations via traps with beetle lures and locating GAtt trees based on the signs and symptoms of bark beetle attacks such as pitch tubes, sawdust, or frass on the bark around beetle entrance holes. In North America, aerial surveys for bark beetle damage in conifers are generally conducted in the late Summer or early Fall. These aerial surveys can be subdivided into aerial overview surveys, which comprise large areas surveyed from fixed-wing aircraft, and detailed aerial surveys (sometimes called heli-gps surveys), which typically focus on smaller regions of high interest or on challenging terrain surveyed using rotary-wing aircraft \cite{AerialSurvey_manual,AerialSurvey_BC}. In aerial surveys, surveyors in fixed-wing aircraft delineate regions impacted by bark beetles using hand-drawn polygons sketched onto digital tablets loaded with digital topographical maps. Each sketched polygon is then assigned an ordinal severity rating based on the surveyor's assessment of the percentage of susceptible host trees that were recently infested \cite{AerialSurvey_manual,AerialSurvey_BC}. In contrast, detailed aerial surveys capitalize on helicopters that are able to fly slower and lower such that the location of clusters of infested trees can be marked on the tablet using point data--often the number of trees in the cluster will be recorded. In both cases (overview and detailed), aerial surveyors are limited to quantifying only the abundance of recently killed trees with red crowns (RAtt), and it can be difficult to distinguish recently killed trees with red crowns from those that were killed more than one year ago that has retained red foliage \cite{Devin_MPB_AerialSurvey}. Detailed aerial surveys are typically preferred in areas where beetles in infested trees will be controlled (removed) because they provide the necessary information for more efficient control efforts. 
In areas of high management interest, control efforts involve ground crews searching radially around aerially identified RAtt trees to locate GAtt trees using the presence of pitch tubes or bark beetle entrance holes on the tree stem \cite{Book_MPB_Alberta_Gov} as identifying characteristics because the green crowns of GAtt are often not distinguishable from healthy crowns.  {Trees identified as GAtt are usually removed prior to the emergence of bark beetle broods in the late spring or early summer}. 
Another form of control is the proactive harvesting of heavily infested stands \cite{Book_MPB_Alberta_Gov}. This approach requires early detection so that harvest roads can be built and harvesting can be done in the winter. 
 {Regardless of whether individual GAtt trees are located and removed or large stands of GAtt trees are controlled by sanitary harvesting, bark beetles developing under the tree bark must be eliminated prior to the dispersal of beetles from the infested trees to infest new trees}.
For this reason, felling and burning as well as prophylactic harvesting of infested GAtt trees occurs in the winter months when immature bark beetles are inactive \cite{Book_MPB_Alberta_Gov}. 
Hence, proactive management strategies and sanitation felling are crucial to slow down the spread of outbreaks \cite{Living_Bark_Beetles}. These strategies demand precise and detailed information regarding the location of trees infested with bark beetles. However, field surveys are costly, irregular, time-consuming, and impractical for large areas and inaccessible locations. Hence, an overarching question that motivated this review is: To what extent can recent RS and ML advances complement or replace the traditional way in which current bark beetle outbreaks are detected and monitored? \\
\indent Remote sensing (RS, capturing images or other types of data from the air or space, and analyzing them) of bark beetle outbreaks was, until recently, limited to retrospective studies of past outbreaks and was hardly used operationally \cite{Review_Wulder,Review_Wulder_Challenges,Review_Guillermo}. Spaceborne sensors were either too spatially coarse to detect individual trees or too expensive to utilize when they had enough resolution. Aerial photography has long been used to assess bark beetle damage (see brief historical review in \cite{Review_Guillermo}), but the cost of acquiring the imagery and having it manually interpreted to create useful maps was prohibitive compared to the less accurate but much more cost-effective aerial surveys, which in addition can take place below the cloud ceiling and thus do not require cloud-free conditions. However, new advances in both RS and ML/DL can potentially overcome these barriers for detection issues \cite{RS_ForestEcology_Management}. 
There are currently geospatial intelligence companies (e.g., Maxar, Planet) that offer data and services based on new satellites and \textit{artificial intelligence} (AI) \cite{RS_Advances}. There are constellations of earth observation satellites that can provide daily images of sites of interest at spatial resolutions of half a meter or finer, which could replace aerial imagery. In addition, small civilian/consumer drones provide a low-cost highly flexible platform equipped with miniaturized interchangeable sensors for ecological and environmental monitoring \cite{Review_Guillermo,Book_Conservation_Drones}. The images are analyzed by ML- or DL-based methods that could replace human photo interpreters. Figure~\ref{fig:overview} is an allegory for the evolution of RS platforms and ML methods that have been used to detect bark beetle attacks. Together, these advances could bring a new scenario where near-real-time geospatial information is readily available as a service to agencies and even the public on a variety of topics including forest health. The developers of such a service for the detection of bark beetle damage would require knowledge of bark beetles and their interactions with host trees, as well as RS, ML, and DL.  {This paper aims to provide a brief introduction to key aspects of those three realms (see \mbox{Fig.~\ref{fig:mindmap}}) as well as a comprehensive review of the past and current developments and possible future directions for early detection of tree mortality due to bark beetle attacks (i.e., GAtt detection)}. 

\subsection{Preceding Reviews on Bark Beetle Damage Detection}
\begin{figure}[!t]
\centering
\includegraphics[width=0.73\linewidth]{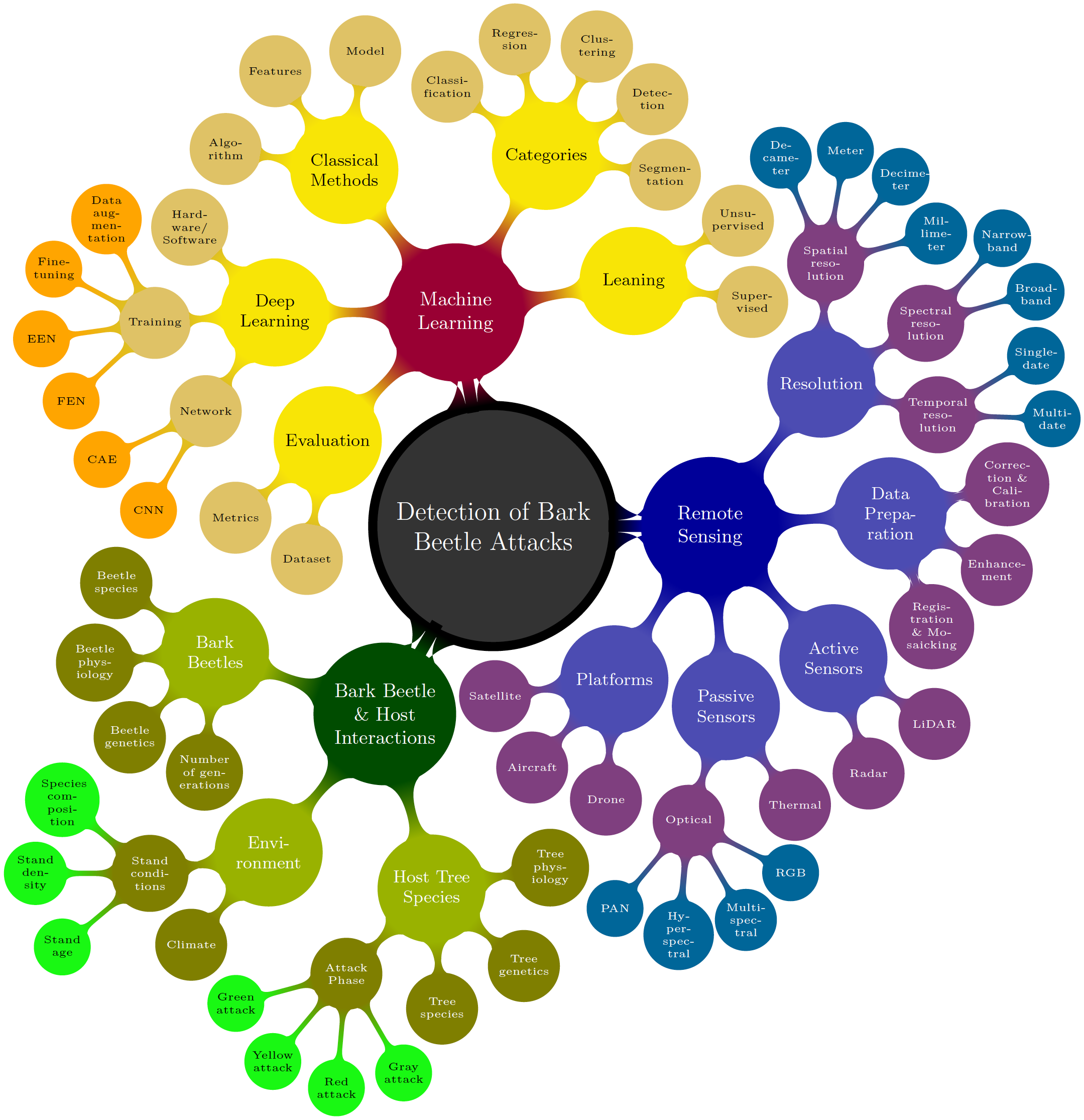}
\vspace{-0.2cm}
\caption{{An overview of three key aspects of: bark beetles \& host interactions, remote sensing, and machine learning (\& deep learning) for early detection of bark beetle attacks. This review comprehensively covers these aspects to serve as a gentle use guide for researchers and shed light on promising research directions.}} 
\vspace{-0.3cm}
\label{fig:mindmap} 
\end{figure}

 {In this section, we summarize previous surveys on the topic \mbox{\cite{Review_Wulder,Review_Wulder_Challenges,Review_Guillermo,Review_InsectDisturbances,Review_Zabihi}}.
In \mbox{\cite{Review_Wulder}}, the primary focus is on the RS efforts aimed at monitoring the impact of the \textit{mountain pine beetle} (MPB) infestation, particularly detecting the RAtt phase in Canada. Following an overview of the biology and forest management aspects, it highlights the key image characteristics and operational information required for monitoring at different forest management scales. This includes regional scale assessments using aerial overview surveys, landscape-scale evaluations using moderate-spatial resolution satellite imagery, and local scale investigations employing high-spatial resolution satellite imagery. This review recommends leveraging RS data as complementary information to address the spatial/temporal limitations of other data collection methods for RAtt detection.
Meanwhile, the biological, logistical, and technological challenges associated with the operational utility of RS data for GAtt detection are explored in \mbox{\cite{Review_Wulder_Challenges}}. This communication recommends consideration of timing factors (e.g., MPB flight period, infestation period, and crown fading rate), image acquisition conditions (e.g., sun angles, snow cover, and area coverage), and technological factors (e.g., complexity and cost of data acquisition) in accordance with forest management intentions. 
Another review in \mbox{\cite{Review_Guillermo}} discusses the application of RS in monitoring eight forest insect pests in Canada (i.e., two bark beetles and six defoliators), explaining the complementary role of RS in conjunction with field \& aerial surveys. It provides a comprehensive discussion of remote sensor characteristics and pre-processing steps for analyzing RS data, with particular attention to the usefulness of RS information in detecting and assessing RAtt damage. It highlights the importance of understanding the pest-host-image triangle for effective damage detection when utilizing either image pairs or single-date imagery.
Another review \mbox{\cite{Review_InsectDisturbances}}, this time from Europe, on the application of RS for three types of insects (bark beetles, broadleaf defoliators, and needle-leaf defoliators) delves into forest insect disturbances. It briefly presents some approaches for detecting RAtt and GRAtt and provides suggestions to overcome related challenges, such as separating insect disturbances from other factors, integrating spatial-temporal analysis, utilizing dense time series data, and improving reference data quality. \\
\indent The use of drones to monitor forest health is covered in two recent reviews. 
In \mbox{\cite{Review_Advances_ForestInsect_UAV}}, the review explores the application of drones in studying ten forest pests (including several bark beetle studies) and nine diseases. However, it mainly examines the type of drone, sensor, data collection, pre-processing, and analytical methods.
The other review \mbox{\cite{Review_UAV_ForestHealth}} takes a broader perspective by discussing both biotic (pests (including bark beetles), diseases, and phytoparasites) and abiotic (e.g., forest fires) stressors. It offers general suggestions including employing \textit{hyperspectral} (HS) sensors, collecting multi-temporal \& long-term data, and leveraging multi-sensor and multi-source approaches.
Finally, the review paper \mbox{\cite{Review_Zabihi}} investigates the impact of spatial, spectral, and temporal resolutions of satellite imagery on the accuracy of GAtt detection. It suggests utilizing satellite imagery with a spatial resolution of less than \mbox{{4} $m$} and explores the potential of individual bands and \textit{spectral vegetation indices} (SVIs) to detect GAtt signatures. Accordingly, it highlights the importance of the \textit{visible} (VIS) and red edge over \textit{near infrared} (NIR) or \textit{shortwave infrared} (SWIR) but neglects to mention \textit{thermal infrared} (TIR). Additionally, it discusses the importance of temporal resolution in relation to flight activity periods and the utilization of time series imagery during the emergence and dispersal period of bark beetle attacks. Despite recommending non-parametric methods (e.g., \textit{Random Forests} (RF)) over parametric ones (e.g., maximum likelihood), this review does not delve into a comprehensive exploration of other RS platforms or ML methods from various aspects.}\\
\indent  {Our review takes a comprehensive and systematic approach to explore distinctive aspects of GAtt detection, as depicted in \mbox{Fig.~\ref{fig:mindmap}}. It acknowledges GAtt detection as an ongoing open problem that served as the primary motivation of our review paper to summarize the progress made thus far, emphasizing the need for a deeper understanding and consideration of all relevant factors involved. 
Unlike previous works, this review encompasses all RS systems and especially ML/DL methods, categorizing them based on their application in \textit{multispectral} (MS) or HS analyses to highlight the strengths and weaknesses of existing methods. It presents a detailed investigation that provides an understanding of these methods based on their distinct capabilities and performance effectiveness to guide informed decision-making when tackling GAtt detection challenges. Toward this goal, we include all relevant studies that provide insights, findings, or experiments related to the GAtt phase utilizing RS and ML/DL techniques to ensure a thorough discussion of the problem. 
Furthermore, overviews of late detection studies (including YAtt, RAtt, and GRAtt phases) are merely listed in tables to summarize the systems and methods employed in other related ongoing research.
To the best of our knowledge, this is the first attempt to provide detailed analyses of methods of bark beetle detection using RS \& particularly ML/DL and to compare the characteristics of existing approaches to serve as a gentle guide to facilitate a discussion on ongoing issues and shed light on promising research directions for practitioners.} 
\subsection{Contribution and Organization}
 {Integrating RS and ML techniques aims to facilitate GAtt detection through timely identification and informed decision-making. However, the systems and techniques exhibit substantial diversity due to the wide range of factors and limitations. For instance, RS tools include multiple platforms and sensors that offer unique capabilities and characteristics, such as different spatial and temporal resolutions, spectral bands, and data collection strategies. Besides, various scenarios and environmental conditions in which data are collected (e.g., weather conditions, seasonal changes, and forest types) contribute to the overall diversity in RS approaches. Similarly, utilized methods can span from classical ML algorithms to \textit{deep neural networks} (DNNs) with their own advantages and limitations. The selection of these methods depends on various factors such as the data complexity, training samples, availability of ground-truth data, and desired accuracy \& generalization. Thus, various analyses and conclusions have been drawn from different investigations on GAtt detection.}

\begin{table*}[!b]
\vspace{-0.3cm}
\rowcolors{2}{}{gray!10}
\caption{List of abbreviations used in this review.} 
\vspace{-0.3cm}
\centering 
 \resizebox{\textwidth}{!}{
\begin{tabular}{c c | c c  c c | c c  c c} 
\hline
\multicolumn{2}{c|}{Bark Beetle \& Host Interactions} & \multicolumn{4}{c|}{Remote Sensing} & \multicolumn{4}{c}{Machine Learning} \\
Abbreviation & Description & Abbreviation & Description & Abbreviation & Description & Abbreviation & Description & Abbreviation & Description   \protect\\ 
\hline\hline
\textbf{ESBB} & European spruce bark beetle & 
\textbf{ARD} & Analysis-ready data                                       & 
\textbf{OLI} & Operational land imagery                       & 
\textbf{ANN} & Artificial neural network                        
&  \textbf{LOGR} & Logistic regression                                     \protect\\
 \textbf{DSBB} &  Dendroctonus spp.          & 
 \textbf{ARI} &  Anthocyanin Reflectance Index                          & 
\textbf{PAN} & Panchromatic                          & 
\textbf{BETA} &  Beta regression                      & 
\textbf{LREG} & Linear regression                         \protect\\
\textbf{FFBB} &  Four-eyed fir bark beetle  & 
\textbf{CIR} &  Color-infrared                              & 
\textbf{PRI} & Normalized difference photochemical reflectance index           & 
\textbf{BLOGR} &  Boosted logistic regression                                & 
\textbf{LSVM} & Linear support vector machine                                   \protect\\
\textbf{GAtt} & Green attack                & 
\textbf{CRI} &  Carotinoid reflectance index                                   & 
\textbf{RDI} & Ratio drought index               & 
\textbf{BRT} & Boosted regression tree                     & 
\textbf{MAXE} & Maximum entropy                                      \protect\\ 
\textbf{GRAtt} & Gray attack                & 
\textbf{ENDVI} & Enhanced normalized difference vegetation index                            & 
\textbf{REIP} & Red-edge inflection point                          & 
\textbf{CAE} & Convolutional auto-encoder                         & 
\textbf{MAXL} & Maximum likelihood          \protect\\   
\textbf{IPSEB} &  Ips spp.                  & 
\textbf{ETM+} &   Enhanced thematic mapper plus                                & 
\textbf{RGI} & Red-green index              & 
\textbf{CART} &  Classification and regression trees                      & 
\textbf{MFDNN} & Multi-layered feedforward deep neural network                                   \protect\\    
\textbf{MEXPB} &  Mexican pine beetle       & 
\textbf{EWDI} &  Enhanced wetness difference index                           & 
\textbf{RS} & Remote sensing                                       &  
\textbf{CLAS} &  Classification            &  
\textbf{ML} & Machine learning                                           \protect\\   
\textbf{MPB} &  Mountain pine beetle       & 
\textbf{GI} & Greenness index                             & 
\textbf{SAR} & Synthetic aperture radar                         & 
\textbf{CLUS} & Clustering                                  & 
\textbf{NBAYES} & Naive Bayes                                 \protect\\  
\textbf{PSB} &  Pine shoot beetle           & 
\textbf{GNDVI} & Green NDVI   & 
\textbf{SR} & Simple ratio                                              &  
\textbf{CNN} &  Convolutional neural networks                                        &  
\textbf{NPAR} & Non-parametric                                                      \protect\\
\textbf{RAtt} & Red attack                  & 
\textbf{GOSAVI} & Green optimized soil adjusted vegetation index                  & 
\textbf{SVI} & Spectral vegetation index          & 
\textbf{DET} & Detection                 & 
\textbf{PAR} & Parametric                         \protect\\    
\textbf{RTB}  &  Red turpentine beetle      & 
\textbf{GRVI} & Green ratio vegetation index                    &  
\textbf{SWIR} & Short-wave infrared                      &
\textbf{DL} & Deep learning                                         &  
\textbf{PCA} & Principal component analysis             \protect\\   
\textbf{SBB} &  Spruce beetle               & 
\textbf{HI} &  HySpex index               &                 
\textbf{TCARI} & Transformed chlorophyll absorption in reflectance index                  & 
\textbf{DT} & Decision tree                                      & 
\textbf{PLS-DA} & Partial least square discriminant analysis                     \protect\\    
\textbf{SFBB} &  Sakhalin fir bark beetle   & 
\textbf{HS} & Hyperspectral                  & 
\textbf{TCC} & True color composites                                 &
\textbf{EA} & Evolutionary algorithm                        & 
\textbf{PSVM} & Polynomial support vector machine               \protect\\     
\textbf{SPB} &  Southern pine beetle        & 
\textbf{IS} & Imaging spectroscopy           &   
\textbf{TCW} & Tasseled cap wetness                       &  
\textbf{EEN} &  End-to-end network                                  &  
\textbf{QDA} & Quadratic discriminant analysis                    \protect\\    
\textbf{WPB} &  Western pine beetle         & 
\textbf{LI} &  Laboratory index         & 
\textbf{TIR} & Thermal infrared     & 
\textbf{EIGEN} &  Clustering-based on Eigenspace transformation                          &  
\textbf{RCNN} & Region-based CNN            \protect\\
\textbf{YAtt} & Yellow attack               & 
\textbf{MSI} & Moisture stress index                              & 
\textbf{TIRS} & Thermal infrared sensor                                        & 
\textbf{ETC} & Extra tree classifier   &  
\textbf{REG} & Regression                              \protect\\    
 &                             & 
 \textbf{LiDAR} &  Light detecting and ranging                                     & 
 \textbf{TM} & Thematic mapper                              &                    
\textbf{FCN} & Fully convolutional network  & 
\textbf{RF} & Random forest                                                  \protect\\     
& & 
\textbf{LWCI} &  Leaf water content index                       & 
 \textbf{UAV} & Unmanned aerial vehicle          &   
\textbf{FEN} &  Feature extraction network                    &   
\textbf{RFG} & Random frog                                          \protect\\  
 &                             & 
\textbf{MS} &  Multispectral                                     & 
\textbf{VHR} & Very high-resolution                                  &  
\textbf{GA} & Genetic algorithm  &  
\textbf{RG} & Region growing                                     \protect\\       
& & 
\textbf{NDRE} &  Normalized difference red-edge index                      & 
\textbf{VNIR} & Visible and near-infrared         &        
\textbf{GLM} & Generalized linear model                          &  
\textbf{SAM} &  Spectral angle mapper                                             \protect\\      
 &                             & 
\textbf{NDRS} & Normalized distance red SWIR                      & 
\textbf{WI} & Water index                                         &        
\textbf{GMM} & Gaussian mixture model                            & 
\textbf{SEG} & Segmentation                                \protect\\     
& & 
\textbf{NDVI} & Normalized difference vegetation index    & 
\textbf{} &                              &   
\textbf{GNBAYES} & Gaussian naive Bayes                         &  
\textbf{SL} & Supervised learning                                           \protect\\    
 &                             & 
\textbf{NDWI} & Normalized difference water index                   & 
 \textbf{} &                                        &    
\textbf{GRADBC} & Gradient boosting classifier                 & 
\textbf{SSEG} & Semantic segmentation                                        \protect\\     
& & 
\textbf{NIR} & Near-infrared                                       & 
\textbf{} &     &    
\textbf{IP} & Image processing                            &  
\textbf{SVM} & Support vector machine                      \protect\\     
 &                             & 
\textbf{NPCI} &  Normalized pigment chlorophyll ratio index &
 \textbf{} &                               &    
\textbf{ISEG} & Instance segmentation                           & 
\textbf{THR} & Thresholding                                          \protect\\    
& & 
\textbf{NSMI} & Normalized difference soil moisture index  &
\textbf{} &                                    &  
\textbf{KNN} & K-nearest neighbors                                  &   
\textbf{USL} &  Unsupervised learning                                         \protect\\     
 &                             & 
\textbf{NWI} & Normalized water index  & 
\textbf{} &                            &   
\textbf{LDA} & Linear discriminative analysis                 & 
\textbf{WSH} & Watershed segmentation                             \protect\\         


\hline 
  \end{tabular}
  \label{tab_Abbrev}  
  }
\end{table*}

 {In this paper, we exclusively focus on a comprehensive review of recent advances and challenges in GAtt detection using RS and ML/DL for practitioners and researchers working in this area. We explore different aspects of the studies based on MS or HS analyses to uncover the unique contributions and insights provided by each approach. The main advantage of MS analysis lies in its simplicity and ease of interpretation because it focuses on a smaller number of spectral bands and enables efficient data acquisition, processing, and evaluation. MS sensors are widely accessible and commonly used, making them practical for many RS applications, but their limited spectral resolution can potentially constrain the ability to differentiate subtle changes to detect specific indicators of GAtt. Alternatively, the HS analysis utilizes a much larger number of contiguous narrow spectral bands to allow a more detailed characterization of subtle changes and spectral signatures. However, the higher spectral dimensionality of this data also presents challenges regarding data storage, processing, and analysis uncertainty \& complexity. On the other hand, the extensive review of ML/DL methods offers a thorough understanding of the existing approaches and their performance in addressing the challenges associated with this problem. More importantly, it helps identify gaps and research opportunities considering the limited success of utilized ML/DL methods for GAtt detection, guiding the development of novel approaches and techniques. 
This paper thus primarily addresses these gaps to present details of recent advances and assess the strengths and limitations of potential approaches for GAtt detection.
We summarize the five main contributions of this paper as follows:}
\begin{enumerate}
    \item  {To set the scene for the problem, we start with a brief introduction to bark beetles and their interactions with host trees. We categorize the studies based on the attack phases examined, bark beetle species involved, tree species studied, geographical locations, and data collection dates. We also emphasize the intricate nature of bark beetles and host interactions, particularly in the presence of diverse environmental conditions and physiological/biological factors.} 

    \item  {Then, we concisely describe the RS approaches and their efficacy for GAtt detection. We categorize these approaches according to the platforms utilized, data collection systems \& analyses, spectral bands \& indices, as well as spatial \& spectral resolutions. In addition, we compare utilized satellite platforms based on their spectral bands and spatial/temporal resolutions to evaluate their suitability for this task.}

    \item  {We then discuss the classical ML algorithms and DL-based methods used for MS/HS analyses of GAtt detection. We categorize these methods based on their approach, learning schemes, defined task, model, algorithm (or network \& architecture), classes/clusters, features, and exploited pixel-/object-based information. Moreover, we highlight the key advantages and weaknesses of the ML/DL methods, then provide a summary of evaluation metrics to clarify their roles in quantifying the performance and effectiveness of these methods.}

    \item  {Next, we quantitatively discuss the effectiveness of ML/DL methods in MS/HS analyses of GAtt detection.  
    Accordingly, we identify promising methods and provide a comprehensive summary of the spectral signatures and indices derived from these analyses, offering helpful insights and potential guidance for practitioners and researchers.} 

    \item  {Finally, we discuss the limitations and requirements of GAtt detection from each perspective and highlight the potential future research directions to address the challenges.}
\end{enumerate}

The rest of this review paper is organized according to the three perspectives of bark beetle \& host interactions (Section~\ref{sec:BarkBeetle}), utilizing the RS for data collection (Section~\ref{sec:RS}), ML/DL methods used for analyses (Section~\ref{sec:ML}). Section~\ref{sec:Evals} discusses the quantitative comparisons of ML/DL methods and conclusions regarding spectral signatures and indices for GAtt detection. Section~\ref{sec:FutureDirections} addresses the open issues, recommendations, and future research directions. Finally, we conclude the paper in Section \ref{sec:Conclusion}. Table \ref{tab_Abbrev} summarizes the abbreviations used in this review. To summarize the categorizations of studies, the key distinctions are provided in the tables, and additional details are presented in the body of the paper. 


\section{ {Bark Beetle-Host Tree Interactions}} \label{sec:BarkBeetle}
Bark beetles (Coleoptera: Curculionidae, Scolytinae) are a well-known group of insects associated with many annual and perennial plant species that occur in all regions of the world \mbox{\cite{Book_BarkBeetles_Biology_Ecology_Vega}} (see Table~{\ref{tab_BB_Tree}} \& Table~{\ref{tab_BB_Env}}). They play key roles in ecosystem function by contributing to nutrient cycling, altering above-ground plant communities, initiating gap dynamics, and altering biodiversity, soil biome, and chemistry. Some bark beetle species have received particular attention due to their ability to kill healthy trees over large areas in many forest ecosystems and hence are considered ecologically and economically important species.
In North America, several species \textit{Dendroctonus} (\textit{Dentro}-tree, \textit{tonus}-destroyer), \textit{D. ponderosae} (MPB), \textit{D. rufipennis} (spruce beetle), \textit{D. brevicomis} (western pine beetle), \textit{D. frontalis} (southern pine beetle) and \textit{D. jeffreyi} (Jeffrey pine beetle) develop outbreaks and are capable of killing a larger number of pine and spruce trees \mbox{\cite{Book_BarkBeetles_Biology_Ecology_Vega}}. 
Likewise, in Eurasia, the European spruce bark beetle (\textit{Ips typographus}) is capable of killing Norway spruce over large areas. 
Most tree-killing bark beetle species have similar basic life history attributes.
 {Bark beetles spend their entire larval stage under tree bark and become pupae in the spring or early summer and emerge as adults in the following years}.
While most bark beetle species have a one-year life cycle, a smaller subset of bark beetle species, particularly those with diapause (a period of suspended development during unfavorable environmental conditions) may need more than one year to complete their development. 

\begin{figure}[!t]
\centering
\includegraphics[width=0.75\linewidth]{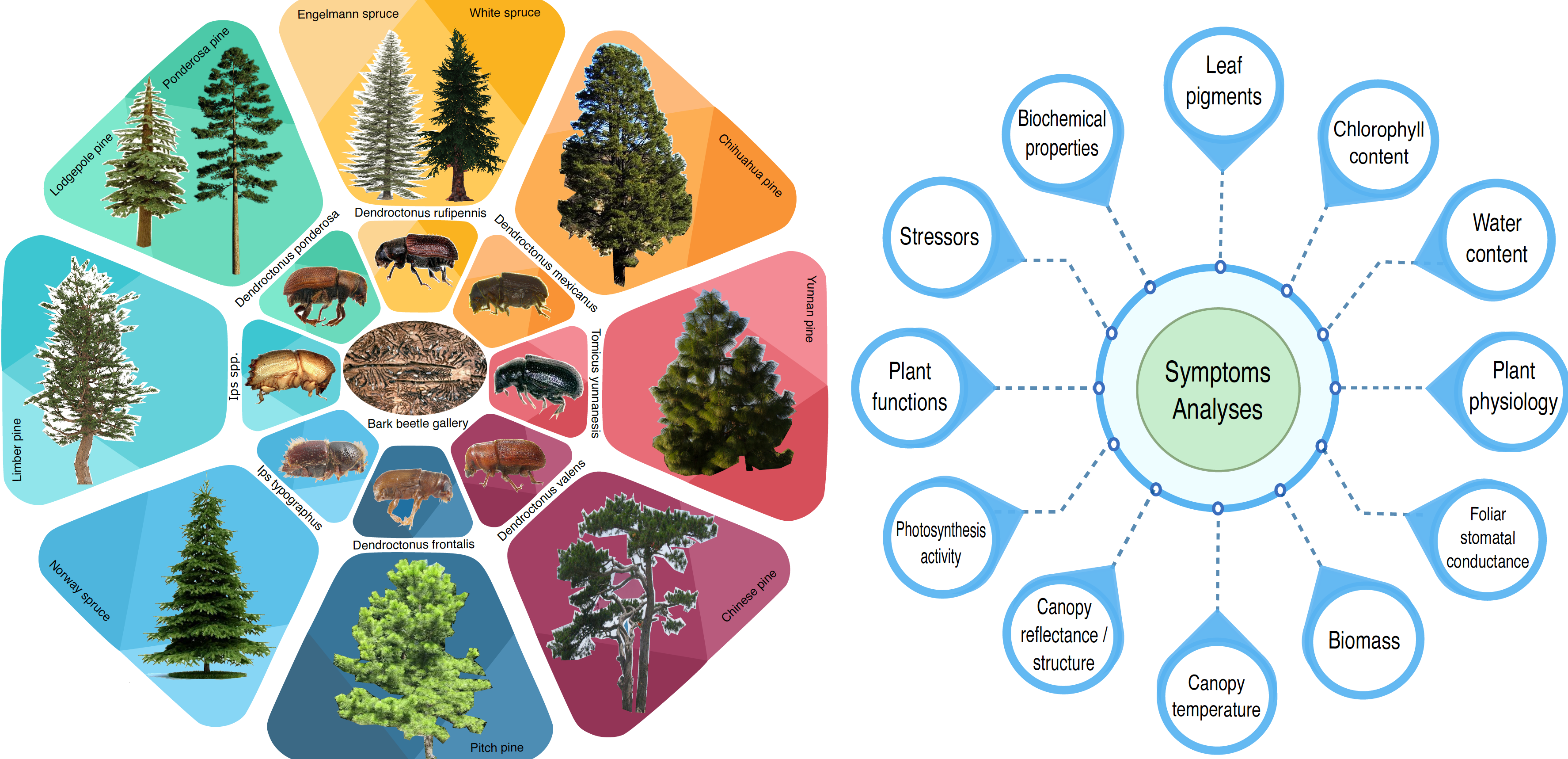}
\vspace{-0.2cm}
\caption{Examples of (left): bark beetle species and host trees, and (right): different symptoms analyses of attacked trees.} 
\vspace{-0.3cm}
\label{fig:beetles} 
\end{figure} 
\begin{table*}[!tbp]
\rowcolors{2}{}{gray!10}
\caption{{Bark beetle species, dominated trees in experimental sites, attack phases, and publication years.}} 
\vspace{-0.3cm}
\centering 
\resizebox*{!}{\dimexpr\textheight-1\baselineskip\relax}{
\begin{tabular}{c c c c c c} 
\hline
& Year & Ref. & Attack Phase & Dominated Tree Species & Bark Beetle Species \protect\\ 
\hline\hline

\global\let\CT@@do@color\relax \multirow{25}{*}{\rotatebox[origin=c]{90}{Focus of Our Review}} & \global\let\CT@@do@color\oriCT@@do@color 2012 & \cite{AngularVegIndex_Spectroscopy_Bavarian} & Green attack, Red attack & Norway spruce & European spruce bark beetle (Ips typographus) \protect\\ 

\global\let\CT@@do@color\relax & \global\let\CT@@do@color\oriCT@@do@color 2013 & \cite{EarlyDet_TerraSarX_RapidEye} & Green attack & Norway spruce & European spruce bark beetle (Ips typographus)     \protect\\ 

\global\let\CT@@do@color\relax & \global\let\CT@@do@color\oriCT@@do@color 2013 & \cite{Forecast_Vital_Hyperspect} & Green attack, Red attack, Gray attack & Norway Spruce & European spruce bark beetle (Ips typographus)  \protect\\ 

\global\let\CT@@do@color\relax & \global\let\CT@@do@color\oriCT@@do@color 2014 & \cite{ObjectExtract_LandsatSPOT} & Green attack, Red attack & Norway spruce & European spruce bark beetle (Ips typographus) \protect\\ 

\global\let\CT@@do@color\relax & \global\let\CT@@do@color\oriCT@@do@color 2014 & \cite{Assess_Hyperspec_Mortality} & Green attack, Red attack, Gray attack & Norway spruce & European spruce bark beetle (Ips typographus)  \protect\\ 
 
\global\let\CT@@do@color\relax & \global\let\CT@@do@color\oriCT@@do@color 2014 & \cite{EarlyDet_NorwaySpruce_WorldView2} & Green attack, Red attack, Gray attack & Norway spruce & \thead{European spruce bark beetle (Ips typographus), \\ Spruce beetle (Dendroctonus rufipennis)} \protect\\ 

\global\let\CT@@do@color\relax & \global\let\CT@@do@color\oriCT@@do@color 2015 & \cite{HyperRS_MPB_Emphasis_Previsual} & Green attack, Red attack & Lodgepole pine & Mountain pine beetle (Dendroctonus ponderosae)  \protect\\ 

\global\let\CT@@do@color\relax & \global\let\CT@@do@color\oriCT@@do@color 2017 & \cite{SpectEvid_Early_Engelmann} & Green attack & Engelmann spruce & Spruce beetle (Dendroctonus rufipennis)    \protect\\ 

\global\let\CT@@do@color\relax & \global\let\CT@@do@color\oriCT@@do@color 2018 & \cite{MPB_BlackHills_SouthDakota} & Green attack & Ponderosa pine & Mountain pine beetle (Dendroctonus ponderosae)    \protect\\ 

\global\let\CT@@do@color\relax & \global\let\CT@@do@color\oriCT@@do@color 2019 & \cite{Sentinel2_GA_Landsat8} & Green attack & Norway spruce, & European spruce bark beetle (Ips typographus)   \protect\\ 

\global\let\CT@@do@color\relax & \global\let\CT@@do@color\oriCT@@do@color 2019 & \cite{Sens_LandSatOLI_TIRS_Foliar} & Green attack, Gray attack & Norway spruce & European spruce bark beetle (Ips typographus)   \protect\\ 

\global\let\CT@@do@color\relax & \global\let\CT@@do@color\oriCT@@do@color 2019 & \cite{PineShootBeetle_Hyperspect_LiDAR} & Green attack, Red attack, Gray attack & Yunnan pine & Pine shoot beetle (Tomicus yunnanensis)   \protect\\

\global\let\CT@@do@color\relax & \global\let\CT@@do@color\oriCT@@do@color 2019 & \cite{UAV_PrecDetection_BBInf} & Green attack, Red attack, Gray attack & Norway spruce & European spruce bark beetle (Ips typographus)   \protect\\ 

\global\let\CT@@do@color\relax & \global\let\CT@@do@color\oriCT@@do@color 2020 & \cite{MultiTemp_HyperMultispect_NorwaySpruce} & Green attack & Norway spruce & European spruce bark beetle (Ips typographus)    \protect\\ 

\global\let\CT@@do@color\relax & \global\let\CT@@do@color\oriCT@@do@color 2020 & \cite{CrownExt_UASMultispec_MixedForest} & Green attack, Yellow attack & Norway spruce & European spruce bark beetle (Ips typographus)   \protect\\

\global\let\CT@@do@color\relax & \global\let\CT@@do@color\oriCT@@do@color 2020 & \cite{Monitor_CentrEuro_Validate} & Green attack, Red attack, Gray attack & Norway Spruce & European spruce bark beetle (Ips typographus)    \protect\\ 

\global\let\CT@@do@color\relax & \global\let\CT@@do@color\oriCT@@do@color 2021 & \cite{EarlyDet_EuropeSpruce_NDRS} & Green attack & Norway spruce & European spruce bark beetle (Ips typographus)     \protect\\ 


\global\let\CT@@do@color\relax & \global\let\CT@@do@color\oriCT@@do@color 2021 & \cite{UAS_Multispect_DL} & Green attack, Yellow attack & Norway spruce & European spruce bark beetle (Ips typographus)  \protect\\ 


\global\let\CT@@do@color\relax & \global\let\CT@@do@color\oriCT@@do@color 2021 & \cite{EarlyDet_Norway_CentEurope_Sentinel} & Green attack, Red attack & Norway spruce & Ips beetles (Ips spp.)    \protect\\ 

\global\let\CT@@do@color\relax & \global\let\CT@@do@color\oriCT@@do@color 2021 & \cite{Map_Gyrocopter_HyperField} & Green attack, Red attack & Norway spruce & European spruce bark beetle (Ips typographus)    \protect\\ 

\global\let\CT@@do@color\relax & \global\let\CT@@do@color\oriCT@@do@color 2021 & \cite{Mapping_Boreal_HealthStatus} & Green attack, Red attack, Gray attack & White spruce & Spruce beetle (Dendroctonus rufipennis)    \protect\\ 



\global\let\CT@@do@color\relax & \global\let\CT@@do@color\oriCT@@do@color 2022 & \cite{Comp_FieldRS_DetBB} & Green attack, Red attack & Norway spruce & European spruce bark beetle (Ips typographus)  \protect\\ 

\global\let\CT@@do@color\relax & \global\let\CT@@do@color\oriCT@@do@color 2022 & \cite{Detection_UAV_YOLOs} & Green attack, Yellow attack, Red attack, Gray attack & Norway spruce & European spruce bark beetle (Ips typographus)    \protect\\ 

\hdashline

\global\let\CT@@do@color\relax \multirow{38}{*}{\rotatebox[origin=c]{90}{Late Detection}} & 
\global\let\CT@@do@color\oriCT@@do@color 2003 & \cite{MPB_RedAttack_Strat_Landsat_BC} & Red attack & Lodgepole pine & Mountain pine beetle (Dendroctonus ponderosae)  \protect\\ 

\global\let\CT@@do@color\relax & \global\let\CT@@do@color\oriCT@@do@color 2005 & \cite{RedAttack_MPB_HSR_Sat} & Red attack & Lodgepole pine & Mountain pine beetle (Dendroctonus ponderosae)  \protect\\ 

\global\let\CT@@do@color\relax & \global\let\CT@@do@color\oriCT@@do@color 2006 & \cite{EstProb_MPB_RedAttack} & Red attack & Lodgepole pine & Mountain pine beetle (Dendroctonus ponderosae)   \protect\\ 
 
\global\let\CT@@do@color\relax & \global\let\CT@@do@color\oriCT@@do@color 2006 & \cite{IntegRS_AncillaryData_CharMPB} & Red attack & \thead{Lodgepole pine} & Mountain pine beetle (Dendroctonus ponderosae)    \protect\\ 

\global\let\CT@@do@color\relax & \global\let\CT@@do@color\oriCT@@do@color 2009 & \cite{Map_WhiteBark_MPB_Sat} & Red attack & Lodgepole pine, whitebark pine & Mountain pine beetle (Dendroctonus ponderosae)  \protect\\ 

\global\let\CT@@do@color\relax & \global\let\CT@@do@color\oriCT@@do@color 2010 & \cite{AssessCanopy_MPB_GeoEye_Sat} & Red attack, Gray attack & Lodgepole pine & Mountain pine beetle (Dendroctonus ponderosae)   \protect\\ 

\global\let\CT@@do@color\relax & \global\let\CT@@do@color\oriCT@@do@color 2011 & \cite{EvalPoten_MultiSpect_MultipleStage} & Red attack, Gray attack & Lodgepole pine & Mountain pine beetle (Dendroctonus ponderosae)  \protect\\ 

\global\let\CT@@do@color\relax & \global\let\CT@@do@color\oriCT@@do@color 2013 & \cite{Eval_DetectBB_SingMultiData_Landsat} & Red attack & \thead{Engelmann spruce, subalpine fir, limber pine, \\ lodgepole pine, Ponderosa pine, Douglas-fir} & Unspecified    \protect\\ 

\global\let\CT@@do@color\relax & \global\let\CT@@do@color\oriCT@@do@color 2013 & \cite{Quanti_MixedSpecies_MultiTempHighRes_Sat} & Gray attack & Piñon pine, Juniper, Ponderosa pine & Unspecified    \protect\\ 

\global\let\CT@@do@color\relax & \global\let\CT@@do@color\oriCT@@do@color 2013 & \cite{LiveDead_Basal_LiDAR_Journal} & Gray attack & Spruces, Pines & \thead{Spruce beetle (Dendroctonus rufipennis), \\ Mountain pine beetle (Dendroctonus ponderosae)}   \protect\\

\global\let\CT@@do@color\relax & \global\let\CT@@do@color\oriCT@@do@color 2014 & \cite{Spatial_Multidate_SPOT_Landsat} & Red attack, Gray attack & Norway spruce & European spruce bark beetle (Ips typographus)  \protect\\ 

\global\let\CT@@do@color\relax & \global\let\CT@@do@color\oriCT@@do@color 2014 & \cite{MPB_GrowthTrend_Landsat} & Gray attack & Lodgepole pine & Mountain pine beetle (Dendroctonus ponderosae)  \protect\\ 

\global\let\CT@@do@color\relax & \global\let\CT@@do@color\oriCT@@do@color 2014 & \cite{Toward_UAV_ForestMonitor} & Red attack, Gray attack & Sitka spruce & European spruce bark beetle (Ips typographus) \protect\\ 

\global\let\CT@@do@color\relax & \global\let\CT@@do@color\oriCT@@do@color 2015 & \cite{Photogram_Hyperspectral_TreeLevel} & Yellow attack, Gray attack & Norway spruce & European spruce bark beetle (Ips typographus)   \protect\\ 

\global\let\CT@@do@color\relax & \global\let\CT@@do@color\oriCT@@do@color 2015 & \cite{Detect_MPB_Ponderosa_HighResAerial} & Red attack & Ponderosa pine & Mountain pine beetle (Dendroctonus ponderosae)     \protect\\ 

\global\let\CT@@do@color\relax & \global\let\CT@@do@color\oriCT@@do@color 2015 & \cite{SpectTemporal_Defoliat_Landsat} & Gray attack & Douglas-fir, Ponderosa pine, Lodgepole pine & \thead{Mountain pine beetle (Dendroctonus ponderosae), \\ Western spruce budworm}  \protect\\ 

\global\let\CT@@do@color\relax & \global\let\CT@@do@color\oriCT@@do@color 2015 & \cite{Spruce_HighMedium_Remote} & Gray attack & Spruce, Fir & Spruce beetle (Dendroctonus rufipennis)  \protect\\ 

\global\let\CT@@do@color\relax & \global\let\CT@@do@color\oriCT@@do@color 2016 & \cite{PercentMortality_MPB_Damage} & Red attack, Gray attack & Lodgepole pine & Mountain pine beetle (Dendroctonus ponderosae)     \protect\\ 

\global\let\CT@@do@color\relax & \global\let\CT@@do@color\oriCT@@do@color 2018 & \cite{UrbanForest_Hyperspec_UAV_Aircraft} & Yellow attack, Gray attack & Norway spruce & European spruce bark beetle (Ips typographus)    \protect\\ 

\global\let\CT@@do@color\relax & \global\let\CT@@do@color\oriCT@@do@color 2018 & \cite{Framework_Ecoregion_Spaceborne} & Red attack & Ponderosa pine & Dendroctonus spp.  \protect\\ 

\global\let\CT@@do@color\relax & \global\let\CT@@do@color\oriCT@@do@color 2018 & \cite{RapidEye_Detect_AreaSpruce} & Red attack, Gray attack & Norway spruce & European spruce bark beetle (Ips typographus) \protect\\

\global\let\CT@@do@color\relax & \global\let\CT@@do@color\oriCT@@do@color 2019 & \cite{BB_Fir_DL} & Red attack, Gray attack & Pine, Fir & Four-eyed fir bark beetle (Polygraphus proximus)   \protect\\ 

\global\let\CT@@do@color\relax & \global\let\CT@@do@color\oriCT@@do@color 2019 & \cite{Comp_WorldViewLandsat_SVMNeuralNet} & Red attack, Gray attack & Spruce & European spruce bark beetle (Ips typographus)  \protect\\ 

\global\let\CT@@do@color\relax & \global\let\CT@@do@color\oriCT@@do@color 2019 & \cite{DeadWood_FCNDenseNet} & Gray attack & N/A & Unspecified    \protect\\ 

\global\let\CT@@do@color\relax & \global\let\CT@@do@color\oriCT@@do@color 2020 & \cite{GF2Sentinel2_Turpentine_China} & Red attack, Gray attack & Chinese pine & Red turpentine beetle (Dendroctonus valens)    \protect\\

\global\let\CT@@do@color\relax & \global\let\CT@@do@color\oriCT@@do@color 2020 & \cite{Class_HealthStat_UAS_Multispect} & Red attack, Gray attack & Norway spruce, Scots pine & Unspecified    \protect\\ 

\global\let\CT@@do@color\relax & \global\let\CT@@do@color\oriCT@@do@color 2020 & \cite{DistSeg_Sat_DL} & Red attack, Gray attack & Sakhalin fir & Unspecified    \protect\\ 

\global\let\CT@@do@color\relax & \global\let\CT@@do@color\oriCT@@do@color 2020 & \cite{Sat_DecisionSupport_TANABBO} & Gray attack & Norway spruce & European spruce bark beetle (Ips typographus)   \protect\\ 

\global\let\CT@@do@color\relax & \global\let\CT@@do@color\oriCT@@do@color 2020 & \cite{Habitat_Dynamics_NorwayDieback} & Gray attack & Norway spruce & European spruce bark beetle (Ips typographus)  \protect\\ 

\global\let\CT@@do@color\relax & \global\let\CT@@do@color\oriCT@@do@color 2021 & \cite{DetectBB_Threshold_CellAutomata} & Yellow attack, Red attack, Gray attack & Pine & Mexican pine beetle (Dendroctonus Maxicanus)  \protect\\ 

\global\let\CT@@do@color\relax & \global\let\CT@@do@color\oriCT@@do@color 2021 & \cite{Early_Hyper_Results_Bavaria} & Yellow attack, Red attack, Gray attack & Norway spruce & Unspecified    \protect\\ 

\global\let\CT@@do@color\relax & \global\let\CT@@do@color\oriCT@@do@color 2021 & \cite{SubtleChange_Landsat_MPBSpruce} & Red attack, Gray attack & Lodgepole pine &  \thead{Mountain pine beetle (Dendroctonus ponderosae), \\ Spruce beetle (Dendroctonus rufipennis)}    \protect\\ 

\global\let\CT@@do@color\relax & \global\let\CT@@do@color\oriCT@@do@color 2021 & \cite{ML_SpatialDist_BBI} & Gray attack & Norway spruce & European spruce bark beetle (Ips typographus)  \protect\\ 

\global\let\CT@@do@color\relax & \global\let\CT@@do@color\oriCT@@do@color 2021 & \cite{SickFir_UAV_DL} & Gray attack & Maries fir & Sakhalin fir bark beetle (Polygraphus proximus)    \protect\\ 

 \global\let\CT@@do@color\relax & \global\let\CT@@do@color\oriCT@@do@color 2022 & \cite{Multispectral_Benefits_Spruce_Decline} & Yellow attack, Red attack, Gray attack & Norway spruce & European spruce bark beetle (Ips typographus)    \protect\\ 
 
\global\let\CT@@do@color\relax & \global\let\CT@@do@color\oriCT@@do@color 2021 & \cite{CrosInteract_Mortality} & Gray attack & Ponderosa pine & Western pine beetle (Dendroctonus brevicomis)    \protect\\

\global\let\CT@@do@color\relax & \global\let\CT@@do@color\oriCT@@do@color 2022 & \cite{Landsat_SouthPineBeet_Severity} & Red attack & Pitch pine & Southern pine beetle (Dendroctonus frontalis)  \protect\\ 

\global\let\CT@@do@color\relax & \global\let\CT@@do@color\oriCT@@do@color 2022 & \cite{Northwenmost_Model_Climate} & Gray attack & Siberian spruce & European spruce bark beetle (Ips typographus)   \protect\\

\hline 
  \end{tabular}
  \label{tab_BB_Tree}  
  }
\end{table*}
\begin{table*}[!b]
\vspace{-0.3cm}
\rowcolors{2}{}{gray!10}
\caption{Experiment site characteristics, conditions, and data collection dates.} 
\vspace{-0.3cm}
\centering 
 \resizebox{\textwidth}{!}{
\begin{tabular}{c c c c c c c} 
\hline
& Ref. & Study Site & Country & Area & Condition & Data Collection Date  \protect\\ 
\hline\hline

\global\let\CT@@do@color\relax \multirow{28}{*}{\rotatebox[origin=c]{90}{Focus of Our Review}}  & \global\let\CT@@do@color\oriCT@@do@color \cite{SpectEvid_Early_Engelmann} & Rocky Mountains & USA & N/A & N/A & Sep 2014     \protect\\ 

\global\let\CT@@do@color\relax  & \global\let\CT@@do@color\oriCT@@do@color \cite{MPB_BlackHills_SouthDakota} & Black Hills & USA & 9781 ha & Ponderosa pine forest; Humid island & Apr 2015   \protect\\ 

\global\let\CT@@do@color\relax  & \global\let\CT@@do@color\oriCT@@do@color \cite{Mapping_Boreal_HealthStatus} & Denali State Park & USA & 25000 ha & Alaskan boreal forest & July 2018   \protect\\ 

\global\let\CT@@do@color\relax  & \global\let\CT@@do@color\oriCT@@do@color \cite{HyperRS_MPB_Emphasis_Previsual} & British Columbia & Canada & 250 ha & N/A & Mid-September 2008, Early-July 2009    \protect\\ 

\global\let\CT@@do@color\relax & \global\let\CT@@do@color\oriCT@@do@color  \cite{EarlyDet_TerraSarX_RapidEye} & Biberach forest district & Germany & N/A & Forest stand age: 10-100 years; Tree heights: 10-35 m & May 2009     \protect\\ 

\global\let\CT@@do@color\relax  & \global\let\CT@@do@color\oriCT@@do@color \cite{Sentinel2_GA_Landsat8} & Bavarian Forest National Park & Germany & 24369 ha & N/A & July 2016    \protect\\ 

\global\let\CT@@do@color\relax  & \global\let\CT@@do@color\oriCT@@do@color \cite{ObjectExtract_LandsatSPOT} & Bavarian Forest National Park & Germany & N/A & Subalpine forest & 2001–2011  \protect\\ 

\global\let\CT@@do@color\relax  & \global\let\CT@@do@color\oriCT@@do@color \cite{Sens_LandSatOLI_TIRS_Foliar} & Bavarian Forest National Park & Germany & 24.222 ha & N/A & May/Ju/Aug 2016    \protect\\ 

\global\let\CT@@do@color\relax  & \global\let\CT@@do@color\oriCT@@do@color \cite{Forecast_Vital_Hyperspect} & Bavarian Forest National Park & Germany & 13000 ha & Old region & 1988-2011  \protect\\ 

\global\let\CT@@do@color\relax  & \global\let\CT@@do@color\oriCT@@do@color \cite{AngularVegIndex_Spectroscopy_Bavarian} & Bavarian Forest National Park & Germany \& Czechia & 3430 ha & Sub-alpine spruce forests & 2009/2010  \protect\\ 

\global\let\CT@@do@color\relax  & \global\let\CT@@do@color\oriCT@@do@color \cite{Assess_Hyperspec_Mortality} & Bavarian Forest National Park & Germany \& Czechia & N/A & Sub-alpine spruce forests & Aug 2009/2010  \protect\\ 

\global\let\CT@@do@color\relax  & \global\let\CT@@do@color\oriCT@@do@color \cite{CrownExt_UASMultispec_MixedForest} & Klánovice forest & Czechia & 1030 ha & Mixed broadleaved \& coniferous tree species & Feb/March 2020    \protect\\

\global\let\CT@@do@color\relax  & \global\let\CT@@do@color\oriCT@@do@color \cite{UAS_Multispect_DL} & Klánovice Forest & Czechia & N/A & Urban heat islands & Sep 2020   \protect\\ 

\global\let\CT@@do@color\relax  & \global\let\CT@@do@color\oriCT@@do@color \cite{EarlyDet_Norway_CentEurope_Sentinel} & Bohemian–Moravian highlands & Czechia & N/A & N/A & 2017-2020     \protect\\ 

\global\let\CT@@do@color\relax  & \global\let\CT@@do@color\oriCT@@do@color \cite{UAV_PrecDetection_BBInf} & Krkonoše Mountains National Park & Czechia & 10.7 ha & Spruce forest & June/Aug/Oct 2017    \protect\\ 

\global\let\CT@@do@color\relax  & \global\let\CT@@do@color\oriCT@@do@color \cite{Monitor_CentrEuro_Validate} & Mendel University (Forest area) & Czechia & 1000 ha & Mixed forest & 2018-2020     \protect\\ 

\global\let\CT@@do@color\relax  & \global\let\CT@@do@color\oriCT@@do@color \cite{Comp_FieldRS_DetBB} & Southern part & Czechia & 300 ha & Norway spruce forest (age classes 30–40 years \& 90–100 years) & Mar-Sep 2020     \protect\\ 

\global\let\CT@@do@color\relax  & \global\let\CT@@do@color\oriCT@@do@color \cite{MultiTemp_HyperMultispect_NorwaySpruce} & Espoo & Finland & 10 ha & N/A & Aug-Oct 2019   \protect\\ 

\global\let\CT@@do@color\relax  & \global\let\CT@@do@color\oriCT@@do@color \cite{EarlyDet_EuropeSpruce_NDRS} & Remningstorp & Sweden & 1602 ha & N/A & Apr-Oct 2018/2019     \protect\\ 
 
\global\let\CT@@do@color\relax  & \global\let\CT@@do@color\oriCT@@do@color \cite{EarlyDet_NorwaySpruce_WorldView2} & Styria & Austria & 1100 ha & Spruce-fir forest with admixture of beech, larch \& sycamore maple & Jun/Jul/Oct 2010  \protect\\ 

\global\let\CT@@do@color\relax  & \global\let\CT@@do@color\oriCT@@do@color \cite{Detection_UAV_YOLOs} & West Balkan Mountains & Bulgaria & N/A & N/A & Aug/Sep 2017   \protect\\ 

\global\let\CT@@do@color\relax  & \global\let\CT@@do@color\oriCT@@do@color \cite{PineShootBeetle_Hyperspect_LiDAR} & Tianfeng Mountain & China & 1000 ha & Plantation forests & Sep 2018     \protect\\

\hdashline



\global\let\CT@@do@color\relax \multirow{42}{*}{\rotatebox[origin=c]{90}{Late Detection}} & \global\let\CT@@do@color\oriCT@@do@color \cite{EstProb_MPB_RedAttack} & Lolo National Forest & USA & N/A & Mountainous, Mixed conifer forest & Aug 1999/ Aug 2002  \protect\\ 

\global\let\CT@@do@color\relax & \global\let\CT@@do@color\oriCT@@do@color \cite{Eval_DetectBB_SingMultiData_Landsat} & Colorado \& Wyoming & USA & 9400 ha & N/A & July/Aug 2011, (base image: Aug 2002)    \protect\\ 

\global\let\CT@@do@color\relax & \global\let\CT@@do@color\oriCT@@do@color \cite{Detect_MPB_Ponderosa_HighResAerial} & Colorado & USA & 35 ha & Ponderosa pine-dominated forests & Sep 2011     \protect\\

\global\let\CT@@do@color\relax & \global\let\CT@@do@color\oriCT@@do@color \cite{Framework_Ecoregion_Spaceborne} & Sierra Nevada & USA & N/A & Mediterranean climate: Hot dry summers \& cool wet winters & 2013-2016   \protect\\ 

\global\let\CT@@do@color\relax & \global\let\CT@@do@color\oriCT@@do@color \cite{Landsat_SouthPineBeet_Severity} & New York & USA & N/A & Moderate humid climate & July-Oct 2000-2011, 2013-2020    \protect\\ 

\global\let\CT@@do@color\relax & \global\let\CT@@do@color\oriCT@@do@color \cite{AssessCanopy_MPB_GeoEye_Sat} & Wyoming & USA & 22 ha & Snowy Mountain range & 2009   \protect\\ 

\global\let\CT@@do@color\relax & \global\let\CT@@do@color\oriCT@@do@color \cite{EvalPoten_MultiSpect_MultipleStage} & Arapahoe National Forest, Colorado & USA & 9400 ha & N/A & Aug 2008  \protect\\ 

\global\let\CT@@do@color\relax & \global\let\CT@@do@color\oriCT@@do@color \cite{SubtleChange_Landsat_MPBSpruce} & Colorado (2 sites) & USA & 484200/1024100 ha & Even-aged stands & 2001-2019     \protect\\ 

\global\let\CT@@do@color\relax & \global\let\CT@@do@color\oriCT@@do@color \cite{Quanti_MixedSpecies_MultiTempHighRes_Sat} & Northern New Mexico & USA & 4600 ha & N/A & 2002, 2006/2011    \protect\\ 

\global\let\CT@@do@color\relax & \global\let\CT@@do@color\oriCT@@do@color \cite{MPB_GrowthTrend_Landsat} & Colorado & USA & 478300 ha & Evergreen forests; Hot \& dry summer & 2000-2011   \protect\\ 

\global\let\CT@@do@color\relax & \global\let\CT@@do@color\oriCT@@do@color \cite{Spruce_HighMedium_Remote} & Colorado & USA & 180000 ha & Spruce–fir forest & 1995-1998/2000-2013   \protect\\ 

\global\let\CT@@do@color\relax & \global\let\CT@@do@color\oriCT@@do@color \cite{PercentMortality_MPB_Damage} & Helena National Forest & USA & 57260 ha & Mountainous region; Coniferous forest \& Montane steppe & July 2013     \protect\\ 

\global\let\CT@@do@color\relax & \global\let\CT@@do@color\oriCT@@do@color \cite{LiveDead_Basal_LiDAR_Journal} & Alaska, Arizona, Colorado, Idaho, Oregon & USA & N/A & N/A & 2008-2010  \protect\\  

\global\let\CT@@do@color\relax & \global\let\CT@@do@color\oriCT@@do@color \cite{Map_WhiteBark_MPB_Sat} & Idaho & USA & N/A & Coniferous forest & Jul 2005   \protect\\ 

\global\let\CT@@do@color\relax & \global\let\CT@@do@color\oriCT@@do@color \cite{CrosInteract_Mortality} & Sierra Nevada (32 sites) & USA & 40 ha & Yellow pine/mixed-conifer forests & Apr-July 2018    \protect\\

\global\let\CT@@do@color\relax & \global\let\CT@@do@color\oriCT@@do@color \cite{MPB_RedAttack_Strat_Landsat_BC} & British Columbia & Canada & 507000 ha & Lodgepole pine \& white spruce forests &  Aug-Sep 1999  \protect\\ 

\global\let\CT@@do@color\relax  & \global\let\CT@@do@color\oriCT@@do@color \cite{RedAttack_MPB_HSR_Sat} & British Columbia (2 sites) & Canada & 19500 ha & \thead{Average age of lodgepole pine stands: 105 years; \\Dominated by Sub-Boreal Spruce; Biogeoclimatic zone} & Oct 2002  \protect\\ 

\global\let\CT@@do@color\relax & \global\let\CT@@do@color\oriCT@@do@color \cite{IntegRS_AncillaryData_CharMPB} & British Columbia & Canada & 544000 ha & N/A & Aug 2001, Jul 2003  \protect\\ 

\global\let\CT@@do@color\relax & \global\let\CT@@do@color\oriCT@@do@color \cite{SpectTemporal_Defoliat_Landsat} & British Columbia & Canada & 14970000 ha & \thead{Douglas-fir Forest; Montane Spruce/ Ponderosa Pine/\\ Sub-Boreal Pine/ Spruce zones} & 1990-2013   \protect\\  

\global\let\CT@@do@color\relax  & \global\let\CT@@do@color\oriCT@@do@color \cite{DetectBB_Threshold_CellAutomata} & Anteojitos & Mexico & 4 ha & Pine forest; Subhumid temperate with rain in summer & June-Aug 2020   \protect\\ 

\global\let\CT@@do@color\relax  & \global\let\CT@@do@color\oriCT@@do@color \cite{Spatial_Multidate_SPOT_Landsat} & Bavarian Forest National Park & Germany & 13000 ha & N/A & 2001-2011   \protect\\ 

\global\let\CT@@do@color\relax  & \global\let\CT@@do@color\oriCT@@do@color \cite{Early_Hyper_Results_Bavaria} & Altotting, Bavaria, & Germany & N/A & Beech forest with few oaks; Subcontinental climate & 2013, 2014     \protect\\

\global\let\CT@@do@color\relax & \global\let\CT@@do@color\oriCT@@do@color \cite{DeadTree_PixObj_WorldView} & Southern Black Forest & Germany & 2700 ha & N/A & Sep 2017   \protect\\ 

\global\let\CT@@do@color\relax & \global\let\CT@@do@color\oriCT@@do@color \cite{Toward_UAV_ForestMonitor} & Bavarian Forest National Park & Germany & N/A & N/A &  2011   \protect\\ 

\global\let\CT@@do@color\relax & \global\let\CT@@do@color\oriCT@@do@color \cite{RapidEye_Detect_AreaSpruce} & Bavarian Forest National Park & Germany & N/A & N/A & 2011/2012  \protect\\ 
 
\global\let\CT@@do@color\relax & \global\let\CT@@do@color\oriCT@@do@color \cite{DeadWood_FCNDenseNet} & Bavarian Forest National Park & Germany & N/A & Complicated temperate forest & August 2012     \protect\\ 
 
\global\let\CT@@do@color\relax & \global\let\CT@@do@color\oriCT@@do@color \cite{Class_HealthStat_UAS_Multispect} & Oplany & Czechia & N/A & \thead{Even-aged coniferous species; Admixture of several \\ broadleaf species, small percentage of understory} & Apr/Jul 2019     \protect\\ 

\global\let\CT@@do@color\relax & \global\let\CT@@do@color\oriCT@@do@color \cite{Sat_DecisionSupport_TANABBO} & Horní Planá & Czechia & 16569 ha & \thead{Coniferous forests; Stands younger than 40 years (23.5\%) \\ \& older than 100 years (32\%)} & 2007-2010    \protect\\ 

\global\let\CT@@do@color\relax & \global\let\CT@@do@color\oriCT@@do@color \cite{ML_SpatialDist_BBI} & Horní Planá region & Czechia & 16569 ha & \thead{Spruce forests; Stands younger than 40 years (23.5\%) \\ \& older than 100 years (32\%)} & 2003-2012  \protect\\  
 
\global\let\CT@@do@color\relax & \global\let\CT@@do@color\oriCT@@do@color \cite{Comp_WorldViewLandsat_SVMNeuralNet} & Sumava National Park & Czechia & 10500 ha & Mountain spruce forests, Peat bogs, Mountain meadows & Sep/Oct 2015   \protect\\  
 
\global\let\CT@@do@color\relax  & \global\let\CT@@do@color\oriCT@@do@color \cite{Photogram_Hyperspectral_TreeLevel} & Lahti & Finland & 5000 ha & Urban forests & Aug 2013   \protect\\ 

\global\let\CT@@do@color\relax  & \global\let\CT@@do@color\oriCT@@do@color \cite{UrbanForest_Hyperspec_UAV_Aircraft} & Lahti & Finland & 407 ha & Urban forests & Aug/Sep 2013     \protect\\ 

\global\let\CT@@do@color\relax  & \global\let\CT@@do@color\oriCT@@do@color \cite{Multispectral_Benefits_Spruce_Decline} & Helsinki city central Park & Finland & 20 ha & N/A & May/Sep 2020 \protect\\ 

\global\let\CT@@do@color\relax & \global\let\CT@@do@color\oriCT@@do@color \cite{Habitat_Dynamics_NorwayDieback} & Białowieża Forest & Poland & 13000 ha & \thead{Complex forest; Average stand age: 89 years; \\ Stands above 140 years old: 27.5\%} & 2015/2017    \protect\\ 

\global\let\CT@@do@color\relax & \global\let\CT@@do@color\oriCT@@do@color \cite{BB_Fir_DL} & Stolby State Nature Reserve & Russia & N/A & Mixed forest & July 2016, August 2018    \protect\\ 

\global\let\CT@@do@color\relax & \global\let\CT@@do@color\oriCT@@do@color \cite{DistSeg_Sat_DL} & Kunashir \& Sakhalin islands & Russia & N/A & Natural non-disturbed forests, Taiga \& coniferous forests & 2018-2019     \protect\\  

\global\let\CT@@do@color\relax & \global\let\CT@@do@color\oriCT@@do@color \cite{Northwenmost_Model_Climate} & Arkhangelsk & Russia & 300420 ha & Taiga area &  2001–2014 (14-year period)   \protect\\ 
 
\global\let\CT@@do@color\relax  & \global\let\CT@@do@color\oriCT@@do@color \cite{GF2Sentinel2_Turpentine_China} & Dahebei & China & 100 ha & Pure forest & Aug 2018    \protect\\ 

\global\let\CT@@do@color\relax & \global\let\CT@@do@color\oriCT@@do@color \cite{SickFir_UAV_DL} & Zao mountain (4 sites) & Japan & 18 ha & Density: 200 trees/ha; Fir trees: 41-103 years old; Average age: 72 years & Summer 2019     \protect\\

\hline 
  \end{tabular}
  \label{tab_BB_Env}  
  }
\end{table*}

\indent Tree-killing bark beetles deploy a similar basic tree-killing strategy. Aggregation on the host trees is usually triggered by the release of aggregation pheromones by attacking beetles within a short time. This enables beetles to overcome host resistance, resulting in tree mortality. 
Figure~{\ref{fig:beetles}} shows major tree-killing bark beetles, associated host trees in the northern hemisphere, and different symptoms used for analyses.
Most species can kill multiple species of hosts in the same genus. 
In almost all cases, attacks by these species result in tree mortality in the year of the attack. Thus, the physiological responses of host trees from the moment beetles attack to tree mortality are similar across species. Attacks by the beetles and their associated phytopathogenic fungi can shut down both nutrient and water flow along tree stems, causing cascading changes in tree foliage. Shortly after beetle attacks, reduced water flow along the tree stem can alter photosynthesis by affecting stomatal conductance. Lower stomatal conductance and decreased evapotranspiration of foliage reduce leaf water content and increase leaf temperature, compared to surrounding healthy trees. Thus, canopy reflectance changes are expected in GAtt trees. In addition, there are several biochemical changes in attacked trees that may affect tree survival. 
In summary, due to changes in water and nutrient flow that impact the foliage in tree crowns, attacked trees can be distinguishable from surrounding healthy trees. \\
\indent Although many studies have focused on bark beetle-host tree interactions {\mbox{\cite{Book_barkbeetle2021}}}, how bark beetles kill trees is not clear as several environmental conditions such as drought and other biological factors such as bark beetle-associated organisms can have a strong influence on the host tree-bark beetle interactions \mbox{\cite{Nadir_Zanganeh_2021}}. 
The physiological function of trees colonized by the bark beetle-symbiotic microbial complex usually suffers from two distinct but dependent interactions between bark beetles and their fungal symbionts. 
In particular, while bark beetles (both adults and larvae) consume tree phloem, which carries photosynthesis products such as sugars, fungal spores introduced by bark beetles germinate and fungal hyphae spread and penetrate water and nutrient-conducting tissues in the xylem. Penetration of host tissues by fungal hyphae disrupts water flow, which results in cellular dehydration. 
As a result of the combination of the disruptive and consumptive effects of bark beetle attacks and fungal infection, the allocation of plant nutrients such as carbohydrates are reduced and they are therefore no longer available for tree respiration and other critical functions. Thus, tree mortality occurs due to a combination of bark beetle and fungal attacks \mbox{\cite{Nadir_Zanganeh_2021}}.


\section{ {Remote Sensing Perspective}} \label{sec:RS} 
\noindent \textit{Remote Sensing} (RS) aims to gather information about a target (e.g., an object like a tree, an area like a forest, a stand, or an event like a wildfire or a pest outbreak) using single or multiple sensors that record electromagnetic radiation reflected, back-scattered, or emitted from the target, where the source of that radiation is either the sun (optical RS), the target (thermal RS) or the sensor itself (active RS, i.e., LiDAR and radar) \cite{Review_Guillermo,Review_Book_EarlyForestDisturbance,RS_ForestEcology_Management,RS_VegetationMapping_Review}. These recordings can, in turn, be related to some biophysical property of the target, like its foliage biomass or its temperature, or used to classify the target (e.g., GAtt phase). 
Successfully detecting bark beetle-induced tree mortality across large areas requires RS systems with suitable spatial, spectral, and temporal resolutions \cite{Review_Wulder_Challenges,Review_Guillermo,Review_Zabihi}. Spatial resolution refers to the minimum size of objects that the sensor can resolve. The spectral resolution is given by the number and width of spectral bands in which the sensor operates. Finally, the temporal resolution indicates the frequency at which new data are acquired over a given site. 
 {Although it may seem ideal to use high spatial-spectral-temporal resolutions to improve the separation of infection phases, it is crucial to consider that such high-density data can also lead to increased costs}.
The incorporation of high spatial-spectral-temporal resolutions could potentially also lead to two main issues: data complexity and model complexity for subsequent analyses. The complexity of data refers to the high volume and dimensionality of data, which demands complex models and optimization problems. As a result, a trade-off is often made between resolution and complexity.

\begin{table*}[!tbp]
\rowcolors{2}{}{gray!10}
\caption{Comparison of remote sensing systems to detect bark beetle attacks (bark beetle species, attack phases, publication years, and software are provided as further details). Abbreviations are denoted in Table~\ref{tab_Abbrev}.} 
\vspace{-0.3cm}
\label{Table_RS_All} 
\centering 
\resizebox*{!}{\dimexpr\textheight-2\baselineskip\relax}{
\begin{tabular}{c c c c c c c c c c c c} 
\hline
 &  & Year & Ref. & Attack Phase & Bark Beetle & Platform, Camera & Spectral Modality & Spectral Bands (ranges) & Spatial Resolution & Pre-Processing Software  \protect\\ %
\hline\hline

\global\let\CT@@do@color\relax \multirow{34}{*}{\rotatebox[origin=c]{90}{Satellite}} & \global\let\CT@@do@color\relax \multirow{17}{*}{\rotatebox[origin=c]{90}{Focus of Our Review}} & \global\let\CT@@do@color\oriCT@@do@color 2013 & \cite{EarlyDet_TerraSarX_RapidEye} & GAtt & ESBB & RapidEye, TerraSAR-X & MS & REye: RGB, Red-edge, NIR; TSAR: X-band & 5/2m & GAMMA   \protect\\  

\global\let\CT@@do@color\relax &  & \global\let\CT@@do@color\oriCT@@do@color 2014 & \cite{EarlyDet_NorwaySpruce_WorldView2} & {GAtt, RAtt, GRAtt} & ESBB, SBB & WorldView-2 & MS &  RGB, NIR1, NIR2, Red-edge, Coastal, Yellow, PAN & 0.5/2cm & ENVI, ERDAS Imagine    \protect\\  

\global\let\CT@@do@color\relax  &  & \global\let\CT@@do@color\oriCT@@do@color 2017 & \cite{SpectEvid_Early_Engelmann} & GAtt & SBB & Landsat TM & MS & RGB, NIR, SWIR1, SWIR2 & 30m & ENVI, ViewSpec Pro   \protect\\  %

\global\let\CT@@do@color\relax  &  & \global\let\CT@@do@color\oriCT@@do@color 2018 & \cite{MPB_BlackHills_SouthDakota} & GAtt & MPB & WorldView-2 & MS &  RGB, NIR1, NIR2, Red-edge, Coastal, Yellow, PAN & 1.85/0.5m & \thead{ENVI, ERDSA Imagine, ArcMap}   \protect\\  

\global\let\CT@@do@color\relax  &  & \global\let\CT@@do@color\oriCT@@do@color 2019 & \cite{Sens_LandSatOLI_TIRS_Foliar} & GAtt, GRAtt & ESBB & Landsat-8 (OLI \& TIRS) & MS &  \thead{9 bands from OLI,\\ thermal band from TIRS} &  100/30m & FLAASH   \protect\\  %

\global\let\CT@@do@color\relax  &  & \global\let\CT@@do@color\oriCT@@do@color 2019 & \cite{Sentinel2_GA_Landsat8}  & GAtt & ESBB & Sentinel-2, Landsat-8 & MS &  \thead{L8OLI: RGB, NIR, SWIR1-2;\\ Sen2: RGB, Red-edge 1-3, NIR, NIR(a), SWIR 1-2} &  10/30m & \thead{MODTRAN4, ENVI,\\ SEN2COR, ArcMap}   \protect\\  

\global\let\CT@@do@color\relax  &  & \global\let\CT@@do@color\oriCT@@do@color 2020 & \cite{Monitor_CentrEuro_Validate} & {GAtt, RAtt, GRAtt} & ESBB & Sentinel-2 & MS & \thead{RGB, Coastal aerosol, Red-edge 1-3, NIR,\\ Narrow NIR, Water vapour, Cirrus, SWIR1-2} & 10cm & -   \protect\\  

\global\let\CT@@do@color\relax  &  & \global\let\CT@@do@color\oriCT@@do@color 2021 & \cite{EarlyDet_Norway_CentEurope_Sentinel} & GAtt, RAtt & IPSEB & Sentinel-2 & MS &  \thead{RGB, B5-B7,\\ B8a, B11-B12} & 20m & Google Earth Engine  \protect\\  

\global\let\CT@@do@color\relax  &  & \global\let\CT@@do@color\oriCT@@do@color 2021 & \cite{EarlyDet_EuropeSpruce_NDRS} & GAtt & ESBB & Sentinel-1, Sentinel-2, Pleiades & MS &  \thead{Sen1: C-band; Sen2: RGB, Coastal aerosol, B5-7,\\ B8, B8a, B11-12} &  10/20/60m & SNAP   \protect\\  

 \cdashline{2-12}
\global\let\CT@@do@color\relax  & \global\let\CT@@do@color\relax \multirow{32}{*}{\rotatebox[origin=c]{90}{Late Detection}}  & \global\let\CT@@do@color\oriCT@@do@color 2003 & \cite{MPB_RedAttack_Strat_Landsat_BC} & RAtt & MPB & Landsat TM & MS &  B1-B5, B7 & 30m & -   \protect\\ 

\global\let\CT@@do@color\relax  &  & \global\let\CT@@do@color\oriCT@@do@color 2005 & \cite{RedAttack_MPB_HSR_Sat} & RAtt & MPB & IKONOS & MS & RGB, NIR, PAN & 4m & \thead{ImageStation Automatic\\ Triangulation}   \protect\\  

\global\let\CT@@do@color\relax  &  & \global\let\CT@@do@color\oriCT@@do@color 2006 & \cite{EstProb_MPB_RedAttack} & RAtt & MPB & Landsat-7 ETM+ & MS & B1-5, B7 & 30m & SPSS   \protect\\  

\global\let\CT@@do@color\relax  &  & \global\let\CT@@do@color\oriCT@@do@color 2006 & \cite{IntegRS_AncillaryData_CharMPB} & RAtt & MPB & Landsat-5 TM, Landsat-7 ETM+ & MS & B1-B7 &  30m & - \protect\\ %

\global\let\CT@@do@color\relax  &  & \global\let\CT@@do@color\oriCT@@do@color 2009 & \cite{Map_WhiteBark_MPB_Sat} & RAtt & MPB & QuickBird & MS & RGB, NIR & 2.4m & ENVI   \protect\\  

\global\let\CT@@do@color\relax  &  & \global\let\CT@@do@color\oriCT@@do@color 2010 & \cite{AssessCanopy_MPB_GeoEye_Sat} & RAtt, GRAtt & MPB & GeoEye-1 & MS &  RGB, NIR, PAN & 0.5m & ENVI   \protect\\  

\global\let\CT@@do@color\relax  &  & \global\let\CT@@do@color\oriCT@@do@color 2011 & \cite{EvalPoten_MultiSpect_MultipleStage} & RAtt, GRAtt & MPB & QuickBird & MS &  RGB, NIR & 2.4-24m & METRO  \protect\\

\global\let\CT@@do@color\relax  &  & \global\let\CT@@do@color\oriCT@@do@color 2013 & \cite{Eval_DetectBB_SingMultiData_Landsat} & RAtt & Unspecified & Landsat-5 TM, Landsat-7 ETM+ & MS &  \thead{RGB, B4-B5, B6,\\ B7, PAN} &  30m & Exelis Visual Information Solutions   \protect\\  

\global\let\CT@@do@color\relax  &  & \global\let\CT@@do@color\oriCT@@do@color 2013 & \cite{Quanti_MixedSpecies_MultiTempHighRes_Sat} & GRAtt & Unspecified & QuickBird, WorldView-2 & MS & WV2 \& QB: RGB, NIR, PAN & 0.6m & GENIE   \protect\\  

\global\let\CT@@do@color\relax  &  & \global\let\CT@@do@color\oriCT@@do@color 2014 & \cite{Spatial_Multidate_SPOT_Landsat}  & {RAtt, GRAtt} & ESBB & \thead{Landsat-5 TM, SPOT-2, \\ Landsat-7 ETM+, SPOT-4} & MS & \thead{Landsats: B1-B5, B7;\\ SPOTs: red, green, NIR} & 30/20m & -    \protect\\  

\global\let\CT@@do@color\relax &  & \global\let\CT@@do@color\oriCT@@do@color 2014 & \cite{ObjectExtract_LandsatSPOT} & RAtt, GRAtt & ESBB & \thead{Landsat-5 TM, SPOT-2, \\ Landsat-7 ETM+, SPOT-4} & MS & \thead{Landsats: B1-B5, B7;\\ SPOTs: B1-B3; PAN} & 30/20/15m & Quantum GIS, ENVI   \protect\\  

\global\let\CT@@do@color\relax  &  & \global\let\CT@@do@color\oriCT@@do@color 2014 & \cite{MPB_GrowthTrend_Landsat} & GRAtt & MPB & Landsat-5 & MS & B4, B7 & 30m & -   \protect\\  

\global\let\CT@@do@color\relax  &  & \global\let\CT@@do@color\oriCT@@do@color 2015 & \cite{Spruce_HighMedium_Remote} & GRAtt & SBB & Landsat TM & MS & B1-B7 &  30/120m & \thead{Ecosystem Disturbance \\ Adaptive Processing}   \protect\\  %

\global\let\CT@@do@color\relax  &  & \global\let\CT@@do@color\oriCT@@do@color 2016 & \cite{PercentMortality_MPB_Damage} & RAtt, GRAtt & MPB & Landsat-8 & MS & B1-B7, B10-B11 &  30/100m & -   \protect\\  %

\global\let\CT@@do@color\relax  &  & \global\let\CT@@do@color\oriCT@@do@color 2018 & \cite{Framework_Ecoregion_Spaceborne} & RAtt & DSBB & \thead{WorldView-2, WorldView-3, \\ Landsat-8 (OLI)} & MS & Landsat OLI: B1-B7, B9 & 1.5m & ENVI   \protect\\  

\global\let\CT@@do@color\relax  &  & \global\let\CT@@do@color\oriCT@@do@color 2018 & \cite{RapidEye_Detect_AreaSpruce}  & RAtt, GRAtt & ESBB & RapidEye, MODIS & MS & REye: RGB, NIR, Red-edge; MODIS: NIR, red &  5/250m & ATCOR-3  \protect\\ 

\global\let\CT@@do@color\relax  &  & \global\let\CT@@do@color\oriCT@@do@color 2019 & \cite{Comp_WorldViewLandsat_SVMNeuralNet} & RAtt, GRAtt & ESBB & WorldView-2, Landsat-8 (OLI) & MS & \thead{WV2: RGB, Coastal, Yellow, Red-edge, NIR1-2;\\ L8OLI: RGB, Coastal, NIR, SWIR 1-2} &  0.46/1.84/15/30m & ENVI   \protect\\  

\global\let\CT@@do@color\relax  &  & \global\let\CT@@do@color\oriCT@@do@color 2020 & \cite{GF2Sentinel2_Turpentine_China}  & {RAtt, GRAtt} & RTB & Sentinel-2, Gaofen-2 & MS & RGB, NIR, VEG1-VEG4, SWIR1-2, PAN &  1/4/10/20/30/60m & Agisoft, eCognition   \protect\\  

\global\let\CT@@do@color\relax  &  & \global\let\CT@@do@color\oriCT@@do@color 2021 & \cite{SubtleChange_Landsat_MPBSpruce} & RAtt, GRAtt & MPB, SBB & Landsat (ARD Tier 1) & MS & Red, NIR, SWIR1 \& SWIR2  & 30m & Google Earth Engine   \protect\\  

\global\let\CT@@do@color\relax  &  & \global\let\CT@@do@color\oriCT@@do@color 2022 & \cite{Northwenmost_Model_Climate} & GRAtt & ESBB & Landsat (time-series) & N/A & N/A & 30m & QGIS  \protect\\

\global\let\CT@@do@color\relax  &  & \global\let\CT@@do@color\oriCT@@do@color 2022 & \cite{Landsat_SouthPineBeet_Severity} & RAtt & SPB & Landsat-5 TM, Landsat-8 (OLI) & MS & \thead{L5TM: RGB, NIR, SWIR, TIR, MIR;\\ L8OLI: RGB, Coastal, NIR, B6-7, PAN, Cirrus} &  15/30/120m & Google Earth Engine, Google Earth Pro   \protect\\  %

\global\let\CT@@do@color\relax  &  & \global\let\CT@@do@color\oriCT@@do@color 2021 & \cite{DeadTree_PixObj_WorldView} & RAtt, GRAtt & ESBB & WorldView-3 & MS &  RGB, NIR1, NIR2, Red-edge, Yellow, Coastal, PAN &  2m/0.5m & \thead{eCognition, ENVI, \\ PCI Geomatics}    \protect\\  

\global\let\CT@@do@color\relax  &  & \global\let\CT@@do@color\oriCT@@do@color 2021 & \cite{DistSeg_Sat_DL} & RAtt, GRAtt & Unspecified & WorldView-2, Worldview-3 & MS & RGB, PAN & 0.3m & -  \protect\\ 

\hline \hline


\global\let\CT@@do@color\relax \multirow{18}{*}{\rotatebox[origin=c]{90}{Aircraft}} & \global\let\CT@@do@color\relax \multirow{9}{*}{\rotatebox[origin=c]{90}{Focus of Our Review}} & \global\let\CT@@do@color\oriCT@@do@color 2012 & \cite{AngularVegIndex_Spectroscopy_Bavarian} & GAtt, RAtt & ESBB & HyMap spectroscopy, CIR images, TCC & HS & 39 narrow bands (VNIR: 0.455-0.986 $\mu m$) &  0.4/7m & ATCOR4, ORTHO  \protect\\  %

\global\let\CT@@do@color\relax  &  & \global\let\CT@@do@color\oriCT@@do@color 2013 & \cite{Forecast_Vital_Hyperspect} & {GAtt, RAtt, GRAtt} & ESBB & HyMap spectroscopy, RGB imagery & HS & 125 bands (0.45-2.48 $\mu m$) & 4/7m & \thead{ATCOR4, ORTHO, ENVI, \\ RapidMiner, StereoAnalyst, LIBSVM}   \protect\\  %
 
\global\let\CT@@do@color\relax  &  & \global\let\CT@@do@color\oriCT@@do@color 2014 & \cite{Assess_Hyperspec_Mortality}  & {GAtt, RAtt, GRAtt} & ESBB & HyMap spectroscopy, CIR images & HS & 125 bands (0.45-2.48 $\mu m$) &  0.4/5/7m & ORTHO   \protect\\  %

\global\let\CT@@do@color\relax  &  & \global\let\CT@@do@color\oriCT@@do@color 2015 & \cite{HyperRS_MPB_Emphasis_Previsual} & GAtt, RAtt & MPB & SPECIM’s AISA Dual spectrometer, LiDAR & HS & VNIR (0.4–0.97 $\mu m$), SWIR (0.97-2.5 $\mu m$) & 0.25/2m & ATCOR 4   \protect\\  %

\global\let\CT@@do@color\relax  &  & \global\let\CT@@do@color\oriCT@@do@color 2021 & \cite{Map_Gyrocopter_HyperField}  & GAtt, RAtt & ESBB & \thead{Gyrocopter, HS HySpex VNIR 1600 camera, \\ Thermal camera, Nikon D800e} & HS/Thermal/RGB & RGB, 80 bands (0.4-1 $\mu m$) &  10/20/30cm & Py6S   \protect\\  

\global\let\CT@@do@color\relax  &  & \global\let\CT@@do@color\oriCT@@do@color 2022 & \cite{Comp_FieldRS_DetBB} & GAtt, RAtt & ESBB & CASI-1500 HS sensor & HS & 48 bands (0.38-1.05 $\mu m$) & 0.5m & -   \protect\\ 

 \cdashline{2-12}
\global\let\CT@@do@color\relax  & \global\let\CT@@do@color\relax \multirow{9}{*}{\rotatebox[origin=c]{90}{Late Detection}} & \global\let\CT@@do@color\oriCT@@do@color 2013 & \cite{LiveDead_Basal_LiDAR_Journal} & GRAtt & SBB, MPB & LiDAR & LiDAR & NIR & 30m & \thead{FUSION, Imagine add-on tool,\\ BCAL}   \protect\\  %
 
\global\let\CT@@do@color\relax  &  & \global\let\CT@@do@color\oriCT@@do@color 2015 & \cite{Detect_MPB_Ponderosa_HighResAerial} & RAtt & MPB & Cessna 441, Leica DO64 telecentric lens & RGB & RGB & 1m & -   \protect\\  %

\global\let\CT@@do@color\relax  &  & \global\let\CT@@do@color\oriCT@@do@color 2018 & \cite{UrbanForest_Hyperspec_UAV_Aircraft} & RAtt, GRAtt & ESBB & Cessna 172 OH-CAH, FPI HS camera; UAV & \thead{RGB/HS (24 bands)/\\Aerial CIR imagery} & 24 bands (Visible green to NIR (0.5–0.9 $\mu m$)) &  9/50cm & AgiSoft, LIBSVM   \protect\\  %
 
\global\let\CT@@do@color\relax  &  & \global\let\CT@@do@color\oriCT@@do@color 2019 & \cite{DeadWood_FCNDenseNet} & GRAtt & Unspecified & DMC aerial camera, CIR imagery & MS & NIR, Red, Green &  20cm & ArcGIS (Polygon annotation) \protect\\  

\global\let\CT@@do@color\relax  &  & \global\let\CT@@do@color\oriCT@@do@color 2020 & \cite{Habitat_Dynamics_NorwayDieback}  & GRAtt & ESBB & \thead{ALS (Full-waveform Riegl LMS-Q680i system, \\ Riegl VQ-1560i-DW system), CIR imagery \\ (UltraCam Eagle camera, DMC II 230 camera)} & MS/LiDAR & RGB, NIR &  0.2/0.5m & TerraSolid   \protect\\

\hline \hline


\multirow{27}{*}{\rotatebox[origin=c]{90}{Drone}} & \global\let\CT@@do@color\relax \multirow{15}{*}{\rotatebox[origin=c]{90}{Focus of Our Review}} & \global\let\CT@@do@color\oriCT@@do@color 2019 & \cite{UAV_PrecDetection_BBInf}  & {GAtt, RAtt, GRAtt} & ESBB & \thead{Zefyros Oktos XL, Sony Alpha A7 camera (RGB),\\ Lumix TZ7 (CIR: CMOS-based full-frame camera)} & RGB/MS & RGB, NIR &  2.3cm & \thead{Agisoft, ENVI,\\ ArcGIS, STATISTICA}  \protect\\ %

\global\let\CT@@do@color\relax  &  & \global\let\CT@@do@color\oriCT@@do@color 2019 & \cite{PineShootBeetle_Hyperspect_LiDAR} & {GAtt, RAtt, GRAtt} & PSB & \thead{HS imagery (pushbroom Nano-Hyperspec), \\ LiDAR (LiAir 200, 40-channel Pandar40 laser sensor)} & HS/LiDAR & 270 bands (VNIR (0.4–0.1 $\mu m$)) & 0.2m & Headwall’s SpectralView   \protect\\  %

\global\let\CT@@do@color\relax  &  & \global\let\CT@@do@color\oriCT@@do@color 2020 & \cite{MultiTemp_HyperMultispect_NorwaySpruce} & GAtt & ESBB & \thead{Quadcopter drone, 2 oblique RGB cameras (Sony A7R II);\\ HS camera (Rikola), MS camera (Micasense Altum camera)} & RGB/HS/MS & RGB, Red-edge, NIR, 46 bands (0.502-0.907 $\mu m$) & 2.1/8.6/6/90cm & \thead{Agisoft Metashape, FGI,\\ RadBA, FUSION}  \protect\\  %

\global\let\CT@@do@color\relax  &  & \global\let\CT@@do@color\oriCT@@do@color 2020 & \cite{CrownExt_UASMultispec_MixedForest} & GAtt, YAtt & ESBB & DJI Matrice 210, MicaSense RedEdge-M camera & MS & RGB, Red-edge (0.717 $\mu m$), NIR (0.84 $\mu m$) &  6.8cm & Agisoft Metashape, QGIS   \protect\\  %

\global\let\CT@@do@color\relax  &  & \global\let\CT@@do@color\oriCT@@do@color 2021 & \cite{UAS_Multispect_DL} & GAtt, YAtt & Unspecified & \thead{DJI Matrice 210 RTK, \\ MicaSense RedEdge-M MS camera} & MS & RGB, Red-edge (0.717 $\mu m$), NIR (0.84 $\mu m$) &  6cm & Agisoft   \protect\\  %

\global\let\CT@@do@color\relax  &  & \global\let\CT@@do@color\oriCT@@do@color 2021 & \cite{Mapping_Boreal_HealthStatus}  & \thead{GAtt, RAtt, GRAtt} & SBB & \thead{DJI Phantom 4 Pro, DJI Matrice 210 (MicaSense \\RedEdge), Goddard Lidar Hyperspectral Thermal Airborne} & \thead{RGB/MS/\\HS/LiDAR} & RGB, VNIR, SWIR (0.4-1 $\mu m$) & \thead{2-3cm, 6cm, \\0.5m, 0.1-0.2m} & Pix4D, ENVI, ArcGIS   \protect\\  

\global\let\CT@@do@color\relax  &  & \global\let\CT@@do@color\oriCT@@do@color 2022 & \cite{Detection_UAV_YOLOs} & GAtt, YAtt, RAtt, GRAtt & ESBB & DJI Phantom 4 Pro, RGB camera & RGB & RGB & 3.75cm & QGIS  \protect\\  

\cdashline{2-12}
\global\let\CT@@do@color\relax  & \global\let\CT@@do@color\relax \multirow{13}{*}{\rotatebox[origin=c]{90}{Late Detection}} & \global\let\CT@@do@color\oriCT@@do@color 2014 & \cite{Toward_UAV_ForestMonitor} & RAtt, GRAtt & ESBB & \thead{RGB camera; Sports airplane \\(Narrow band camera TetraCam Mini MCA 6)} & RGB/MS & \thead{RGB/ B1 (0.55$\pm$0.05 $\mu m$), B2 (0.67$\pm$0.05 $\mu m$), B3 (0.71$\pm$0.05 $\mu m$),\\ B4 (0.78$\pm$0.05 $\mu m$), B5 (0.9$\pm$0.01 $\mu m$), B6 (0.95$\pm$0.02 $\mu m$)} &  6cm & N/A  \protect\\  

\global\let\CT@@do@color\relax  &  & \global\let\CT@@do@color\oriCT@@do@color 2015 & \cite{Photogram_Hyperspectral_TreeLevel}  & YAtt, GRAtt & ESBB & \thead{Octocopter, Ordinary Samsung \\ NX1000 RGB camera, FPI HS camera} & RGB/HS & RGB/22 Bands (0.5-0.9 $\mu m$) &  2.4/9cm & N/A  \protect\\  %

\global\let\CT@@do@color\relax  &  & \global\let\CT@@do@color\oriCT@@do@color 2019 & \cite{BB_Fir_DL} & RAtt, GRAtt & FFBB & \thead{DJI Phantom 3 Pro quadcopter (built-in camera), \\ Yeneec Typhoon H hexacopter (CGO3+ camera)} & RGB & RGB &  5-10cm & QGIS, Agisoft  \protect\\  %

\global\let\CT@@do@color\relax  &  & \global\let\CT@@do@color\oriCT@@do@color 2020 & \cite{Class_HealthStat_UAS_Multispect} & RAtt, GRAtt & Unspecified & DJI S900, MicaSense RedEdge-M camera & MS & RGB, NIR (0.84 $\mu m$), Red-edge (0.717 $\mu m$) &  5cm & \thead{STATISTICA, Agisoft, ArcMap,\\ PCI geomatica, SPSS Statistics}   \protect\\  %

\global\let\CT@@do@color\relax  &  & \global\let\CT@@do@color\oriCT@@do@color 2021 & \cite{DetectBB_Threshold_CellAutomata}  & YAtt, RAtt, GRAtt & MEXPB & \thead{Hexacopter with a Tarot FY680 Pro,\\ Parrot Sequoia MS sensor} & RGB & RGB & 3.8m & OpenDroneMap, ImageMagick   \protect\\  

\global\let\CT@@do@color\relax  &  & \global\let\CT@@do@color\oriCT@@do@color 2021 & \cite{SickFir_UAV_DL} & GRAtt & SFBB & \thead{DJI Mavic 2 pro Hasselblad L1D-20c camera, \\ DJI Phantom 4 Quadcopter} & RGB & RGB & 2.6cm & GS Pro, GIMP, Agisoft, Global Mapper   \protect\\  %

\global\let\CT@@do@color\relax  &  & \global\let\CT@@do@color\oriCT@@do@color 2021 & \cite{CrosInteract_Mortality} & GRAtt & WPB & \thead{DJI Matrice 100, DJI Zenmuse X3 RGB \\ camera, Micasense Rededge3 MS camera} & RGB/MS & RGB, 5 narrow bands &  5/8cm & Pix4Dmapper Cloud, QGIS   \protect\\  

\global\let\CT@@do@color\relax  &  & \global\let\CT@@do@color\oriCT@@do@color 2022 & \cite{Multispectral_Benefits_Spruce_Decline} & YAtt, RAtt, GRAtt & ESBB & \thead{Custom drone (Sony A7R digital camera), Phantom 4\\ Pro V2 (Micasense RedEdge M), LiDAR} & RGB/MS/LiDAR & RGB, NIR, Red-edge & 5/8cm & Metashape  \protect\\  
\hline
  \end{tabular}
  } 
\end{table*}
\begin{table*}[!bp]
\vspace{-0.3cm}
\rowcolors{2}{}{gray!10}
\caption{Comparison of satellite platforms and their spectral bands used for detecting bark beetle attacks ($\mu m$).} 
\vspace{-0.3cm}
\centering
 \resizebox{\textwidth}{!}{
\begin{tabular}{c c c c c c c c c c c c c}
\hline
& Satellite & PAN & Blue & Green & Red & Red-edge & NIR & SWIR & TIR & \global\let\CT@@do@color\relax\thead{Resolution ($m$) \\ (MS, PAN, TIR)} & Revisit time (days) & \thead{Data\\ Availability} \\
\hline\hline

\global\let\CT@@do@color\relax \multirow{18}{*}{\rotatebox[origin=c]{90}{Focus of Our Review}} & \global\let\CT@@do@color\oriCT@@do@color Landsat-5 & - & - & $0.5-0.6$ (B1)
 & $0.6-0.7$ (B2) & - & 
\thead{$0.7-0.8$ (B3) \\
       $0.8-1.1$ (B4)} & - & -  & $(60, -, -)$  & $16$ & Open Data \\

\global\let\CT@@do@color\relax   & \global\let\CT@@do@color\oriCT@@do@color Landsat-5 TM & - & $0.45-0.52$ (B1) & $0.52-0.6$ (B2) & $0.63-0.69$ (B3) & - & $0.76-0.9$ (B4) & 
\thead{SWIR-1: $1.55-1.75$ (B5) \\
       SWIR-2: $2.08-2.35$ (B7)} &  $10.4-12.5$ (B6) & $(30, - ,120)$ & $16$ & Open Data \\

\global\let\CT@@do@color\relax   & \global\let\CT@@do@color\oriCT@@do@color Landsat-7 ETM+ & $0.52-0.9$ (B8) & $0.45-0.52$ (B1) & $0.52-0.60$ (B2) & $0.63-0.69$ (B3) & - & $0.77-0.9$ (B4) & 
\thead{SWIR-1: $1.55-1.75$ (B5) \\
SWIR-2: $2.09-2.35$ (B7)} & $10.4-12.5$ (B6) & $(30, 15, 60)$ & $16$ & Open Data \\

\global\let\CT@@do@color\relax   & \global\let\CT@@do@color\oriCT@@do@color Landsat-8 OLI \& TIRS & $0.5-0.68$ (B8)& $0.45-0.51$ (B2) & $0.53-0.59$ (B3) & $0.64-0.67$ (B4) & - & $0.85-0.88$ (B5) & 
\thead{Cirrus: $1.36-1.38$ (B9) \\
       SWIR-1: $1.57-1.67$ (B6) \\
       SWIR-2: $2.11-2.29$ (B7) }
& \thead{TIR-1: $10.6-11.19$ (B10)\\
TIR-2: $11.5-12.51$ (B11)}  & $(30, 15, 100)$  & $16$ & Open Data \\

\global\let\CT@@do@color\relax   & \global\let\CT@@do@color\oriCT@@do@color Sentinel-2 & - & $0.46-0.52$ (B2) & $0.54-0.58$ (B3) & $0.65-0.68$ (B4) & 
\thead{Red-edge-1: $0.698-0.712 (B5)$ \\                                                                                   Red-edge-2: $0.733-0.747 (B6)$ \\
       Red-edge-3: $0.773-0.795 (B7)$}
& 
\thead{$0.784-0.9$ (B8) \\  
$0.855-0.875$ (B8A)} & 
\thead{SWIR-1: $1.565-1.655$ (B11)  \\
       SWIR-2: $2.1-2.28$ (B12)}
& - & \thead{B2-B4, B8: $10$ \\
             B5-B7, B8A, B11-B12: $20$ \\
             B1, B10: $60$
}
 & $5$ & Open Data \\

\global\let\CT@@do@color\relax   & \global\let\CT@@do@color\oriCT@@do@color WorldView-2 & $0.45-0.8$ & \thead{Coastal Blue: $0.4-0.45$ \\ $0.45-0.51$} & \thead{$0.51-0.58$ \\ Yellow: $0.585-0.625$} & $0.63-0.69$ & $0.705-0.745$ & 
\thead{NIR-1: $0.77-0.895$ \\
       NIR-2: $0.86-1.04$} & - &  - & $(1.84,0.5,-)$  & $1.1$ & Commercial \\

\global\let\CT@@do@color\relax   & \global\let\CT@@do@color\oriCT@@do@color SPOT-2 & $0.51-0.73$ & - & $0.5-0.59$ & $0.61-0.68$ & - & $0.79-0.89$ & - & - &  $(10, 20, -) $ & $26$ & Commercial \\

\global\let\CT@@do@color\relax   & \global\let\CT@@do@color\oriCT@@do@color SPOT-4 & $0.61-0.68$ & - & $0.5-0.59$ & $0.61-0.68$ & - & $0.79-0.89$ & $1.53-1.75$ & - & $(10, 20, -)$  & $26$ & Commercial \\

\global\let\CT@@do@color\relax   & \global\let\CT@@do@color\oriCT@@do@color Gaofen-2 & $0.45-0.89$ & $0.45-0.52$ (B1) & $0.52-0.59$ (B2) & $0.62-0.69$ (B3) & - & $0.77-089$ (B4) & - & - & $(3.2 ,0.8, -)$  & $5$ & Commercial \\

\global\let\CT@@do@color\relax   & \global\let\CT@@do@color\oriCT@@do@color RapidEye  & - & $0.44-0.51$ & $0.52-0.590$ & $0.63-0.685$ & $0.69-0.73$ & $0.76-0.85$ & - & - & $(6.5, -, -)$ & $5.5$ & Commercial \\

\global\let\CT@@do@color\relax   & \global\let\CT@@do@color\oriCT@@do@color Pleiades & $0.48-0.83$ & $0.43-0.55$ & $0.49-0.61$ & $0.6-0.72$ & - & $0.75-0.95$ & - & - & $(2, 0.5,-)$ & $1$ & Commercial \\


\hdashline

\global\let\CT@@do@color\relax \multirow{8}{*}{\rotatebox[origin=c]{90}{Late Detection}} & \global\let\CT@@do@color\oriCT@@do@color WorldView-3  & $0.45-0.8$  & \thead{Coastal Blue: $0.4-0.45$ \\ $0.45-0.51$} & \thead{$0.51-0.58$ \\ Yellow: $0.585-0.625$} & $0.63-0.69$ & $0.705-0.745$ & 
\thead{NIR-1: $0.77-0.895$  \\
      NIR-2: $0.86-1.04$} & 
\thead{SWIR-1: $1.195-1.225$ \\
SWIR-2: $1.55-1.59$ \\
SWIR-3: $1.64-1.68$ \\
SWIR-4: $1.71-1.75$ \\
SWIR-5: $2.145-2.185$ \\
SWIR-6: $2.185-2.225$ \\
SWIR-7: $2.235-2.285$ \\
SWIR-8: $2.295-2.365$}
& - & $(1.24, 0.31, 3.7)$ & $4.5$ & Commercial \\

\global\let\CT@@do@color\relax   & \global\let\CT@@do@color\oriCT@@do@color QuickBird  & $0.45-0.9$  & $0.45-0.52$  & $0.52-0.6$  & $0.63-0.69$ & - & $0.76-0.9$ & - & - & $(2.62,0.65,-)$  & $2.8$  & Open Data \\

\global\let\CT@@do@color\relax   & \global\let\CT@@do@color\oriCT@@do@color MODIS Terra \& Aqua & - & \multicolumn{3}{c}{12 bands (B1, B3-B4, B8-B14)} &  $0.743-0.753$ (B15) & \thead{7 bands \\ (B2, B5, B16-B19, B26)} & 
\thead{$1.628-1.652$ (B6)\\
        $2.105-2.155$ (B7)}
 & \thead{16 bands \\ (B20-B25, B27-B36)} &  
 \thead{B1-B2: $250$ \\
        B3-B7: $500$ \\
        B8-B36: $1000$}
  & $1-2$ & Open Data \\

\global\let\CT@@do@color\relax   & \global\let\CT@@do@color\oriCT@@do@color IKONOS & $0.45-0.9$ & $0.445-0.516$ (B1) & $0.506-0.595$ (B2) & $0.632-0.698$ (B3) & - & $0.757-0.853$ (B4) & - & - & $(4, 0.8, -)$ & $1-3$ & Commercial \\

\global\let\CT@@do@color\relax   & \global\let\CT@@do@color\oriCT@@do@color GeoEye-1 & $0.45-0.8$ & $0.45-0.51$ & $0.51-0.58$ & $0.655-0.69$ & - & $0.78-0.92$ & - & - & $(1.65, 0.41, -)$ & $4.6$ & Commercial \\

\hline
\end{tabular}

  }
\vspace{-0.3cm}  
\label{Table_RS_Sat_Comp}  
\end{table*}

Three types of RS platforms, namely satellites, manned aircraft, and drones have been utilized to detect bark beetle attacks. Characteristics of these platforms are compared in Table~\ref{Table_RS_All} (for more general details, see \cite{RS_review_platform_sensor}). Optical sensors mounted on satellites offer decametric to submetric resolution and cover broad areas (see Table~\ref{Table_RS_Sat_Comp}), but are limited by revisit times and clouds. Aircraft on the other hand can carry a variety of sensors, some of which (e.g., LiDAR) can operate in overcast conditions providing the cloud ceiling is high enough, and provide decimetric to centimetric resolution. More recently, drones with high flexibility in applications and sensors that provide centimetric to millimetric resolution are being used, but with current regulations, they can only cover small areas.

\subsection{Satellite} \label{sec:RS_sat}
Early detection of bark beetle attacks has been conducted using a variety of commercial or non-commercial satellites and sensors. Table~\ref{Table_RS_All} provides a comparative overview of the use of satellites to detect bark beetle attacks. Following is a summary of their effectiveness for early detection of the attacks.
In \cite{EarlyDet_TerraSarX_RapidEye}, RapidEye (optical) and TerraSAR-X (radar) satellite data were used independently and together to detect GAtt plots (i.e., circles including three or more GAtt trees). It was found that combining radar and optical data enhanced performance, with radar data playing a complementary role, while TerraSAR-X results were less accurate than RapidEye ones.
In another work \cite{SpectEvid_Early_Engelmann}, a Landsat spectral band (i.e., band {7} (SWIR-{2}) and associated SVI (i.e., RGI) correlate to a ground-based spectroradiometer were identified as promising for early bark beetle outbreak detection. This study employed ground information (narrow bands at the branch \& needle levels) to scale up to the entire landscape using satellite data by detecting changes at the forest stand level.
In \cite{Sens_LandSatOLI_TIRS_Foliar}, Landsat-{8} images from optical \& thermal sensors were used to differentiate between healthy and infested samples using SVIs and \textit{canopy surface temperature} (CST). Thermal data offered the potential to reveal plants' physiological and biochemical properties and detect their disease before it becomes apparent through symptoms of visual stress. 
The potential of commercial WorldView-{2} imagery (higher spatial \& spectral resolutions than, for instance, Landsat-{5}) was examined in \cite{EarlyDet_NorwaySpruce_WorldView2,MPB_BlackHills_SouthDakota} for the GAtt detection. 
In \cite{MPB_BlackHills_SouthDakota}, pan-sharpened WorldView-{2} images (i.e., fusing MS images with a panchromatic image) were used to extract seemingly green tree crowns (including GAtt \& healthy trees). While slightly different spectral responses were observed for GAtt trees than healthy ones, small effect levels of spectral differences and substantial spectral overlaps of these trees led to low GAtt detection accuracy. 
These findings were also observed for Ponderosa pines when just subtle differences in spectral response were detected by using MS bands and their SVIs due to significant overlaps and variances \cite{EarlyDet_NorwaySpruce_WorldView2}. 

The SVIs sensitive to leaf \& canopy spectral variations from Landsat-{8} and Sentinel-{2} were used in \cite{Sentinel2_GA_Landsat8} for mapping GAtt plots. As expected, the higher spatial and spectral resolutions of Sentinel-{2} imagery resulted in higher sensitivity for its SVIs. 
In \cite{Monitor_CentrEuro_Validate}, the short revisit period of Sentinel-{2} enabled mapping of bark beetle damage and severity in forest areas based on a change detection approach. However, detecting no/minor/moderate damage areas was limited due to overlap in vitality values and subtle signals. 
Dense time-series of Sentinel-{2} images were used in \cite{EarlyDet_Norway_CentEurope_Sentinel} to quantify the impact of spectral bands and their temporal separability between healthy and bark beetle-infested forest stands within seasonal changes of canopy reflectance. Although clouds frequently covered mountainous areas (especially in spring and autumn) resulting in notable gaps in the time-series analysis, analyzing changes in spectral bands' seasonal trajectory was more critical than merely assessing static maps/values.
Lastly, differences between the time series of radar data (Sentinel-{1}) and optical images (Sentinel-{2}) during the whole vegetation season were investigated in \cite{EarlyDet_EuropeSpruce_NDRS} to distinguish stressed pixels due to the bark beetle attacks. It indicated that radar information was less important than the MS and found that forest vulnerability before beetle attacks (i.e., spectral differences caused by prior weaknesses versus stress caused by beetle attack) should be considered in analyses.
\subsection{Aircraft} \label{sec:RS_aircraft}
The manned aircraft platform served to fill the gaps of using various imaging sensors, providing a flexible imaging time, and alleviating discrepancies (i.e., mixed information due to coarse spatial resolution) compared to non-commercial satellites. As shown in Table~\ref{Table_RS_All}, most related studies utilized HS sensors to record a set of contiguous narrow bands that could target specific ranges of the spectrum. As a consequence of the availability of many bands, these sensors assessed particular features (e.g., physiological structures) closely linked to the response of  {bark beetle attacked trees} (see Fig.~\ref{fig:beetles}) and provided a more detailed shape of spectral signatures for further analyses. 
 {For instance, the HyMap sensor was used in \mbox{\cite{AngularVegIndex_Spectroscopy_Bavarian,Forecast_Vital_Hyperspect,Assess_Hyperspec_Mortality}}. It provided {{125}} spectral bands with a spectral range from \mbox{{0.45}-{2.48} $\mu m$} with resolutions between \mbox{{13}-{17} $\mu m$} to identify relevant signals required to detect bark beetle damages, while airborne CIR images were used as the reference data in the absence of field data. 
Using this sensor, angular indices (i.e., the combination of wavelengths and reflectance values of three bands) were calculated in \mbox{\cite{AngularVegIndex_Spectroscopy_Bavarian}} to identify the most suitable spectral information and develop forest damage sensitivity indices. Despite the potential of these indices to detect medium \& heavy beetle damages, the study could not distinguish low-damaged trees from healthy trees due to spectral similarities. 
Moreover, various mortality stages of forest trees were mapped using original \& normalized HS bands in  \mbox{\cite{Assess_Hyperspec_Mortality}}.
The most useful spectral information to predict vitality stages of tree stands from this sensor was also investigated in \mbox{\cite{Forecast_Vital_Hyperspect}} from changes in biochemical-biophysical characteristics in response to vegetation stress levels. It was assumed that bark beetle infestation has a secondary role (following, e.g., climatic change) in reducing vegetation vitality. However, substantial spectral overlaps between GAtt and non-attacked trees resulted in poor separation due to coarse spatial resolution (i.e., mixed information) and likely the influence of other factors (e.g., stand age and condition).}

On the other hand, the changes in leaf/canopy reflectance were analyzed by continuum removal (i.e., normalization of specific features for comparison with a variety of targets or acquisitions) with shape-related metrics \cite{HyperRS_MPB_Emphasis_Previsual}. In this work, LiDAR data with a fine spatial resolution served as a guide for sampling HS data from treetop locations, decreasing the impact of crown shape and position on brightness to focus on tree health status. However, subtle reflectance differences were identified for GAtt detection in a small stand of infested trees.  
As the previous studies could not detect the GAtt phase due to their low spatial resolution-based analysis, the HS data in \cite{Map_Gyrocopter_HyperField} almost simultaneously acquired by a gyrocopter (airborne data) and a field spectrometer (ground data) to analyze airborne \& laboratory indices with different spatial scales. However, needle-level data acquired from the ground could not effectively be scaled to airborne pixel-level data and did not differentiate healthy trees from GAtt trees due to external factors (e.g., shadowing and under-canopy vegetation). Further, the lack of quality and quantity of reference data (e.g., limited samples and unfavorable data acquisition conditions) prevented definitive conclusions. 
Finally, the limitations of RS methods to assess the sequence (i.e., transition across infestation stages) and timing of attack symptoms were studied in \cite{Comp_FieldRS_DetBB} by comparing aerial HS images and classical field surveys. To this end, weekly field surveys and monthly aerial scans (i.e., HS time series) were conducted during the same period in which the spatial resolution of {0.5} $m$ enabled monitoring of early infestation stages in young trees (e.g., {30}-{40} years). However, airborne HS data were found to take longer to detect infested trees than field surveys as a result of the different frequencies of data acquisition and the inability to recognize subtle changes in tree conditions.
\subsection{Drone} \label{sec:RS_uav}
Drones allow end-users to customize acquisition settings (e.g., sensor type, view angle, spatial resolution, and acquisition time \& interval) for small-scale forest monitoring \cite{UAV_Sat_Review}. Table~\ref{Table_RS_All} presents the drone-acquired data (mostly from MS sensors) for monitoring bark beetle attacks. 
In \cite{UAV_PrecDetection_BBInf}, a drone mounted with RGB and NIR cameras was used to examine the spectral differences between healthy and GAtt trees based on SVIs. While RGB bands were partially effective (\& limited success using the NIR band) and the trees differed in spectral responses for the images captured at the beginning (June), during (August), and after (October) the outbreak, further investigation with more professional MS/HS cameras was recommended. 
Another study \cite{UAS_Multispect_DL} explored drone-captured VHR images to address the issue of low resolution in MS images. However, insufficient differences between GAtt, YAtt, and healthy trees led to the merging of GAtt and YAtt classes for further analyses of this study. 
In \cite{Detection_UAV_YOLOs}, drone-collected RGB images were used to detect early attacked trees. It was indicated that enhancing image contrast allowed for more accurate analysis and reduced false positives/negatives as the quality of images can be degraded by various factors (e.g., equipment quality).
Drone-captured MS images were utilized in \cite{CrownExt_UASMultispec_MixedForest} to develop an individual tree crown delineation and consequently detect beetle disturbances in a mixed urban forest. Although spectral information was combined with four tree crown elevation metrics to address the similar responses of spectral bands and SVIs for various species/infestation stages, no specific signal could differentiate between GAtt \& YAtt stages and tree species of pine \& spruce. 
In \cite{Multispectral_Benefits_Spruce_Decline}, RGB and MS images were captured from four forest areas in spring and fall to examine the effects of phenology, data acquisition time, and life cycle stages of bark beetles on tree vitality. Accordingly, the data collection in the fall or the end of summer yielded better results due to the lower deviations and variability of spectral responses.

A comparison of drone-based HS and LiDAR data was conducted in \cite{PineShootBeetle_Hyperspect_LiDAR} to quantify tree crown damage severity based on the predictive power of each and their fusion. Evidently, the HS data including {270} bands were more sensitive to tree crown damage than the LiDAR metrics from a single NIR band. While combining the advantages of these data improved the prediction accuracy of crown damage, the error sources that prevented accurate estimation were identified as uncertainties \& overestimation of HS features and exclusion of shaded pixels containing damage information. 
In \cite{MultiTemp_HyperMultispect_NorwaySpruce}, the RGB, MS, and HS (from a \textit{Fabry–P\'erot interferometer} (FPI) camera with lightweight \& low-cost HS sensors) images were collected by a quadcopter drone for spectral analysis of crown symptoms. The HS data achieved slightly better results than the MS data in the highly challenging task of detecting trees suffering from GAtt or root-rot. 
Lastly, the fusion of MS/HS data with structural information using either drone or \textit{airborne Goddard LiDAR HS thermal} (G-LiHT) was compared in \cite{Mapping_Boreal_HealthStatus} to discriminate the health status of individual trees. Although data fusion and structural or {3}D spectral information led to improved performance, the analyses from structural \& RGB information and MS images indicated that spectral-only data could not distinguish healthy from GAtt trees.
%
\section{ {Machine Learning Perspective}} \label{sec:ML}
\textit{Machine Learning} (ML) is the principal component of artificial intelligence systems that automatically learns from data/experiences and makes informed decisions about future data. It can roughly be categorized into classical- and DL-based methods. 
 {Classical methods (e.g., \textit{random forest} (RF) or \textit{support vector machines} (SVM)) refer to traditional ML algorithms that require an in-depth understanding and sufficient domain expertise of the problem. These methods rely on manually selecting handcrafted features and require mathematical models to analyze data, but they may not generalize well to new or complex data.
In contrast, DL-based methods (e.g., \textit{convolutional neural networks} (CNNs) and \textit{convolutional auto-encoders} (CAEs)) involve training DNNs to automatically discover data structure and learn relevant features from the raw input data, without the need for explicit feature engineering. It is worth noting that these methods require large amounts of training data compared to classical methods due to their reliance on data-driven discovery rather than domain knowledge, which poses a critical challenge in detecting bark beetle attacks}.
As shown in Table~\ref{tab_ML_all}, a wide range of these methods was used with various inputs, features, and different numbers of samples for training, validation, and testing procedures. 
These models are either parametric or non-parametric, where a fixed number of parameters defines parametric models, and are independent of the training data size. However, non-parametric models are more likely to learn non-linear data distributions, but their complexity increases as training data and model parameters are increased. The categorization of classical ML methods includes the algorithms, types of used handcrafted features, and exploited information (pixel- or region-based), while the DL-based methods are generally compared in terms of their networks and architectures. The general comparison of the most commonly used methods for this purpose is summarized in Table~\ref{tab_ML_comp}. From the task category aspect, bark beetle attacks are modeled as classification (binary- or multi-class), segmentation (traditional image-processing based, or semantic), detection, regression, or clustering tasks. Further, supervised and unsupervised learning strategies are adopted to train the models. 
 {Despite the lack of established ML milestones specific to bark beetle attack detection, advancements in ML/DL techniques have significantly improved the detection of bark beetle infestations, including feature engineering, data fusion and integration, time-series analysis, change detection algorithms, ensemble methods, unsupervised learning, transfer learning, and various DNNs.
The following sections describe these methods for GAtt detection and summarize their overall performance \& evaluation metrics, while quantitative comparisons are presented in \mbox{Sec.~\ref{sec:Evals}}.}
\subsection{ {Classical ML Algorithms used for Multispectral Analysis}} \label{sec:ML_classic_MS}
As the most widely used algorithm for bark beetle attack detection, the RF ensembles multiple decision trees and randomly selects training samples and variables. The RF algorithm in various objectives helped the methods for detecting the early stages of bark beetle infestation, e.g., the forest/non-forest pixels classification \cite{Monitor_CentrEuro_Validate}, tree species classification \cite{Monitor_CentrEuro_Validate}, feature importance evaluation \cite{EarlyDet_Norway_CentEurope_Sentinel}, and forest disturbance/tree vitality classification \cite{EarlyDet_Norway_CentEurope_Sentinel,EarlyDet_EuropeSpruce_NDRS,Multispectral_Benefits_Spruce_Decline}. 
In \cite{Monitor_CentrEuro_Validate}, the pixels that belonged to non-forest areas were discarded and the species that were prone to the attacks at a forest stand were classified by the RF algorithm. Then, a bi-temporal ordinary least squares regression and score-based rules determined the deviations of disturbed areas from healthy ones (i.e., anomaly detection) as well as the classification of vitality changes. Having confusion between GAtt and no damage classes, this method disregards minor damage classes that could be attributed to either forest vitality or bark beetle damage. 
Another work \cite{EarlyDet_Norway_CentEurope_Sentinel} used the RF classifier to evaluate the most sensitive spectral bands and SVIs of Sentinel-{2} images with the aim of developing an early forest disturbance model. However, several factors (e.g., windfallen trees or small clear-cuts) led to incorrectly classified forest stands in addition to confused spectral responses for analysis of areas at a smaller scale than Sentinel-{2} pixel size. In addition, this study was conducted on monoculture stands which means the performance for early infestation detection could be less for multi-species forests. 
The RF classifier was employed in \cite{EarlyDet_EuropeSpruce_NDRS} to quantify the separability of healthy and stressed trees at early-stage of infestation from Sentinel-{1} and Sentinel-{2} data. Although three RF models with different combinations of radar and optical bands were evaluated to classify stressed tree crowns, the spectral differences were not sufficient to detect GAtt symptoms. 
Lastly, the RF algorithm was applied for the vitality classification of trees and model transferability investigation to other areas \& areas with similar spectral responses \cite{Multispectral_Benefits_Spruce_Decline}. It trained two independent RF classifiers on collected data in May \& September, resulting in low/varied accuracy for declined tree predictions. However, applying this method to different areas led to instability in results.
\begin{table*}[!tbp]
\rowcolors{2}{}{gray!10}
\caption{Comparison of ML and DL-based methods for detecting bark beetle attacks [Abbreviations are denoted in Table~\ref{tab_Abbrev}].} 
\vspace{-0.3cm}
\centering 
 \resizebox{\textwidth}{!}{
\begin{tabular}{c c c c c c c c c c c c c c c} 
\hline
 &  & Year & Ref. & Attack stage & Bark beetle & Approach & Learning & Category & Model & Algorithm / Network & Classes / Clusters & Features & Information/ Network Architecture  \protect\\ 
\hline\hline
\global\let\CT@@do@color\relax \multirow{58}{*}{\rotatebox[origin=c]{90}{Classical Machine Learning}} & \global\let\CT@@do@color\relax \multirow{32}{*}{\rotatebox[origin=c]{90}{Focus of Our Review}} & \global\let\CT@@do@color\oriCT@@do@color 2012 & \cite{AngularVegIndex_Spectroscopy_Bavarian} & GAtt, RAtt & ESBB & ML & SL & CLAS & NPAR & SVM & \thead{Heavy/medium/green damages, healthy coniferous,\\ healthy broadleaved, bare soil} & Angle/vegetation indices & Pixel-based \protect\\  

\global\let\CT@@do@color\relax  &   & \global\let\CT@@do@color\oriCT@@do@color 2013 & \cite{EarlyDet_TerraSarX_RapidEye}  & GAtt & ESBB & ML & SL & CLAS & NPAR/ PAR & RF/ MAXE, GLM & GAtt, healthy  & \thead{Std, max, min, mean, median, the first \& \\ third quartiles of the backscatter distribution} & Pixel-based    \protect\\ 

\global\let\CT@@do@color\relax  &   & \global\let\CT@@do@color\oriCT@@do@color 2013 & \cite{Forecast_Vital_Hyperspect}  & GAtt, RAtt, GRAtt & ESBB & ML & SL & CLAS & NPAR/ PAR & DT (ID3) & \thead{(grey–green), (red–grey), (infestation 2010),\\ (infestation 2011), (healthy, green)} & Spectral information/derivatives/indices & Pixel-based  \protect\\  

\global\let\CT@@do@color\relax  &   & \global\let\CT@@do@color\oriCT@@do@color 2014 & \cite{Assess_Hyperspec_Mortality}  & GAtt, RAtt, GRAtt & ESBB & ML & USL/ SL & CLAS & NPAR & GA/ SVM & \thead{Healthy coniferous, healthy broad-leaved, \\bare soil,  sparsely vegetated soil} & Spectral bands & Pixel-based   \protect\\   


\global\let\CT@@do@color\relax  &   & \global\let\CT@@do@color\oriCT@@do@color 2014 & \cite{EarlyDet_NorwaySpruce_WorldView2} & GAtt, RAtt, GRAtt & ESBB, SSBB & ML & SL & CLAS & NPAR/ PAR & RF/ LOGR & Dead (RAtt, GRAtt), GAtt, healthy (non-attacked) & Spectral signatures & Pixel-based \protect\\ 

\global\let\CT@@do@color\relax  &   & \global\let\CT@@do@color\oriCT@@do@color 2015 & \cite{HyperRS_MPB_Emphasis_Previsual} & GAtt, RAtt & MPB & ML & SL & CLAS & NPAR & SAM, NN &  Healthy, previsual GAtt, RAtt & Pigment \& water absorption features & Pixel-based   \protect\\  

\global\let\CT@@do@color\relax  &   & \global\let\CT@@do@color\oriCT@@do@color 2017 & \cite{SpectEvid_Early_Engelmann}  & GAtt & SBB & ML & SL & CLAS & NPAR & RF & Infested/non-infested trees & Different wavelengths & -    \protect\\ 

\global\let\CT@@do@color\relax  &   & \global\let\CT@@do@color\oriCT@@do@color 2018 & \cite{MPB_BlackHills_SouthDakota}  & GAtt & MPB & ML & SL & CLAS & NPAR/ PAR & RF/ LOG, LDA & GAtt \& non-attack  & Spectral bands and indices & Pixel-based     \protect\\  

\global\let\CT@@do@color\relax  &   & \global\let\CT@@do@color\oriCT@@do@color 2019 & \cite{UAV_PrecDetection_BBInf}  & GAtt, RAtt, GRAtt & ESBB & ML & SL & CLAS & PAR & MAXL & Dead, healthy, infested & GI, SR, GRVI, NDVI, GNDVI & Pixel-based    \protect\\  

\global\let\CT@@do@color\relax  &   & \global\let\CT@@do@color\oriCT@@do@color 2019 & \cite{Sens_LandSatOLI_TIRS_Foliar}  & GAtt, GRAtt & ESBB & ML & SL & CLAS & PAR & LREG & Severely stressed/moderately stressed/healthy trees  & \thead{Leaf traits (stomatal conductance, \\ chlorophyll fluorescence, water content)} & Pixel-based    \protect\\  

\global\let\CT@@do@color\relax  &   & \global\let\CT@@do@color\oriCT@@do@color 2019 & \cite{Sentinel2_GA_Landsat8}  & GAtt & ESBB & ML & SL - USL & CLAS - CLUS & PAR & RFG, PLS-DA - PCA & Healthy \& infested areas & SVIs & Pixel-based   \protect\\  

\global\let\CT@@do@color\relax  &   & \global\let\CT@@do@color\oriCT@@do@color 2019 & \cite{PineShootBeetle_Hyperspect_LiDAR}  & GAtt, RAtt, GRAtt & PSB & ML & SL & REG, SEG & NPAR/ PAR & RF, WSH & \thead{Healthy tree, slightly-/moderately-/\\severely-infected tree, dead tree} & Hyperspectral features, LiDAR metrics & Region-based  \protect\\  


\global\let\CT@@do@color\relax  &   & \global\let\CT@@do@color\oriCT@@do@color 2020 & \cite{MultiTemp_HyperMultispect_NorwaySpruce}  & GAtt & ESBB & ML & SL & CLAS & NPAR & RF & GAtt, root-rot, healthy & Spectral \& index features & Pixel-based   \protect\\  

\global\let\CT@@do@color\relax  &   & \global\let\CT@@do@color\oriCT@@do@color 2021 & \cite{Map_Gyrocopter_HyperField}  & GAtt, RAtt & ESBB & ML & SL & CLAS & NPAR & SAM, THR & Healthy, infested spruces (early: GAtt, late: brown crown) & \thead{Laboratory, HySpex \& hyperspectral \\ vegetation indices} & Pixel-based     \protect\\  

\global\let\CT@@do@color\relax  &   & \global\let\CT@@do@color\oriCT@@do@color 2021 & \cite{EarlyDet_Norway_CentEurope_Sentinel}  & GAtt, RAtt & IPSEB & ML & SL & CLAS & NPAR & RF & \thead{Healhty, RAtt during summer 2018, RAtt during autumn 2018,\\ green during 2018, RAtt during summer 2019}  & Spectral bands, SVIs, seasonal changes & Pixel-based    \protect\\  

\global\let\CT@@do@color\relax  &   & \global\let\CT@@do@color\oriCT@@do@color 2021 & \cite{EarlyDet_EuropeSpruce_NDRS}  & GAtt & ESBB & ML & SL & CLAS & NPAR/ PAR &  RF/ LDA  & Healthy/stressed trees  & Radar \& optical bands & Pixel-based  \protect\\  

\global\let\CT@@do@color\relax  &   & \global\let\CT@@do@color\oriCT@@do@color 2021 & \cite{Mapping_Boreal_HealthStatus}  & GAtt, RAtt, GRAtt & SBB & ML & SL & CLAS & NPAR &  RF  & Non-infested, GAtt, dead  & Spectral \& structural features & Pixel-based  \protect\\

\cdashline{2-15}
\global\let\CT@@do@color\relax  & \global\let\CT@@do@color\relax \multirow{50}{*}{\rotatebox[origin=c]{90}{Late Detection}} & \global\let\CT@@do@color\oriCT@@do@color 2003 & \cite{MPB_RedAttack_Strat_Landsat_BC}  & RAtt & MPB & ML & SL & CLAS & PAR & MAXL & RAtt/non-attack & Spectral response & Pixel-based  \protect\\  

\global\let\CT@@do@color\relax  &  & \global\let\CT@@do@color\oriCT@@do@color 2005 & \cite{RedAttack_MPB_HSR_Sat} & RAtt & MPB & ML & USL & CLUS & NPAR & ISODATA & Non-attacked, lightly/moderately/heavily RAtt & Spectral images & Region-based  \protect\\ 

\global\let\CT@@do@color\relax  &   & \global\let\CT@@do@color\oriCT@@do@color 2006 & \cite{EstProb_MPB_RedAttack}  & RAtt & MPB & ML & SL & CLAS & PAR & MAXL & RAtt/non-attack & EWDI, elevation, slope & Pixel-based   \protect\\  
 
\global\let\CT@@do@color\relax  &   & \global\let\CT@@do@color\oriCT@@do@color 2006 & \cite{IntegRS_AncillaryData_CharMPB} & RAtt & MPB & ML & SL & CLAS & NPAR/ PAR & DT/ LOGR & RAtt/non-RAtt stands & \thead{Forest inventory attributes, structural \& terrain information, \\ stand area, stand age, number of stems, location information} & Pixel-based \protect\\  

\global\let\CT@@do@color\relax  &   & \global\let\CT@@do@color\oriCT@@do@color 2009 & \cite{Map_WhiteBark_MPB_Sat}  & RAtt & MPB & ML & SL & CLAS & PAR & MAXL & \thead{Green \& brown herbaceous cover,\\ green (live) \& RAtt (dead) tree cover}  & RGI, green reflectance & Pixel-based  \protect\\  

\global\let\CT@@do@color\relax  &   & \global\let\CT@@do@color\oriCT@@do@color 2010 & \cite{AssessCanopy_MPB_GeoEye_Sat}  & RAtt, GRAtt & MPB & ML & SL & CLAS & PAR & MAXL & Green/red/gray canopy \& shadow & \thead{Pixels within 4 spectral classes \\ (green/red/gray canopy, shadow)} & Pixel-based   \protect\\  

\global\let\CT@@do@color\relax  &   & \global\let\CT@@do@color\oriCT@@do@color 2011 & \cite{EvalPoten_MultiSpect_MultipleStage}  & RAtt, GRAtt & MPB & ML & SL & CLAS & PAR & MAXL & \thead{Green trees, dead trees with red needles (RAtt), \\dead trees without needles (GRAtt), \& non-forest} & Pixels from several spectral bands & Pixel-based \protect\\  

\global\let\CT@@do@color\relax  &   & \global\let\CT@@do@color\oriCT@@do@color 2013 & \cite{Quanti_MixedSpecies_MultiTempHighRes_Sat}  & GRAtt & Unspecified & ML & SL/ USL & CLAS/ CLUS & NPAR/ PAR & EA/ GMM  & Live tree foliage, dead tree & SVIs & Pixel-based  \protect\\  

\global\let\CT@@do@color\relax  &   & \global\let\CT@@do@color\oriCT@@do@color 2013 & \cite{LiveDead_Basal_LiDAR_Journal}  & GRAtt & SBB, MPB & ML & SL & REG & NPAR & RF & Live \& Dead basal area & LiDAR canopy and topographic metrics & -    \protect\\  

\global\let\CT@@do@color\relax  &   & \global\let\CT@@do@color\oriCT@@do@color 2013 & \cite{Eval_DetectBB_SingMultiData_Landsat}  & RAtt & Unspecified & ML & SL & CLAS & PAR & MAXL & Undisturbed forest, RAtt, herbaceous, masked locations  & Pixels from several spectral bands & Pixel-based  \protect\\  

\global\let\CT@@do@color\relax  &   & \global\let\CT@@do@color\oriCT@@do@color 2014 & \cite{Toward_UAV_ForestMonitor} & RAtt, GRAtt & ESBB & ML & SL & CLAS & NPAR & RF & \thead{(spruces 0-1 or young), (spruces 2-3), \\ (dead spruces or wood), (beech), (the rest)} & Pixels & Pixel-based  \protect\\ 

\global\let\CT@@do@color\relax  &   & \global\let\CT@@do@color\oriCT@@do@color 2014 & \cite{Spatial_Multidate_SPOT_Landsat}  & RAtt, GRAtt & ESBB & ML & SL & CLAS & NPAR & RF & \thead{Old-attacked, current year-attacked, non-attacked, \\current year 1-attacked, current year 2-attacked} & \thead{Spectral signatures of SPOT \& Landsat \\ original bands} & Pixel-based  \protect\\  

\global\let\CT@@do@color\relax  &   & \global\let\CT@@do@color\oriCT@@do@color 2014 & \cite{MPB_GrowthTrend_Landsat} & GRAtt & MPB & ML & SL & CLAS & NPAR & RF, MAXL, EIGEN, DT & Healthy, MPB mortality, clearcuts & Spectral bands and indices & Pixel-based  \protect\\  

\global\let\CT@@do@color\relax  &   & \global\let\CT@@do@color\oriCT@@do@color 2015 & \cite{Photogram_Hyperspectral_TreeLevel}  & YAtt, GRAtt & ESBB & ML & SL & CLAS & NPAR & KNN & Healthy/infested/dead trees & Spectral bands & Region-based  \protect\\   
 
\global\let\CT@@do@color\relax  &   & \global\let\CT@@do@color\oriCT@@do@color 2015 & \cite{Detect_MPB_Ponderosa_HighResAerial}  & RAtt & MPB & ML & SL & CLAS & PAR & MAXL, THR & RAtt/non-RAtt & Pixels from several spectral bands & Pixel-based     \protect\\

\global\let\CT@@do@color\relax  &   & \global\let\CT@@do@color\oriCT@@do@color 2015 & \cite{SpectTemporal_Defoliat_Landsat} & GRAtt & MPB, WSB & ML & SL & CLAS, SEG & NPAR & RF & Harvest, fire, insects (MPB, WSB) & Disturbance \& recovery metrics & Pixel-based  \protect\\  

\global\let\CT@@do@color\relax  &   & \global\let\CT@@do@color\oriCT@@do@color 2016 & \cite{PercentMortality_MPB_Damage} & RAtt, GRAtt & MPB & ML & SL & REG, CLAS & NPAR/ PAR & \thead{RF, LSVM, PSVM, BLOGR/\\ GLM, BETA} & Dead/live tree, live vegetation (non-tree), bare, shadow & \thead{Elevation, slope, aspect, Landsat-8 bands,\\ NDVI, principal components} & Pixel-based   \protect\\  

\global\let\CT@@do@color\relax  &   & \global\let\CT@@do@color\oriCT@@do@color 2018 & \cite{UrbanForest_Hyperspec_UAV_Aircraft} & YAtt, GRAtt & ESBB & ML & SL &  CLAS & NPAR & SVM & Healthy/infested/dead trees & Spectral bands, SVIs & Pixel-based   \protect\\ 

\global\let\CT@@do@color\relax  &   & \global\let\CT@@do@color\oriCT@@do@color 2018 & \cite{Framework_Ecoregion_Spaceborne}  & RAtt & DSBB & ML & SL & CLAS & NPAR & RF & Green/red-stage conifer, non-conifer  & \thead{Eight radiometrically normalized bands,\\ SVIs} & Pixel-based    \protect\\  

\global\let\CT@@do@color\relax  &   & \global\let\CT@@do@color\oriCT@@do@color 2018 & \cite{RapidEye_Detect_AreaSpruce}  & RAtt, GRAtt & ESBB & ML & SL & CLAS & NPAR & RF & Dead wood \& buffer (not infested) & 24 synthetic NDVI time steps & Pixel-based   \protect\\  

\global\let\CT@@do@color\relax  &   & \global\let\CT@@do@color\oriCT@@do@color 2019 & \cite{Comp_WorldViewLandsat_SVMNeuralNet}  & RAtt, GRAtt & ESBB & ML & SL &  CLAS & NPAR & SVM, ANN & \thead{Healthy/affected/regenerating/dead/clear-cut forests, wetlands, \\permanent grasslands, water bodies, artificial surface} & Spectral bands & Pixel-based   \protect\\  

\global\let\CT@@do@color\relax  &   & \global\let\CT@@do@color\oriCT@@do@color 2020 & \cite{Class_HealthStat_UAS_Multispect}  & RAtt, GRAtt & Unspecified & ML & SL &  CLAS & NPAR & SVM & \thead{Broadleaves, dead/healthy/infested Norway spruce, \\ dead/healthy/infested Scots pine} & SVIs, textural bands & Pixel-based     \protect\\  

\global\let\CT@@do@color\relax  &   & \global\let\CT@@do@color\oriCT@@do@color 2020 & \cite{Sat_DecisionSupport_TANABBO} & GRAtt & ESBB & ML & SL & REG & PAR & LOGR, LREG & Strong damage, very strong damage & \thead{Distance from existing spots, NDVI, solar radiation, age \\ structure, stand density, wood volume per hectare} & Pixel-based      \protect\\  

\global\let\CT@@do@color\relax  &   & \global\let\CT@@do@color\oriCT@@do@color 2020 & \cite{Habitat_Dynamics_NorwayDieback} & GRAtt & ESBB & ML & SL & CLAS, REG & NPAR & RF, BRT & Shadows, dead, living trees & CIR aerial images & Pixel-based   \protect\\  

\global\let\CT@@do@color\relax  &   & \global\let\CT@@do@color\oriCT@@do@color 2020 & \cite{GF2Sentinel2_Turpentine_China}  & GAtt, RAtt, GRAtt & RTB & ML & SL & CLAS, REG & NPAR & RF, SVM, CART & RAtt, GRAtt & \thead{Spectral information, spectral indices,  \\textural information} & \thead{Pixel-based, \\ Region-based}    \protect\\ 

\global\let\CT@@do@color\relax  &   & \global\let\CT@@do@color\oriCT@@do@color 2021 & \cite{Early_Hyper_Results_Bavaria} & YAtt, RAtt, GRAtt & Unspecified & ML & SL & CLAS & NPAR & RF & \thead{Single needle age (four needle classes (2013-2010)) \\ \& the tree crown}  & \thead{Individual spectra \& their first and \\ second derivatives, SVIs} & Pixel-based   \protect\\  

\global\let\CT@@do@color\relax  &   & \global\let\CT@@do@color\oriCT@@do@color 2021 & \cite{SubtleChange_Landsat_MPBSpruce}  & RAtt, GRAtt & MPB, SBB & ML & SL & REG, CLAS & NPAR/ PAR & PTCLAS/ LOGR & \thead{MPB (RAtt, GAtt, herbaceous, bare soil, shadow); \\SBB (shadows, non-forest, green trees, GRAtt)} & SVIs, spectral bands & Pixel-based  \protect\\  

\global\let\CT@@do@color\relax  &   & \global\let\CT@@do@color\oriCT@@do@color 2021 & \cite{CrosInteract_Mortality}  & GRAtt & WPB & ML & SL & REG & NPAR & BLOGR & live/dead tree, ponderosa pine/other trees & \thead{Proportion of host trees, mean height of trees, \\count of trees, site-level climatic water deficit} & Pixel-based    \protect\\  

\global\let\CT@@do@color\relax  &   & \global\let\CT@@do@color\oriCT@@do@color 2021 & \cite{ML_SpatialDist_BBI}  & GRAtt & ESBB & ML & SL & REG, CLAS & NPAR/ PAR & \thead{DT, RF, KNN, SVM, ETC, GRADBC/ \\LOG, LDA, QDA, GNBAYES} & Damaged/undamaged forest & \thead{Distances, global solar radiation, NDVI, forest age,\\ spruce percentage, wood volume, stocking} & Pixel-based   \protect\\ 

\global\let\CT@@do@color\relax  &   & 2022 & \cite{Multispectral_Benefits_Spruce_Decline}  & GRAtt, RAtt, GRAtt & ESBB & ML & SL & CLAS & NPAR & RF & Healthy, declined, dead & Statistical features, SVIs & Pixel-based   \protect\\

\global\let\CT@@do@color\relax  &   & \global\let\CT@@do@color\oriCT@@do@color 2022 & \cite{Northwenmost_Model_Climate} & GRAtt & ESBB & ML & SL & REG & PAR & LREG &  - & Meteorological variables  & Pixel-based    \protect\\  

\global\let\CT@@do@color\relax  &   & 2022 & \cite{Landsat_SouthPineBeet_Severity}  & RAtt & SPB & ML & SL & CLAS, REG & NPAR & RF, MDC & RAtt with varied severity values (1–100\%) & SVIs, transformation values & Pixel-based   \protect\\ 

\hline \hline


\global\let\CT@@do@color\relax \multirow{7}{*}{\rotatebox[origin=c]{90}{Deep Learning}} & \global\let\CT@@do@color\relax \multirow{1}{*}{\rotatebox[origin=c]{90}{Focus}} & \global\let\CT@@do@color\oriCT@@do@color 2021 & \cite{UAS_Multispect_DL} & GAtt, YAtt & Unspecified & DL & SL & CLAS & NPAR & CNN & Pines, YAtt, GAtt, non-infested trees & Deep features & \thead{Custom-CNNs, \\ DenseNet-169}      \protect\\ 

\global\let\CT@@do@color\relax  &   & \global\let\CT@@do@color\oriCT@@do@color 2022 & \cite{Detection_UAV_YOLOs} & GAtt, YAtt, RAtt, GRAtt & ESBB & DL & SL & DET & NPAR & CNN & Four attack phases & Deep features &  YOLOv2, YOLOv3, YOLOv4    \protect\\  

\cdashline{2-15}

\global\let\CT@@do@color\relax  & \global\let\CT@@do@color\relax \multirow{9}{*}{\rotatebox[origin=c]{90}{Late Detection}} & \global\let\CT@@do@color\oriCT@@do@color 2019 & \cite{BB_Fir_DL} & RAtt, GRAtt & FFBB & DL & SL & DET, CLAS & NPAR & CNN & \thead{Healthy tree/recently attacked, beetle-colonized tree,\\ recently died tree, deadwood} & Deep features & IP, Custom-CNN        \protect\\  

\global\let\CT@@do@color\relax  &   & \global\let\CT@@do@color\oriCT@@do@color 2019 & \cite{DeadWood_FCNDenseNet} & GRAtt & Unspecified & DL & SL & SSEG & NPAR & CNN & Standing \& fallen dead tree & Deep features &  FCN-DenseNet    \protect\\  

\global\let\CT@@do@color\relax  &  & \global\let\CT@@do@color\oriCT@@do@color 2020 & \cite{ForestHealth_Aerial_DL} & GRAtt & Unspecified & DL & SL & ISEG & NPAR & CNN & N/A & Deep features & Mask-RCNN       \protect\\ 

\global\let\CT@@do@color\relax  &   & \global\let\CT@@do@color\oriCT@@do@color 2020 & \cite{DistSeg_Sat_DL} & RAtt, GRAtt & Unspecified & DL & SL &  SSEG  & NPAR & CAE & Damaged \& healthy trees & Deep features & U-Net   \protect\\  

\global\let\CT@@do@color\relax  &   & \global\let\CT@@do@color\oriCT@@do@color 2021 & \cite{DeadTree_PixObj_WorldView} & RAtt, GRAtt & ESBB & DL/ ML & SL & CLAS & NPAR & CNN & \thead{Standing dead tree, declining tree, live conifer, live broadleaf, \\bare ground, grassland, artificial surface} & Deep features & \thead{MFDNN/ RF, \\ SVM, KNN}  \protect\\  
\global\let\CT@@do@color\relax  &   & \global\let\CT@@do@color\oriCT@@do@color 2021 & \cite{SickFir_UAV_DL} & GRAtt & SFBB & DL & SL & CLAS & NPAR & CNN & Healthy, sick tree, deciduous tree & Deep features & \thead{AlexNet, SqueezeNet, VGG, \\ ResNet, DenseNet}   \protect\\

\global\let\CT@@do@color\relax  &   & \global\let\CT@@do@color\oriCT@@do@color 2022 & \cite{Ours_DL_BarkBeetle} & YAtt, RAtt, GRAtt & MEXPB & DL & SL & CLAS & NPAR & CNN & Healthy, YAtt, RAtt, leafless & Deep features & \thead{RetinaNet}   \protect\\


\hline 
  \end{tabular}
  
  }
\vspace{-0.3cm}
\label{tab_ML_all}  
\end{table*}
\begin{table*}[!b]
\vspace{-0.5cm}
\caption{Most common machine learning approaches used to detect bark beetle attacks: advantages and disadvantages.} 
\vspace{-0.3cm}
\centering 
 \resizebox{\textwidth}{!}{
\begin{tabular}{ c  c  c } 
\hline
Method & Advantages & Disadvantages   \protect\\ %
\hline\hline
MAXL &
\thead{
\textbullet\ Intuitively appealing (most likely outcome)
\textbullet\ Well-developed theoretical foundation \\
\textbullet\ Handling covarying data
\textbullet\ Performing well over a range of input  types and  conditions
}
&    
\thead{
\textbullet\ Strong assumptions about data structure \\
\textbullet\ Computationally expensive (for large-area high-resolution images)
}
\protect\\ 
\hdashline

LREG &
\thead{
\textbullet\ Wide applications
\textbullet\ Simple and efficient
}
&    
\thead{
\textbullet\ Linearity assumption
\textbullet\ Prone to overfitting, noise, and multicollinearity
\textbullet\ Sensitive to outliers
}
\protect\\ 
\hdashline
LOGR &
\thead{
\textbullet\ Wide applications (probability output) \\
\textbullet\ Handling nonlinearity, interaction effect, and power term
}
&  
\thead{
\textbullet\ Requiring large sample size to achieve stable results \\
\textbullet\ Suffering multicollinearity
}
\protect\\ 
\hdashline



PLS &
\thead{
\textbullet\ Handling multiple dependent/independent variables
\textbullet\ Maximizing covariance between data sets \\
\textbullet\ Handling multicollinearity, small data, different variables
\textbullet\ Robust to noise and missing data
}
&   
\thead{
\textbullet\ Difficult interpretation
\textbullet\ Not significant performance
}
\protect\\  
\hdashline
DT &  
\thead{
\textbullet\ Handleing feature interactions
\textbullet\ Dealing with linearly inseparable data \\ 
\textbullet\ Handling variety of data, missing values,  and redundant attributes
\textbullet\ Good generalization \\
\textbullet\ Robust to noise
\textbullet\ High performance for relatively small computational effort \\
\textbullet\ Simple method (easy to understand and visualize)
\textbullet\ Fast (quick training, low computational time) \\
\textbullet\ Requiring less data preprocessing
\textbullet\ Dealing with both categorical and numerical data \\
\textbullet\ Not constrained to class distributions
}
&    
\thead{
\textbullet\ Difficult to deal with high dimension data 
\textbullet\ Easily over-fit \\
\textbullet\ Computationally expensive to build a tree
\textbullet\ Cannot deal with complex interactions \\
\textbullet\ Error-propagation through trees
\textbullet\ Problem of data fragmentation \\
\textbullet\ Not good generalization for complex tree structures
\textbullet\ Unstable model \\
\textbullet\ Less appropriate for continuous variable estimation 
\textbullet\ Problematic for time-series data  \\
\textbullet\ Having trouble with non-rectangular regions
}
\protect\\  
\hdashline
SVM &
\thead{
\textbullet\ High accuracy
\textbullet\ Avoiding overfitting
\textbullet\ Flexible selection of kernels for nonlinearity \\
\textbullet\ Independent performance from number of features
\textbullet\ Good generalization \\
\textbullet\ Dealing with high-dimensional data
\textbullet\ Memory efficient
}
 &    
\thead{
\textbullet\ Complex algorithm
\textbullet\ Relatively high training time (very large datasets) \\
\textbullet\ Low performance when overlapped target classes or noisy dataset \\
\textbullet\ Requiring n-fold cross-validation (computationally expensive)
}
\protect\\  
\hdashline 
NBAYES & 
\thead{
\textbullet\ Not requiring huge dataset (small-sized training data is good enough) 
\textbullet\ Explicit probability calculation  \\
\textbullet\ Intensely fast (efficiently solve diagnostic problems) 
\textbullet\ Relatively simple 
}
&   
\thead{
\textbullet\ Comparatively a bad estimator
\textbullet\ Zero conditional probability problem \\
\textbullet\ Very strong assumption (independent class features)
}
\protect \\ 
\hdashline
MAXE & 
\thead{
\textbullet\ No inherent conditional independence assumptions
\textbullet\ Good accuracy \\
\textbullet\ Convex objective function (converge to a global optimum)
}
&    
\thead{
\textbullet\ Prone to overfitting
\textbullet\ Limited generalization
}
\protect\\ 
\hdashline 


RF &  
\thead{
\textbullet\ Fast and scalable
\textbullet\ Robust to noise
\textbullet\ Avoiding overfitting \\
\textbullet\ Simple explanation and visualization of input (no parameters required)
}
&    
\thead{
\textbullet\ Slow down as the number of trees increases
}
\protect\\  
\hdashline 
KNN & 
\thead{
\textbullet\ Simple 
\textbullet\ Well-suited for multi-model classes
\textbullet\ Independency (sample distribution from classification) \\
\textbullet\ Handling noisy training data
\textbullet\ Handling large training data
}
&    
\thead{
\textbullet\ Low efficiency (density estimation, parameter selection, noise and geometry irrelevant features) \\
\textbullet\ Dependent performance (data size and parameter value) 
\textbullet\ Computationally expensive
}
\protect\\  
\hdashline 
SAM & 
\thead{
\textbullet\ Fast and simple
\textbullet\ Comprehensible \\
\textbullet\ Handling illumination differences and scaled noise
}
&    
\thead{
\textbullet\ Failure if vector magnitude is important for discriminating information \\
\textbullet\ Insensitivity to certain physiologic changes 
\textbullet\ Misclassifications for similar spectra
}
\protect\\  
\hdashline
 

CART & 
\thead{
\textbullet\ Multivariate logistic regression
\textbullet\ No assumptions about distribution \\
\textbullet\ Dealing with missing data
\textbullet\ Simple interpretation
}
&    
\thead{
\textbullet\ Computationally expensive
\textbullet\ Different performance (small data change) \\
\textbullet\ Not support continuous values
}
\protect\\ 
\hdashline 




NN & 
\thead{
\textbullet\ Dealing with nonlinear or dynamic relationships \\
\textbullet\ No strong or priori assumptions 
\textbullet\ Robust to irrelevant input and noise \\
\textbullet\ Readily adopting auxiliary data (textural information, slope, aspect, elevation)
\textbullet\ Quite flexible \\
\textbullet\ Good performance (particular problems)
}
&    
\thead{
\textbullet\ Computationally expensive (training)  \\
\textbullet\ Dependent performance(data size, number of hidden layers, parameter values) \\
\textbullet\ Difficult interpretation
\textbullet\ No guarantee finding global optimum
}
\protect\\ 
\hdashline 
CNN & 
\thead{
\textbullet\ Not requiring feature engineering
\textbullet\ Superior performance
\textbullet\ Good generalization
}
&   
\thead{
\textbullet\ Computationally expensive
\textbullet\ Demanding large data
\textbullet\ Not explicit interpretations
}
\protect\\ 
\hline
  \end{tabular}
  \label{tab_ML_comp}  
  }
\end{table*}

The use of other ML algorithms has been also explored in several studies.
For instance, the random frog algorithm \cite{RandomFrog} was used to initially search for the most relevant SVIs (derived from Sentinel-{2} and Landsat-{8}) in the model space, in which the partial least squares discriminant analysis models healthy \& GAtt trees \cite{Sentinel2_GA_Landsat8}. In addition, the \textit{principal component analysis} (PCA) visually examined the separability of these trees, which showed a slight crossover of SVIs from Sentinel-{2} and apparent overlap for Landsat-{8} SVIs. 
Other than that, the impact of GAtt symptoms on SVIs calculated from a Landsat-{8} image was evaluated using partial least squares regression \cite{Sens_LandSatOLI_TIRS_Foliar}. The selected SVIs were then examined using either the PCA for determining the importance of each SVI or infestation map generation using linear regression. 
Another example is the maximum likelihood classification of healthy and GAtt trees using SVIs \cite{UAV_PrecDetection_BBInf}. It was noted that the time for data acquisition had more impact on classification accuracy than feature selection; However, the accuracy was affected by the changes in SVIs, and GAtt trees were overestimated. 

Different ML algorithms have been examined to compare their effectiveness or achieve higher reliability and performance. 
In \cite{EarlyDet_TerraSarX_RapidEye}, the capability and feature selection of the \textit{generalized linear model} (GLM), maximum entropy, and RF were compared for the classification of GAtt and healthy areas. In this regard, the maximum entropy identified the most suitable features and achieved the best classification accuracy. The accuracy of this algorithm was not affected by applying cross-validation, indicating that it was well-fitted with a limited number of samples and was not prone to overfitting. However, cross-validation and sample size heavily influenced the accuracy of the RF classifier. 
Apart from that, logistic regression and RF algorithms were used to classify two (GAtt vs. healthy) or three (GAtt, healthy, and dead) classes of trees using various features (i.e., single/combined spectral bands \& SVIs) \cite{EarlyDet_NorwaySpruce_WorldView2}. Both algorithms achieved comparable results when all spectral bands were included. However, the classification accuracy was low for the healthy \& GAtt classes due to significant overlap in their spectral signatures, despite the simple separation of the dead class. Although selecting unsuitable features negatively affected the LR model, the RF classifier was able to handle this problem more effectively. 
In \cite{MPB_BlackHills_SouthDakota}, the RF and logistic regression, as well as LDA using different feature sets, were also used to classify GAtt and non-attacked tree crowns. In most cases, the logistic regression achieved a higher classification accuracy through cross-validation with training data due to its robustness to invalid assumptions. However, the RF model performed comparably or even better after being transferred to an independent dataset. The logistic regression and RF outperformed the LDA algorithm based on their robustness to non-normality \& multicollinearity, in contrast to LDA's susceptibility to assumption violation.
\subsection{ {Classical ML Algorithms used for Hyperspectral Analysis}} \label{sec:ML_classic_HS}
In the HS analysis, the RF algorithm was primarily applied for feature selection and model evaluation \cite{SpectEvid_Early_Engelmann,PineShootBeetle_Hyperspect_LiDAR,Mapping_Boreal_HealthStatus,MultiTemp_HyperMultispect_NorwaySpruce}. 
For instance, this algorithm determined important spectral bands \& SVIs to identify early-stage infestation \cite{SpectEvid_Early_Engelmann} or measure the importance of different features at the crown scale \cite{PineShootBeetle_Hyperspect_LiDAR}. 
In \cite{PineShootBeetle_Hyperspect_LiDAR}, the RF algorithm was utilized to compare three models using HS features, LiDAR metrics, or their combination after the data dimension was reduced through PCA. However, all models failed to accurately predict even moderately damaged trees despite the higher performance of the combined model. 
A comparison of the accuracy of tree health status classification using MS and HS images was conducted using this algorithm \cite{MultiTemp_HyperMultispect_NorwaySpruce}. Although the analysis indicated slightly higher accuracy for HS observations, the performance was not much better than random classification due to its difficulties to identify GAtt trees and root-rot classes. 
Lastly, this algorithm was applied to classify GAtt trees at the crown level based on specific-domain information and fused data \cite{Mapping_Boreal_HealthStatus}. To this end, various RF classifiers were trained on the most critical MS/HS features (from corresponding RF feature selector) using spectral-only, structural-only (e.g., SfM \& LiDAR data), or fused data to predict health status probabilities.  {It was found that the fusion models achieved a higher performance even though classification accuracy did not improve with higher spatial resolutions than \mbox{{6} $cm$ \cite{Mapping_Boreal_HealthStatus}}}.

Analysis of HS data has been also conducted using other ML algorithms.
For instance, the SVMs were used to produce bark beetle infestation maps, where the evolutionary genetic algorithm selected the most stable features of high dimensional HS data \cite{AngularVegIndex_Spectroscopy_Bavarian}. However, the healthy \& GAtt trees and also medium \& high damages were not distinguishable as the primary limitations resulted from the missing reference data and visual definition of damage classes.
In \cite{Assess_Hyperspec_Mortality}, the SVM classifier was applied to distinguish tree mortality stages, while the genetic algorithm was used to prune the feature space dimension and identify relevant wavelengths. Accordingly, the genetic algorithm reduced the complexity of the classifier and improved its transferability. However, the mortality stages were not sufficiently classified due to highly confusing diagnostics of critical classes, i.e., healthy \& GAtt trees. Therefore, the combination of these two algorithms was insufficient operationally as the most important classes with similar spectral characteristics were substantially overclassified. 
In another study \cite{HyperRS_MPB_Emphasis_Previsual}, the \textit{nearest-neighbor} (NN) algorithm was employed to cluster the absorption feature shape and cell structure continuum changes of healthy and GAtt trees. The clusters were shape metrics to assess pigment and water absorption features, which resulted in more variation in prenormalized depths than healthy trees, possibly related to foliar tissue changes or decreased canopy volume. Additionally, the \textit{spectral angle mapper} (SAM) algorithm was used to evaluate the effects of normalization of different scale data (i.e., benefit from continuum removal analysis) on classification accuracy. Accordingly, the pigment and water absorption features were able to distinguish infected from healthy trees. However, the results for GAtt trees were less stable, and the band depth metric was less consistent. 
The SAM classifier was applied in \cite{Map_Gyrocopter_HyperField} as a preprocessing step to mask pixels that contain irrelevant or ambiguous information (e.g., background \& mixed pixels), considering the applicability of this algorithm for high-dimensional HS data. Following that, a threshold-based approach was used to evaluate the classification performance and transferability of laboratory \& airborne HS indices.
Lastly, the decision tree classifier in \cite{Forecast_Vital_Hyperspect} examined the ability to discriminate between HS data and spectral information, derivatives, and indices based on its relatively small structure and potential ability to apply to this data. However, the separation was only partial or low to moderate because of the mixed information of classes and changes in spectral signatures associated with forest regeneration, ground vegetation layer, and stand age.
\subsection{ {Deep Learning-based Methods used for Multispectral Analysis}} \label{sec:ML_DL}
Although DL approaches have been extensively developed for diverse computer vision applications (e.g., \cite{DL_Tracking,DL_Detection}), there are two methods that use DNNs for GAtt detection \cite{UAS_Multispect_DL,Detection_UAV_YOLOs}. 
In \cite{UAS_Multispect_DL}, three CNNs were evaluated to determine their potential to classify infested trees. It trained i) a customized network with three convolutional layers from scratch (without data augmentation), ii) a customized network with six convolutional layers from scratch (with and without data augmentation), and iii) a pre-trained DenseNet-{169} network \cite{DenseNet} applying transfer learning (with data augmentation). The DenseNet weights were pre-trained using the ImageNet dataset \cite{ImageNet}, while the top classification layers were fine-tuned using MS image patches. To avoid information loss and degradation of model accuracy, raw spectral bands (RGB bands or RBG \& red-edge bands) were used to train the networks. As a result, the two customized networks provided higher performance than the DenseNet model, and the models using RGB bands were more effective than those using the red-edge band. Despite not significantly improving classification accuracy, data augmentation did reduce misclassifications of GAtt and healthy trees. However, despite partially alleviating this problem through data augmentation, several GAtt trees were classified as healthy.  

In another work \cite{Detection_UAV_YOLOs}, three versions of YOLO network architectures were used to detect trees at different stages of infestation. First, the quality of low-contrast images was improved using the balanced contrast enhancement technique \cite{BCET_contrast}, equalizing the histogram of pixel intensity distribution without changing its pattern. Following that, various data augmentations were applied to an acquired unbalanced dataset, including rotation, horizontal/vertical flipping, and resizing. Then, three networks, YOLO-v{2} \cite{YOLO_v2} (based on Darknet-{30}), YOLO-v{3} \cite{YOLO_v3} (based on Darknet-{53} \& ResNet \cite{ResNet}), and YOLO-v{4} \cite{YOLO_v2}, were trained to detect infested trees. With the YOLO-v{4}, the performance was higher \& the speed was much faster, as opposed to the YOLO-v{2} and YOLO-v{3} with limitations of detecting small objects and being highly complex. The YOLO-v{4} architecture was shown to produce better results with fewer false positives, even though the dataset was relatively small. Moreover, pixel contrast enhancement increased classification accuracy.
\subsection{ {Evaluation Metrics}} \label{sec:ML_PerfAssess}
The methods for detecting bark beetle-induced tree mortality are commonly evaluated by calculating the confusion matrix, \textit{average precision} (AP, or \textit{mean average precision} (mAP)), \textit{root mean square error} (RMSE), coefficient of determination ($R^2$, or goodness of fit indicator), and \textit{intersection-over-union} (IoU or Jaccard-index) metrics. 
The confusion matrix compares the predicted disturbance values/classes to the ground-truth ones and achieves four outputs of \textit{true-positive} (TP), \textit{true-negative} (TN), \textit{false-positive} (FP), and \textit{false-negative} (FN). While the TP and TN represent the correct predictions of infested and non-infested trees, the FP \& FN show the false predictions for the non-infested and infested trees, respectively. The confusion matrix summarizes the model performance and can also be used to calculate various performance metrics, e.g., accuracy, precision, etc.
Given $Total=TP+TN+FP+FN$, the agreement metrics are derived as follows.
\begin{alignat}{4}
    \rm{Accuracy = \frac{TP+TN}{Total}} \quad \text{(1),} \quad  && 
    \rm{Precision = \frac{TP}{TP+FP}} \quad \text{(2),} \quad    && 
    \rm{Recall = \frac{TP}{TP+FN}} \quad \text{(3),}  \quad      &&
    \rm{F1 = \frac{2 * (Precision * Recall)}{Precision + Recall}} \quad \text{(4)} \notag
\end{alignat}
These metrics evaluate the performance of a model from different aspects. While accuracy is a general metric that represents the overall correctness of the model's predictions, the precision and recall (or sensitivity) metrics assess the model's performance for specific instances. The precision (or user’s accuracy) is more critical when false positives are more costly than false negatives since high precision indicates that the model is usually correct when it predicts positives. However, recall (or producer’s accuracy) is of particular importance where positive instances are relatively rare because high sensitivity implies that the model can detect most positive instances. The F{1} score is a harmonic mean of precision and recall to provide a balanced measure of the two metrics, ranging from zero (the worst) to one (the best).

In addition, Cohen's Kappa metric can be calculated by
\begin{alignat}{4}
    \rm{kappa = \frac{Accuracy-P_e}{1-P_e}} \;\;\;\; \text{(5)} \quad \text{ where} \quad && \rm{P_e=P_0+P_1}  \;\; \text{,} \qquad  && \rm{P_0 = \frac{TP+FP}{Total}\times\frac{TP+FN}{Total}} \;\; \text{,} \quad && \rm{P_1 = \frac{FN+TN}{Total}\times\frac{FP+TN}{Total}}. \notag  
\end{alignat}
This metric measures the level of chance agreement and ranges from {-1} (i.e., complete disagreement) to {1} (i.e., almost perfect agreement). While a kappa value of zero indicates no agreement beyond chance, kappa values higher than {0.8} are generally considered strong agreement and a value below {0.6} reflects poor agreement. Moreover, the error metrics of \textit{commission error} (CE), \textit{omission error} (OE), and \textit{relative bias} (RB) are defined by Eq.~(6)-(8).
\begin{alignat}{3}
    \rm{CE = \frac{FP}{TP+FP}} \qquad \text{(6)} \qquad  \qquad  \qquad  \qquad  && \rm{OE = \frac{FN}{TP+FN}} \qquad \text{(7)} \qquad  \qquad  \qquad  \qquad  && \rm{RB = \frac{FP-FN}{TP+FN}} \qquad \text{(8)} \notag  
\end{alignat}
The CE metric (or false positive error) indicates the error of inclusion when samples are included in a class to which they do not belong, whereas the OE metric (or false negative error) represents the error of exclusion when samples are excluded from a class to which they do belong. As well, the RB metric is a biased assessment of model accuracy, indicating the tendency of a model to consistently predict too high (over-prediction) or too low (under-prediction) values compared to the actual values.

The mAP metric is computed by Eq.~(9) where $AP$ and $M$ represent the area under the precision-recall curve and the number of classes, respectively. Besides, the RMSE is defined by Eq.~(10) in which the $\hat{y}$, $y$, and $n$ are the predictions, ground-truths, and the number of ground-truths, respectively. Also, the $\rm{R^2}$ can be calculated by Eq.~(11) where the mean of $y$ is denoted as $\bar{y}$.
\begin{alignat}{3}
    \rm{mAP = \frac{1}{M}\sum_{i=1}^{M}AP_i} \qquad \text{(9)} \qquad  \qquad  \qquad  && \rm{RMSE = \sqrt{\sum_{i=1}^{n}\frac{({y_{i}-\hat{y_{i}})}^2}{n}}} \qquad \text{(10)} \qquad  \qquad  && \rm{R^2 = 1-\frac{\sum_{i=1}^{n}({y_i}-\hat{y_i})^2}{\sum_{i=1}^{n}({y_i}-\bar{y})^2}} \qquad \text{(11)} \notag  
\end{alignat}
The mAP metric measures the accuracy of the model in localizing objects in an image, ranging from {0} to {1} (i.e., perfect accuracy). The RMSE metric is defined as the average deviation of the predictions from the actual values, where a lower value indicates better performance. The $\rm{R^2}$ metric indicates the goodness of fit of a regression model with values ranging from {0} to {1}, with higher values indicating a better fit.
Finally, the IoU metric represents the relative spatial overlap between the predicted bounding boxes of infested trees and ground-truth ones (i.e., $TP/(TP+FP+FN)$). This metric uses both FP and FN for evaluating the overall performance of a detection model and it is more sensitive to the size of the predicted area (compared to precision and recall metrics) since even a slight deviation in the estimated area size can have a significant impact on the IoU score.
\section{ {Empirical Evaluations}} \label{sec:Evals}
 {This section compares the ML/DL methods, spectral signatures, and SVIs used for the early detection of bark beetle attacks on the basis of their MS or HS analyses.}

\subsection{{Machine Learning Methods for Early Detection}} \label{sec:ML_quantitative_comp}

\subsubsection{\textbf{{Classical Algorithms for Multispectral Analysis:}}} \label{sec:Quant_MS}
 {In this section, we discuss the quantitative comparison of various algorithms and then explore the accuracy of independently applied algorithms.
For example, the maximum entropy, RF, and GLM were compared in \mbox{\cite{EarlyDet_TerraSarX_RapidEye}} to classify plots with GAtt ({{15}} plots) and without GAtt ({{230}} plots), where the ground-truth data for GAtt areas was recorded in the field and those for healthy (or background) data were selected using aerial images. Although the RF and GLM utilized healthy \& GAtt plots to analyze their relationship and features extracted from forest hexagon areas, maximum entropy required no healthy plots to explain GAtt distributions. The maximum entropy achieved the highest classification accuracy following \textit{leave-one-out cross-validation} (LOOCV), as the best models on RapidEye imagery and TerraSAR-X data resulted in a kappa of 0.51 and 0.23, respectively. In addition, combining the spectral information of both sensors improved the performance to a kappa of 0.74. The prediction maps with the maximum entropy model indicated no OE (i.e., FN) but many CEs (i.e., FP) because of high spectral variability in areas with mixed conifers and deciduous trees and areas close to forest borders. Moreover, the accuracy of the results depended heavily on the sample size, the number of ground-truth data, and the selection of handcrafted features.
Another study \mbox{\cite{EarlyDet_NorwaySpruce_WorldView2}} used RF (with defining two or three classes) and logistic regression (with healthy and GAtt classes) to classify infestation stages based on four or eight bands of WorldView-2 images taken in June and July. 
The overall accuracy to classify healthy, GAtt, and dead trees using the RF over the June and July images were respectively 74.4\% and 76.2\%, while the kappa coefficients were approximately 0.6. In addition, the overall accuracy was decreased by using only four bands to 71.6\% and 70.2\% and the kappa coefficients to 0.53 and 0.51, respectively. 
To specifically distinguish between the healthy and GAtt trees, the overall accuracy and kappa coefficient obtained from RF and logistic regression using eight bands were similar for two images around 70\% and 0.4, respectively. 
However, for the June (or July) image, the overall accuracy for the RF and logistic regression using four bands decreased to 68\% (or 65\%) and 72\% (or 70\%), respectively. Moreover, difference indices and band ratio indices resulted in very low kappa values (i.e., close to a random model) indicating poor model accuracy. 
The RF achieved the best results when using eight spectral bands of the July image, but the spectral differences between healthy and GAtt trees were minor and blurred by high within-class variances. 
Although the overall accuracy of some studies has been euphemized by the fact that more than 90\% of the samples come from healthy trees (e.g., \mbox{\cite{EarlyDet_TerraSarX_RapidEye}}), a balanced set of reference trees for each image (i.e., 257 vs 272 trees in June, and 256 vs 245 trees in July for healthy vs GAtt trees) was used in this study.
In \mbox{\cite{MPB_BlackHills_SouthDakota}}, the RF, logistic regression, and LDA algorithms were compared to classify healthy and GAtt trees based on 36 features (8 spectral bands from WorldView-2 and 28 SVIs). An assessment of these algorithms was conducted using the LOOCV and a validation data split, ensuring more aligned splits of data. 
While the logistic regression achieved the best overall accuracy of 75.9\% using all features and LOOCV (vs 73.1\% and 68.9\% respectively for the RF and LDA), the RF provided the best overall accuracy of 70.6\% using eight spectral bands on validation data split (vs 67.7\% and 66.3\% respectively for the LDA and logistic regression). The logistic regression algorithm had the least transferability among the three algorithms, as its performance decreased by about 9.6\% using validation data split. It was also found that there were high within-class variances and spectral confusion for the GAtt and healthy tree classes. 
In \mbox{\cite{EarlyDet_EuropeSpruce_NDRS}}, the LDA and RF were employed to quantify the separability of individual bands and SVIs from Sentinel-1 \& Sentinel-2 data and classify healthy trees as well as stressed trees affected by the early-stage of bark beetle attacks (i.e., GAtt phase from May to July). The overall accuracy of the LDA algorithm to classify healthy and stressed trees using the NDRS, NDWI, and RDI indices from April to August increased from 83\%, 75\%, and 77\% to 84\%, 77\%, and 78\%, respectively. Although classification accuracy varied between 70\% and 85\% for RF models combining multiple bands, there was no significant difference between the two classes when comparing them before the attack (April) and the end of the GAtt phase (August), indicating little impact on spectral differences. \\
\indent The importance of spectral bands and SVIs from Sentinel-2 cloud-free images for early detection and classification of healthy, GAtt and RAtt pixels was evaluated in \mbox{\cite{EarlyDet_Norway_CentEurope_Sentinel}} using the RF algorithm. Sentinel-2 bands were found to have an overall importance accuracy of 87\%, whereas seasonal change and absolute index values obtained overall importance accuracy of 96\% and 87\%, respectively. The results highlighted the importance of temporal separation to increase accuracy by collecting data throughout the season. Following that, the RF algorithm classified early-infested trees from healthy ones with an overall classification accuracy of 78\% for the GAtt class. The results showed that among 40 trees classified as healthy, eight trees belonged to the GAtt class. Meanwhile, 30 of the overall 40 trees classified as GAtt trees were correctly predicted, and ten were healthy. Accordingly, the performance based on the accuracy, precision, and recall metrics were 77.5\%, 75\%, and 78\%, respectively. In this study, misclassifications of bark beetle infestation symptoms were attributed to confusion over spectral responses when the size of individual trees or areas was smaller than the Sentinel-2 pixel size. 
On the other hand, the Random Frog algorithm was used to select SVIs (from Sentinel-2 and Landsat-8) that influenced spectral separability between the healthy and GAtt plots \mbox{\cite{Sentinel2_GA_Landsat8}}. Accordingly, Sentinel-2 images showed higher sensitivity to bark beetle GAtt-induced changes (i.e., stressed canopies) than Landsat-8 images, revealing more SVIs with differences between healthy and GAtt samples (i.e., 17 out of 35 SVIs in contrast to 8 out of 24 for Landsat-8). Sentinel-2 had more correctly-matched pixels with reference infestation data than Landsat-8 (i.e., 67\% vs 36\%) based on selected SVIs and a defined threshold. Also, it was found that the pixels detected as GAtt were located within 500 meters of last year's infestation zone.
In \mbox{\cite{Sens_LandSatOLI_TIRS_Foliar}}, the linear regression method was used to generate a map of canopies stressed by bark beetle GAtt from Landsat-8 imagery. Taking 40 healthy and 21 GAtt plots, this method assessed the stress intensity of generated map, such that 66\%, 21\%, and 13\% of GAtt pixels were assigned to severely stressed, moderately stressed, and healthy (i.e., false negative) pixels. Most GAtt pixels were located within the severely stressed class, enabling the stress map to identify areas with a high potential of bark beetle GAtt.
Lastly, the maximum likelihood classifier was used in \mbox{\cite{UAV_PrecDetection_BBInf}} to distinguish between healthy and GAtt trees based on the difference of SVIs in June. For this period, the CE and OE were 3\% and 25\% for healthy trees, and 53\% and 10\% for GAtt trees, respectively. The best overall accuracy of this method was 78\%, although a slight overestimation of the number of infested trees was made because the opposite is more harmful in practice.} 

\subsubsection{\textbf{Deep Learning-based Methods for Multispectral Analysis:}} 
 {The potential of DL-based methods for detecting bark beetle attacks was investigated as an alternative to classical methods requiring heuristic selections of appropriate transformations and handcrafted features \mbox{\cite{UAS_Multispect_DL,Detection_UAV_YOLOs}}. 
In \mbox{\cite{UAS_Multispect_DL}}, first, a simple three-layer CNN was trained from scratch for 40 epochs with a batch size of one and without data augmentation. The experiments were conducted on two models with three (RGB, denoted as 3b) or four spectral bands (RGB and red-edge, denoted as 4b), resulting in F1-scores of 0.83 \& 0.77 and kappa of 0.8 \& 0.74 for the GAtt class, respectively. Based on the confusion matrix, the best 3b model (denoted as Model-1) classified GAtt trees with TP, FP, FN, and TN of 38, 10, 6, and 78, respectively. 
Second, two six-layer CNNs were trained from scratch with a batch size of 27 for 160 epochs without data augmentation (denoted as Model-2) and 400 epochs with data augmentation (denoted as Model-3). 
For the 3b model and 4b model, the F1-scores were 0.77 and 0.68 without data augmentation and 0.77 and 0.72 with data augmentation, respectively. Moreover, the TP, FP, FN, and TN of the best 3b models were 33, 9, 11, and 79 for Model-2, and 34, 10, 10, and 78 for Model-3, respectively.
Third, transfer learning was applied to the classifier of the pre-trained DenseNet-169 model with a batch size of 27 and training for 160 epochs with augmented data. The F1-scores of the 3b model and 4b model were respectively 0.72 and 0.6, while the best 3b model (denoted as Model-4) achieved TP, FP, FN, and TN of 30, 9, 14, and 79. 
Fourth, the performance of the RF algorithm using original spectral bands without data augmentation was compared with the DL-based methods. The 3b model and the 4b model of the RF algorithm respectively had F1-scores of 0.6 and 0.63, while TP, FP, FN, and TN for the best 4b model (as this algorithm tend to benefit from more features) were 28, 17, 16, and 71.
The best performance was accomplished by Model-1, which achieved the fewest misclassifications with accuracy, precision, recall, CE, OE, RB, and IoU metrics of 0.88, 0.79, 0.86, 0.21, 0.14, 0.09, and 0.7, respectively. 
There was the highest misclassification of GAtt trees as healthy trees in Model-2, although data augmentation in Model-3 reduced this inaccuracy from about 13\% to 4\%}. 
 {The Model-4 had the worst performance such that 30\% of GAtt trees were misclassified as YAtt trees. However, all CNN models outperformed the best RF model, which misclassified 37\% of YAtt trees to GAtt trees and 34\% of GAtt trees to YAtt trees. The results showed that CNN models which used RGB bands performed better than those that added red-edge bands. There may be several reasons for these results, including spectral overlaps of this band with disturbance classes associated with spruce trees or overfitting due to the increased number of parameters with a small training dataset (i.e., a total of 672 tree crowns)}. \\
\indent  {In another study \mbox{\cite{Detection_UAV_YOLOs}}, three versions of the YOLO network were compared (w/o data augmentation and w/o image contrast enhancement) to identify four stages of GAtt, YAtt, RAtt, and GRAtt on two external evaluation plots. The YOLO-v2 network was trained for 4500 and 2000 iterations on original and contrast-enhanced images, while the YOLO-v3 and YOLO-v4 were trained for 6500 and 8000 iterations on mentioned images, respectively. The best overall IoU, precision, recall, and mAP were achieved using contrast-enhanced images for the YOLO-v2 network with 0.76, 0.9, 0.96, and 0.92, for the YOLO-v3 with 0.74, 0.96, 0.91. and 0.97, and for the YOLO-v4 with 0.73, 0.95, 0.76, and 0.94, respectively. The YOLO-v2, YOLO-v3, and YOLO-v4 networks predicted 17, 16, and 8 trees out of 12 GAtt trees in the first test plot and 18, 4, and 17 trees out of 10 GAtt trees in the second one, respectively. Overall, the YOLO-v4 network provided the best GAtt detections, with fewer false positives and more true positives. However, the relatively small dataset of this study limited its ability to offer practical recommendations.}

\subsubsection{\textbf{Classical Algorithms for Hyperspectral Analysis:}} \label{sec:Quant_HS} 
 {As with the MS analysis, various classical ML algorithms were used to analyze HS data for the early detection of bark beetle attacks. However, no DL-based approach has yet been studied to analyze HS data for this application.
In \mbox{\cite{SpectEvid_Early_Engelmann}}, the RF algorithm was applied to HS data to find important wavelengths that showed subtle disturbance changes associated with bark beetle infestations. It resulted in eight more appropriate wavelengths to distinguish healthy and GAtt trees being selected after ensuring minimal contributions from other wavelengths. However, the mean difference between the healthy and GAtt trees was quite small (i.e., between 9e-5 and 0.096 nm) despite selecting wavelengths with an accuracy above 80\%.
In another work \mbox{\cite{PineShootBeetle_Hyperspect_LiDAR}}, five damage degrees caused by pine shoot beetles (including severely damaged and dead trees) were estimated by three RF regression models using HS, LiDAR, or both HS \& LiDAR features. The best results were achieved by a combined approach (i.e., both HS and LiDAR) with {$\rm{R^2=0.83}$} (i.e., model goodness) \& {$\rm{RMSE=9.93\%}$} (i.e., average deviation from actual values) compared to the LiDAR approach with {$\rm{R^2=0.69}$} \& {$\rm{RMSE=12.28\%}$} and HS approaches with {$\rm{R^2=0.67}$} \& {$\rm{RMSE=15.87\%}$}. However, all models failed to accurately estimate slightly infected trees with an accuracy of 66\% to 69\% primarily due to the uncertainty of HS feature extraction, canopy reflectance change during crown delineation, and exclusion of shaded pixels that might contain damaged trees. 
The RF classifier was also used in \mbox{\cite{MultiTemp_HyperMultispect_NorwaySpruce}} to separate healthy, GAtt, and root-rot classes from HS and MS data captured from August to October with an overall accuracy of 40\%-55\% (kappa of 0.3) and 40\%-50\% (kappa of 0.23), respectively. From 28 GAtt trees, the confusion matrix indicated that 18 trees were correctly classified, but eight trees were misclassified as healthy trees and two trees to the root-rot class. In addition, it incorrectly identified five trees to the GAtt and four trees to the root-rot class among 28 healthy trees.
Accordingly, the results were just slightly better than random classification, indicating the models failed to detect GAtt.
In \mbox{\cite{Mapping_Boreal_HealthStatus}}, the RF algorithm was employed to classify tree crowns into healthy, GAtt, and dead classes based on drone and G-LiHT spectral and structural data. The spectral-only and structure-only models using G-LiHT data achieved overall accuracies of 62\% \& 77\% and kappa of 0.42 \& 0.64 compared to the drone-data models, which achieved overall accuracies of 55\% \& 75\% and kappa of 0.29 \& 0.59, respectively. The best results were achieved by the fusion models (i.e., fused spectral and structural data from drone and G-LiHT) with a similar overall accuracy of 78\% for both models and kappa coefficients of 0.64 and 0.65 for the drone fusion model and the G-LiHT fusion model, respectively. However, the discrimination between healthy and GAtt crowns was confusing for all models. For instance, the precision and recall values of the drone spectral-only model for healthy crowns were 32\% and 33\%, respectively, while 68\% and 54\% for GAtt crowns. Although the approach sought to address the difficulty of GAtt detection with only spectral data by fusion models, it was concluded that merely a modest degree of accuracy could be achieved for detecting the GAtt phase due to the confusion among crown health conditions and imbalance in data distribution. \\
\indent Besides the wide applicability of the RF algorithm, other ML methods were also employed for HS analysis of the GAtt detection. 
In \mbox{\cite{HyperRS_MPB_Emphasis_Previsual}}, the separability of continuum removal spectra for healthy, GAtt, and RAtt clustering using shape metrics was qualitatively evaluated by the NN algorithm. Also, the SAM classifier was used for further objective evaluation of the accuracy of the results. Using a small dataset, the best results detected {{37}} correctly classified GAtt trees and six incorrectly classified healthy trees using pigment features. In addition, the best results using water absorption features identified 28 correctly classified GAtt trees and five incorrectly classified trees in the healthy and RAtt classes. However, the spectral reflectance effects were primarily related to decreased nutrient and water delivery to foliage.
The genetic algorithm was used in \mbox{\cite{AngularVegIndex_Spectroscopy_Bavarian}} to select the most stable features, which were then used in the SVM algorithm to classify GAtt trees with an overall accuracy of 70.2\%. However, the classification accuracy was not confirmed when the experiment was extended to the full image extent. In this study, there was confusion between GAtt trees and healthy \& medium-damaged trees due to the lack of field data.
With similar purposes, the genetic algorithm and the SVM classifier were employed in \mbox{\cite{Assess_Hyperspec_Mortality}} to search relevant wavelengths, prune the high-dimensionality of feature space, and classify bark beetle attack stages. While the overall accuracy and kappa coefficient were respectively around 78.4\%-82.9\% and 0.73-0.79, the results were not applicable for operational use due to the confusion of the GAtt phase with other classes and lack of field data.
Lastly, the vitality status of five classes (see Table~\mbox{\ref{tab_ML_all}}) from HyMap spectral data was classified using the ID3 decision tree algorithm in \mbox{\cite{Forecast_Vital_Hyperspect}} with an overall accuracy of 51.56\% and a kappa of 0.395. The results showed that the lowest class recall (i.e., 28.12\%) was for the possible GAtt class, in which needles were green with reduced vitality and without visibility of bark beetle attack. Additionally, the overall classification accuracy between possible GAtt trees and healthy trees was about 64\% because their spectral signatures overlapped considerably.}

\subsubsection{\textbf{Discussion:}} \label{sec:Disc_ML_Exp}
 {For the MS analyses, the RF algorithm provides more effective performance and transferability than other classical ML methods due to its robustness to noise and outliers in the data, reduced risk of overfitting, feature importance measurement, handling of non-linear relationships between features and targets, and scalability to handle large datasets. However, a comparative analysis of the results indicated that the effectiveness of GAtt detection is quite limited. The overall accuracy is less than 80\% and cannot be generalized to other study sites since the predictions were mainly achieved by overfitted models on limited training data. 
The main challenges reported include 
i) excessive spectral variability, 
ii) high intra-class variances, 
iii) insufficient spatial/temporal/spectral data resolution, 
iv) limited availability of data samples and ground-truth data, and 
v) the selection of appropriate handcrafted features. 
In general, the performance and spectral separation of classical ML methods can be improved by combining high-resolution representative information from multiple sensors and collecting data throughout the season. However, the lack of large-scale datasets impacts the effectiveness and generalization of models in accurately predicting GAtt trees. In addition, no unique feature set has been found to reliably predict the GAtt phase. Meanwhile, investigating stressed trees has the potential to distinguish between healthy and GAtt trees. \\
\indent It was expected that HS analysis provides a higher level of capability than MS in detecting subtle changes during the GAtt phase as HS data consists of more narrow bands and detailed information. However, classical ML algorithms used for this analysis failed to detect GAtt trees, demonstrating that trained models and handcrafted features were not sufficiently reliable for practical purposes. Some of the main challenges include i) the uncertainty of HS feature extraction, ii) confusion caused by overlapped spectral signatures, iii) imbalanced data, and iv) a lack of field data. 
Moreover, conducting these analyses is more time-consuming and expensive (particularly for large forest areas) and requires more computational power to process the substantial volume of collected data. This makes it less accessible for those with limited resources. Furthermore, the high cost and technical expertise required to collect and process HS data restrict the number of studies conducted using this data, thereby limiting our understanding of early bark beetle attack detection. \\
\indent Recently, DL-based methods have been utilized for the analysis of bark beetle attacks using MS data, demonstrating their superior performance compared to classical ML algorithms across various evaluation metrics. However, these DL methods are still in the beginning stages and hold potential for further advancements. Currently, most investigations focus on RGB-designed models due to the transferability of pre-trained models, as well as the cost-effectiveness and simplicity of RGB imaging. However, more research is required to explore various model architectures, model ensembles, multi-modal fusion of data from multiple sensors, semi-/unsupervised learning with limited data, and additional spectral bands. But, the fact remains that DL-based models generally require large-scale datasets for training and are not human-interpretable, making it difficult to understand the underlying basis for the predictions and decisions made.}
\subsection{Spectral Signatures for Early Detection} \label{sec:RS_SpecSignature}
\begin{figure}[!t]
\centering
\includegraphics[width=0.8\linewidth]{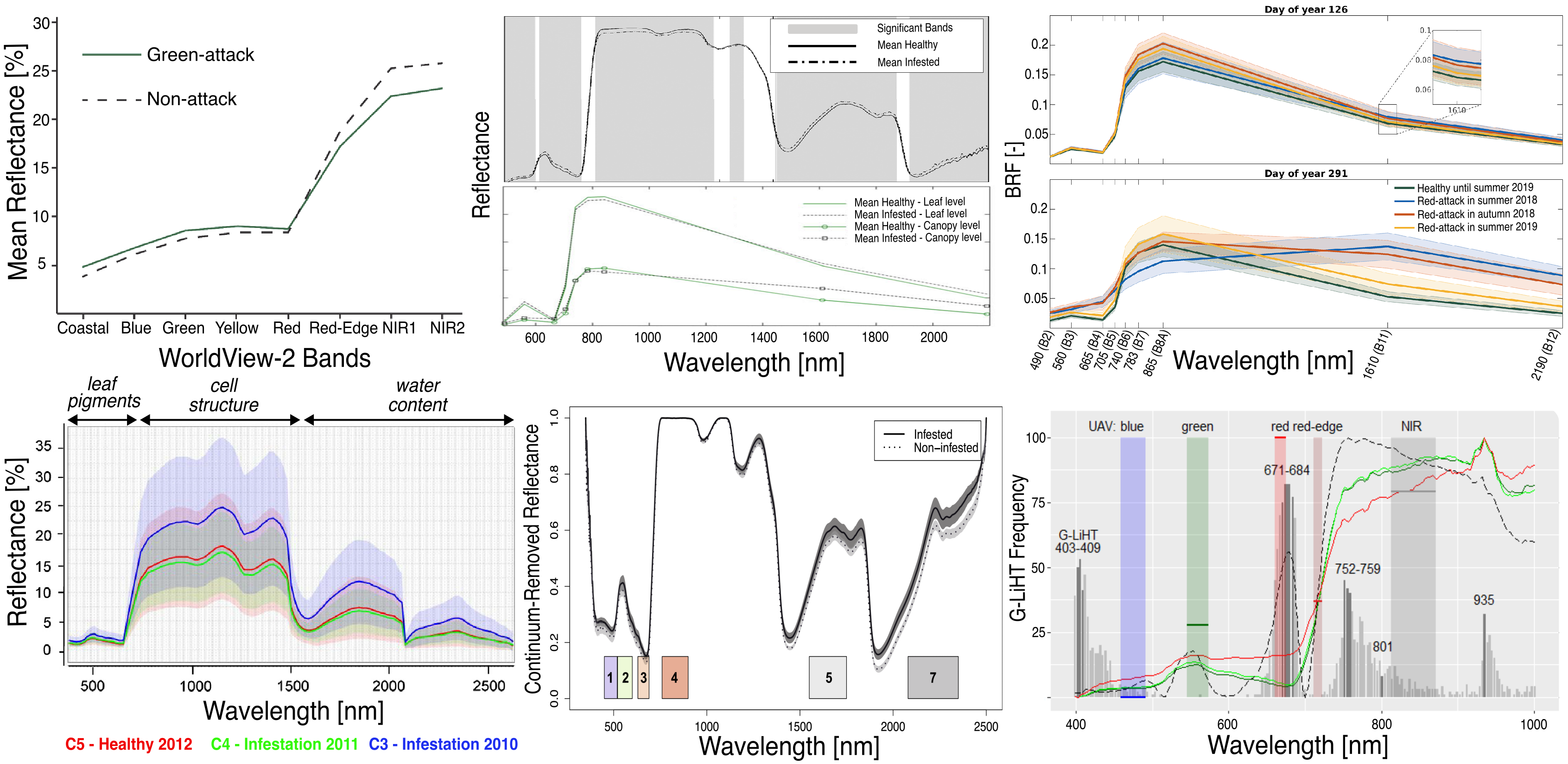}
\vspace{-0.2cm}
\caption{{Examples of spectral signatures based on MS (top row) and HS (bottom row) analyses for early detection of bark beetle attacks. 
(top-left): Comparison of mean reflectance values for healthy and GAtt trees based on WorldView-{{2}} bands} {\cite{MPB_BlackHills_SouthDakota}}{. (top-middle): Comparison of mean reflectance values of healthy and infested foliage (top graph based on field collected data) and canopy (bottom graph based on Sentinel-{{2}} bands)} {\cite{Sentinel2_GA_Landsat8}}{. (top-right): Gradual changes of canopy reflectance from healthy to RAtt phase during the vegetation season based on Sentinel-{{2}} bands} {\cite{EarlyDet_Norway_CentEurope_Sentinel}}{. (bottom-left): Comparison of mean spectral reflectances of HyMAP data for healthy (C{5}) and possible GAtt (C{{3}} \& C{{4}}) trees} {\cite{Forecast_Vital_Hyperspect}}{. (bottom-middle): Comparison of average continuum removal spectral reflectance of healthy and GAtt trees based on Landsat bands (color bars)} {\cite{SpectEvid_Early_Engelmann}}{. (bottom-right): Comparison of normalized spectral responses of healthy, GAtt, and dead trees based on UAV MS bands (five color rectangles) and HS G-LiGHT bands (gray bars)} \cite{Mapping_Boreal_HealthStatus}.}
\vspace{-0.5cm}
\label{fig:signatures} 
\end{figure}

A common approach for detecting bark beetle attacks is to investigate the spectral signatures of GAtt trees (see Fig.~\ref{fig:signatures}). The following conclusions have been drawn regarding which wavelengths were most effective based on the used sensors and the resulting signals captured.

\subsubsection{\textbf{Multispectral Analysis:}} \label{sec:RS_SpecSignature_MS}
The work \cite{EarlyDet_TerraSarX_RapidEye} combined passive and active signals from two satellites to detect GAtt areas, leveraging the complementary role of radar microwave data with optical data for improved detection capabilities. The reflectance patterns related to water stress in leaves/needles had higher responses for GAtt plots than background areas, as the reflectance difference of the NIR band was higher than one for the red band. However, the combination of red and red-edge bands was selected as an important indicator, associated with chlorophyll \textit{a} and \textit{b} to minimize the effects of soil reflectance and leaf biomass. In addition, radar data revealed higher variances in spectral signatures and water-stress signals masked by crown structure and needle mass \& orientation. 
In \cite{EarlyDet_NorwaySpruce_WorldView2}, two summer images (June \& July) of WorldView-{2} were used for spectral analysis of the health status of healthy, GAtt, and dead trees. The analyses were performed using single/combined bands and difference/ratio indices. However, the best results were achieved by combining all eight bands for July-acquired images with significant spectral overlaps \& minor distinctions. In the analysis of individual bands, the yellow band (or red \& green bands) was found helpful in the July (or June) image. 
In another work \cite{MPB_BlackHills_SouthDakota}, two WorldView-{2} images (in April) were collected that pointed to the lower reflectance of the NIR-{1} band as the key to differentiating with non-attacked trees. Accordingly, the NIR bands were linked to the structure of spongy mesophyll in plants and suppressed with developing stress in trees. 
As combined spectral bands could represent foliar biochemical and biophysical properties, the work \cite{Sentinel2_GA_Landsat8} evaluated the effectiveness of various SVIs sensitive to detecting stress-induced variations in chlorophyll content (visible bands), biomass (NIR), and water content (SWIR) from Sentinel-{2} \& Landsat-{8} at the leaf and canopy levels. For Sentinel-{2} data, healthy foliar showed higher chlorophyll and leaf water content than infested one at both levels, which led to differences in visible ({0.52}-{0.685} $\mu m$, NIR ({0.74}-{1.13} $\mu m$), and SWIR ({1.42}-{1.85} $\mu m$ \& {2}-{2.2} $\mu m$) regions. However, due to lower spatial and spectral resolutions, Landsat-{8} data showed limited (or even no) differences at the leaf (or canopy) level. Overall, the red-edge bands and water-related indices of Sentinel-{2} were the most sensitive signatures to the GAtt.

In \cite{Sens_LandSatOLI_TIRS_Foliar}, optical and TIR data from three Landsat-{8} images (May, July, and August for early, advanced, and after infestation) was compared using single bands, SVI, and CST for the detection of stressed canopies using leaf properties (stomatal conductance, chlorophyll fluorescence, and water content). The CST, derived from TIR data, was the best for distinguishing between healthy and GAtt sample plots. The CST revealed surface temperature associated with plant functions (e.g., reduction of leaf water content) and photosynthesis activity (e.g., using chlorophyll fluorescence to examine stress impacts on vegetation physiology). Nevertheless, neither SVIs nor single bands indicated a difference between healthy and GAtt plots. 
The ability of drones to precisely detect GAtt trees with great detail was evaluated in \cite{UAV_PrecDetection_BBInf}. In this study, the efficacy of the NIR band (acquired by a customized CIR camera) for GAtt detection was rejected, and higher spectral differences in the red band as the crucial spectral region were highlighted. Although there were no significant differences in spectral properties between GAtt and healthy trees in June-acquired data, August-acquired data showed significantly lower humidity levels and higher temperatures. It concluded that the SVIs may apply to data acquired at the end of August or the beginning of October as a consequence of the extensive incorrect identification of trees in previous periods. 
In \cite{CrownExt_UASMultispec_MixedForest}, a combination of spectral and elevation metrics was employed to detect individual beetle-infested trees to address overlapping spectral responses (e.g., for different species) in a heterogeneously mixed forest. In spite of this, spectral similarities remained such that no metric could differentiate between all classes of disturbance. 
The effect of data acquisition time on the detection of declined trees (a combination of GAtt and yellowish trees) was examined in \cite{Multispectral_Benefits_Spruce_Decline} for May \& September collected data. Accordingly, fall images revealed a more significant difference in reflectance between declined and healthy trees than spring images because of the higher spectral and phenology variability of individual trees in spring. Also, the result indicated that visible bands were more helpful than red-edge or NIR bands.

In \cite{Monitor_CentrEuro_Validate}, four SVIs were extracted from two Sentinel-{2} images (summer and winter) for damage detection, yet resulted in overlaps between minor damaged trees and those without or moderate damage. Also, Sentinel-{2} time-series data ({14} images from April to November) were examined to determine the pre-visual stages of bark beetle attack within seasonal changes in canopy reflectance. Based on this, the SWIR band (B-{12} with central wavelength at {2.19} $\mu m$) provided the most contribution for discriminating newly infested trees from healthy ones, followed by the red (B-{4} at {0.665} $\mu m$) and red-edge (B-{5} at {0.705} $\mu m$) bands. Considering the importance of tracking seasonal changes to improve accuracy over time, the most sensitive SVIs were calculated by using red (B-{4}), red-edge (B-{5}), and SWIR (B-{11} \& B-{12}) bands. The SWIR, red, and red-edge bands reflect changes in leaf water content \& canopy structure, chlorophyll absorption, and leaf pigment changes \& canopy structure, respectively. Similar to some prior findings (e.g., \cite{UAV_PrecDetection_BBInf}), forest canopy structural effects led to the poor separation of infested trees in NIR bands. 
Lastly, Sentinel-{1} and Sentinel-{2} time-series were used in \cite{EarlyDet_EuropeSpruce_NDRS} for investigating single bands, SVIs, and the combination of multiple bands during vegetation season (April to October). While NIR and red-edge bands achieved the lowest separation and highest uncertainties, the most significant differences were associated with visible (B-{2}, B-{3}, B-{4}) and SWIR (B-{11}, B-{12}) bands before and during the GAtt phase (middle April to June). However, no significant spectral changes were identified during the GAtt phase, which linked the previous differences to the weakness and stress of trees. As well, radar data offered slight effectiveness, making spectral information of greater importance. Based on the analysis, the SVIs appeared to be stable before the attacks, decreased between May and June, and then gradually increased between June and August.
\subsubsection{\textbf{Hyperspectral Analysis:}} \label{sec:RS_SpecSignature_HS}
As mentioned earlier, the analysis of HS data provides the possibility of extracting more detailed information than the MS one.
To develop an index for the detection of bark beetle infections, the study \cite{AngularVegIndex_Spectroscopy_Bavarian} combined HyMap spectral bands in the visible and NIR bands (i.e., {0.455}-{0.986} $\mu m$) based on identified stress-sensitive wavelengths in the forest context. However, the GAtt stands were confused with medium damage and healthy coniferous stands. 
The spectral bands from the HyMap sensor were also analyzed in \cite{Forecast_Vital_Hyperspect} to determine the vitality stages of the spruce stands. Most of the relevant indicators were located in the visible infrared and NIR ({0.45}-{0.89} $\mu m$) related to chlorophyll absorption features and leaf pigments, while a few were identified in the SWIR region (related to the needle water content) between {1.4}-{1.8} $\mu m$. As before, the spectral signatures of GAtt trees and healthy trees overlapped significantly. In pursuit of HyMap data analysis, stress symptoms of bark beetle-infested trees were detected at green peak ($\sim${0.56} $\mu m$), chlorophyll absorption region ({0.68} $\mu m$), and red-edge rise ({0.69} $\mu m$) wavelengths. It was also possible to identify stages of damage based on vegetation vigor information found in the cell structure and water content regions (e.g., {1.076} $\mu m$ and {1.532} $\mu m$). The spectral regions in these bands were fairly similar to those in MS Sentinel-{2}, allowing satellite data to be used for this purpose.
In \cite{HyperRS_MPB_Emphasis_Previsual}, subtle changes in chlorophyll and water levels were considered for detecting bark beetle-induced stress that decreased nutrient and water delivery to foliage. This study focused on the VNIR bands up to {1.3} $\mu m$ since spectral reflectances of healthy and GAtt trees were found not separable with longer wavelengths. The spectral effects in pigment (centered on {0.68} $\mu m$) and foliar water (centered on {0.94} $\mu m$ and {1.25} $\mu m$) regions offered differences between healthy and GAtt stands. However, the visible wavelengths resulted in similar spectra. 
To address the limitation of HS data collection for large forest landscapes (i.e., high cost and time-consuming), ground-based HS measurements at the branch and needle levels were analyzed in \cite{SpectEvid_Early_Engelmann} to adapt for low-cost broadband satellite imagery. As a result, early-stage infested trees exhibited somewhat higher reflectance in the SWIR region ({1.9}-{2.48} $\mu m$) than other wavelengths (e.g., NIR). This higher reflectance was attributed to decreased canopy conductance caused by slowly losing water from infested trees. 

In \cite{PineShootBeetle_Hyperspect_LiDAR}, the combination of HS data with LiDAR data for quantifying the severity of tree crown damage (collected in September) was examined to overcome the limitations of HS imagery in the delineation of crown structure and the spectral variations associated with tree crown shadow. The LiDAR data complemented more sensitive HS data to canopy bio-physiological conditions resulting in spectral differences at the red-edge ({0.66}-{0.75} $\mu m$) and NIR ({0.75}-{1} $\mu m$) regions. 
The study \cite{Mapping_Boreal_HealthStatus} fused high-resolution MS/HS data with structural data captured from aerial and drone platforms and confirmed the crucial role of assessing physiological changes in red and red-edge bands. However, G-LiHT- and drone-captured red-edge bands were referred to as {0.752}-{0.759} $\mu m$ and {0.711}-{0.723} $\mu m$, respectively. The SWIR band information derived from LiDAR data was also useful for stress detection.
In another study \cite{MultiTemp_HyperMultispect_NorwaySpruce}, multitemporal HS data collected from August to October were assessed to detect early infestation stages. This analysis indicated that infested trees have higher spectral responses on the visible bands ({0.5}-{0.68} $\mu m$ linked to the chlorophyll absorption) than healthy trees, with minor spectral differences at the onset of infestation (no noticeable symptoms). 
Simultaneous spectral analysis of ground- \& airborne HS data in \cite{Map_Gyrocopter_HyperField} highlighted the importance of the visible and red-edge spectral bands with sufficient spatial and temporal resolutions. This study focused on the range of {0.4}-{1} $\mu m$, in which airborne (or laboratory) SVIs using red-edge band (or green, blue, and red-edge bands) in the {0.731}-{0.753} $\mu m$ range were helpful for early-stage detection. These ranges indicated the nutrient deficiency in tree crowns in response to bark beetle feeding behavior. 
At last, time-series analysis of HS data revealed that the progression of infestations impacts the reflectance of red to red-edge regions first and NIR region second \cite{Comp_FieldRS_DetBB}. During the progression towards the RAtt, the spectral reflectance of GAtt coniferous in the red \& red-edge bands (chlorophyll absorption) increased while it decreased in the NIR region (canopy structure) compared to the healthy ones \cite{Comp_FieldRS_DetBB}. 
\subsubsection{\textbf{Discussion:}} \label{sec:Disc_SpecSignature}
 {The variety of data collection methods, instruments, sensors, and detection approaches used to identify GAtt trees hinder the ability to draw a unified and conclusive conclusion.
Despite the potential demonstrated by MS analyses in utilizing SWIR, TIR, and visible bands, as well as their combinations, the challenge lies in the limited ability to make clear distinctions between healthy trees and those affected by GAtt due to significant spectral similarities. However, the use of high spatial-spectral resolutions to monitor seasonal changes can yield more dependable findings, characterized by lower variances in MS spectral signatures. 
On the other hand, spectral signatures obtained from HS analyses still face challenges in differentiating minor differences or overlapping signatures between healthy or slightly-damaged trees despite the potential for extracting detailed information. The SWIR, red-edge, and visible bands were identified as the most promising spectral ranges for investigating the GAtt phase of trees affected by bark beetle infestations. However, the similarities observed between healthy and infested trees in many studies make it difficult to rely solely on spectral information for accurate classification. In addition, the exploration of complementary data with sufficient spatial and temporal resolution has been pursued to improve detection capabilities and enhance the differentiation of bark beetle-infested trees from healthy ones. However, the results were not effective enough and should be investigated more with sufficient samples and ground-truth data.}
\subsection{Spectral Vegetation Indices for Early Detection} \label{sec:RS_SVIs}
\subsubsection{\textbf{Multispectral Analysis-based SVIs:}} \label{sec:SVIs_MS}
 {The advantages of using SVIs include enhancing the sensitivity of vegetation detection, reducing atmospheric and background effects, providing standardized measures across different sensors and platforms, and enabling rapid mapping of the extent and severity of infestations}. The most helpful SVIs to detect bark beetle attacks are summarized in Table~\ref{tab_SVIs}.
In \cite{EarlyDet_NorwaySpruce_WorldView2}, the {2}-band ratio (i.e., ${\frac{wavelength1}{wavelength2}}$) and normalized difference (i.e., ${\frac{(wavelength1-wavelength2)}{(wavelength1+wavelength2)}}$) indices calculated from the WorldView-{2} bands achieved comparable results to the combination of its eight spectral bands. While combining red-edge and NIR-{2} bands resulted in the worst performance, the combination of NIR-{1} with the green band (or yellow band) for the June image (or July image) was effective. 
The capability of SVIs derived from Sentinel-{2} and Landsat-{8} bands was examined in \cite{Sentinel2_GA_Landsat8}, which Sentinel-{2} SVIs demonstrating a higher sensitivity than Landsat-{8} ones in detecting stressed canopy states. The red-edge (NDRE-{2} \& NDRE-{3} calculated from {0.705}-{0.783} $\mu m$ region) and water-related (SR-SWIR, NDWI, DWSI, \& LWCI) indices computed from Sentinel-{2} were promising for detecting changes. However, the water-related indices (NDWI, DWSI, and RDI) from Landsat-{8} were partly sensitive (no considerable differences in pigment-dependent indices) given the insufficient spatial and spectral resolutions of imagery. As a result of combining the SWIR (leaf water content) with NIR or visible bands, the resulting indices yielded promising results with both satellite data. But, the SVIs computed from the blue region (Sentinel-{2} \& Landsat-{8}) could not detect spectral variations associated with early infestations.  

\begin{table*}[!bp]
\vspace{-0.3cm}
\rowcolors{2}{}{gray!10}
\caption{Most effective SVIs for early detection of bark beetle attacks. The multispectral and hyperspectral analyses are denoted by MS and HS, and $RX$ denotes reflectance at wavelength $X$ $\mu m$. } 
\vspace{-0.3cm}
\centering 
 \resizebox{\textwidth}{!}{
\begin{tabular}{c c c c c} 
Analysis & Spectral Vegetation Indices  &  Formula  &  Ref.    \\
\hline \hline    

\global\let\CT@@do@color\relax \multirow{14}{*}{\rotatebox[origin=c]{90}{MS}} & \global\let\CT@@do@color\oriCT@@do@color Disease Water Stress Index (DWSI) &  $\frac{\rm NIR+Green}{\rm SWIR1+Red}$ & \cite{Sentinel2_GA_Landsat8,EarlyDet_EuropeSpruce_NDRS}  \protect\\ 

\global\let\CT@@do@color\relax  & \global\let\CT@@do@color\oriCT@@do@color Normalized Difference Red-edge {2} (NDRE-{2}) & $\frac{\rm{NIR}-RedEdge1}{\rm{NIR}+RedEdge1} $ & \cite{Sentinel2_GA_Landsat8}  \protect\\ 
   
\global\let\CT@@do@color\relax  & \global\let\CT@@do@color\oriCT@@do@color Normalized Difference Red-edge {3} (NDRE-{3}) & $\frac{\rm{NIR}-\rm{RedEdge2}}{\rm{NIR + RedEdge2}}$ & \cite{Sentinel2_GA_Landsat8}  \protect\\  

\global\let\CT@@do@color\relax  & \global\let\CT@@do@color\oriCT@@do@color Simple Ratio/Short Wave Infrared (SR-SWIR) & $\frac{\rm SWIR1}{\rm SWIR2}$ & \cite{Sentinel2_GA_Landsat8}  \protect\\  

\global\let\CT@@do@color\relax  & \global\let\CT@@do@color\oriCT@@do@color Normalized Difference Water Index (NDWI or NDVI {0.819}/{1.649}) & $\frac{\rm NIR-SWIR}{\rm NIR+SWIR}$  & \cite{Sentinel2_GA_Landsat8,EarlyDet_Norway_CentEurope_Sentinel,EarlyDet_EuropeSpruce_NDRS}  \protect\\  

\global\let\CT@@do@color\relax  & \global\let\CT@@do@color\oriCT@@do@color Leaf Water Content Index (LWCI) & $\frac{{\rm{log}}(1-(\rm{NIR-SWIR}))}{{-\rm{log}}(1-(\rm{NIR-SWIR}))}$  & \cite{Sentinel2_GA_Landsat8}  \protect\\ 

\global\let\CT@@do@color\relax  & \global\let\CT@@do@color\oriCT@@do@color Ratio Drought Index (RDI) & $\frac{\rm{SWIR2}}{\rm{NIR}}$ & \cite{Sentinel2_GA_Landsat8,EarlyDet_EuropeSpruce_NDRS}  \protect\\ 

\global\let\CT@@do@color\relax  & \global\let\CT@@do@color\oriCT@@do@color Enhanced Normalized Difference Vegetation Index (ENDVI) & $\frac{\rm{(NIR + Green)} - \rm{(2 \times Blue)}}{\rm{(NIR + Green)} + \rm{(2 \times Blue)}}$ & \cite{CrownExt_UASMultispec_MixedForest}  \protect\\ 

\global\let\CT@@do@color\relax  & \global\let\CT@@do@color\oriCT@@do@color Tasseled Cap - Wetness (TCW) & $0.1509 \times {\rm{Blue}}$ + 0.1973 $\times$ {\rm{Green}} + 0.3279 $\times$ {\rm{Red}} + 0.3406 $\times$ {\rm{NIR}} - 0.7112 $\times$ {\rm{SWIR1}} - 0.4572 $\times$ {\rm{SWIR2}} & \cite{EarlyDet_Norway_CentEurope_Sentinel}     \protect\\  


\global\let\CT@@do@color\relax  & \global\let\CT@@do@color\oriCT@@do@color Normalized Distance Red SWIR (NDRS) & $\frac{\rm{DRS}-\rm{DRS^{'}_{min}}}{\rm{DRS^{'}_{max}}+ \rm{DRS^{'}_{min}}}$ & \cite{EarlyDet_EuropeSpruce_NDRS}  \protect\\ 

\global\let\CT@@do@color\relax  & \global\let\CT@@do@color\oriCT@@do@color Distance Red SWIR (DRS) $\dagger$ & $\rm{\sqrt{(Red)^2 + (SWIR)^2}}$ & \cite{EarlyDet_EuropeSpruce_NDRS}   \protect\\ 
 
\hdashline 

\global\let\CT@@do@color\relax \multirow{22}{*}{\rotatebox[origin=c]{90}{HS}} & \global\let\CT@@do@color\oriCT@@do@color Moisture Stress Index (MSI) & $\frac{\rm R1.599}{\rm R0.819} $ & \cite{Forecast_Vital_Hyperspect}  \protect\\ 

\global\let\CT@@do@color\relax  & \global\let\CT@@do@color\oriCT@@do@color Carotinoid Reflectance Index {1} (CRI-{1}) & $\frac{\rm (1/R0.51)}{\rm (1/R0.55)} $ & \cite{Forecast_Vital_Hyperspect}  \protect\\ 

\global\let\CT@@do@color\relax  & \global\let\CT@@do@color\oriCT@@do@color Anthocyanin Reflectance Index {2} (ARI-{2})  &  $ \rm{R0.8 \times [\frac{1}{R0.55}-\frac{1}{R0.7}]} $ & \cite{Forecast_Vital_Hyperspect}  \protect\\ 
\global\let\CT@@do@color\relax  & \global\let\CT@@do@color\oriCT@@do@color Green Normalized Difference Vegetation Index (GNDVI)  &  $\frac{\rm (R0.78 - R0.55)}{\rm (R0.78 + R0.55)} $ & \cite{Forecast_Vital_Hyperspect}  \protect\\  

\global\let\CT@@do@color\relax  & \global\let\CT@@do@color\oriCT@@do@color Normalized Difference Soil Moisture Index (NSMI)  &  $\frac{\rm (R1.8-R2.119)}{\rm (R1.8+R2.119)} $ & \cite{Forecast_Vital_Hyperspect}  \protect\\ 

\global\let\CT@@do@color\relax  & \global\let\CT@@do@color\oriCT@@do@color Normalized Water Index {2} (NWI-{2})  &  $\frac{\rm (R0.97-R0.85)}{\rm (R0.97+R0.85)} $ & \cite{Forecast_Vital_Hyperspect}  \protect\\ 

\global\let\CT@@do@color\relax  & \global\let\CT@@do@color\oriCT@@do@color Green Optimized Soil Adjusted Vegetation Index (GOSAVI)  &  $\frac{\rm (1+0.16)(R0.8-R0.67)}{\rm (R0.8+R0.67+0.16)} $ & \cite{Forecast_Vital_Hyperspect}  \protect\\ 

\global\let\CT@@do@color\relax  & \global\let\CT@@do@color\oriCT@@do@color Normalized Pigment Chlorophyll Ratio Index (NPCI)  &  $\frac{\rm (R0.68-R0.43)}{\rm (R0.68+R0.43)} $ & \cite{Forecast_Vital_Hyperspect}  \protect\\ 

\global\let\CT@@do@color\relax  & \global\let\CT@@do@color\oriCT@@do@color Transformed Chlorophyll Absorption in Reflectance Index (TCARI)  &  $ \rm{3 \times [(R0.7-R0.67)-0.2 \times (R0.7-R0.55) \frac{R0.7}{R0.67}]} $ & \cite{Forecast_Vital_Hyperspect}  \protect\\ 

\global\let\CT@@do@color\relax  & \global\let\CT@@do@color\oriCT@@do@color Difference Index {1} (DI-{1}) & $ \rm{R0.8-R0.55} $ & \cite{Forecast_Vital_Hyperspect}  \protect\\ 

\global\let\CT@@do@color\relax  & \global\let\CT@@do@color\oriCT@@do@color Red-Green Index (RGI) & $\frac{\rm{Red}}{\rm{Green}}$ & \cite{SpectEvid_Early_Engelmann}  \protect\\  

\global\let\CT@@do@color\relax  & \global\let\CT@@do@color\oriCT@@do@color Water Index (WI) & $\rm{\frac{R{0.9}}{R{0.97}}}$ & \cite{SpectEvid_Early_Engelmann}   \protect\\   

\global\let\CT@@do@color\relax  & \global\let\CT@@do@color\oriCT@@do@color Normalized Difference Photochemical Reflectance Index (PRI) & $\rm{\frac{R{0.531}-R{0.570}}{R{0.531}+R{0.570}}}$  &  \cite{Comp_FieldRS_DetBB}  \protect\\ 
\global\let\CT@@do@color\relax  & \global\let\CT@@do@color\oriCT@@do@color Laboratory Index {3} (LI-{3}) & $\rm{1.5 \times [(R0.724-R0.716)-(R0.716-R0.709)]-2 \times [(R0.549-R0.541)-(R0.49-R0.483)]+ 0.5 \times [(R0.541-R0.534)-(R0.52-R0.512)] }$ & \cite{Map_Gyrocopter_HyperField}  \protect\\   
 
\global\let\CT@@do@color\relax  & \global\let\CT@@do@color\oriCT@@do@color HySpex Index {1} (HI-{1}) & $\rm{\frac{R0.75252-R0.73067}{2}}$ & \cite{Map_Gyrocopter_HyperField}  \protect\\   
 
\global\let\CT@@do@color\relax  & \global\let\CT@@do@color\oriCT@@do@color Red-edge inflection point (REIP) & $0.75 + 0.035 \rm{ \left( \frac{\left(\frac{R{0.665}+R{0.783}}{2} \right)-R{0.705}}{R{0.74}+R{0.705}} \right)}$ & \cite{Comp_FieldRS_DetBB}    \protect\\ 

\global\let\CT@@do@color\relax  & \global\let\CT@@do@color\oriCT@@do@color ANCB index CR ({0.65}–{0.72}) $\ddagger$ & $\frac{1}{2} \sum_{i=1}^{n-1}{(\lambda_{i+1}-\lambda_i)(R_{{\rm{CR}}(\lambda_{i+1})}+ R_{{\rm{CR}}(\lambda_i)})}\frac{1}{{\rm{BD}}_{0.68}} $ &  \cite{Comp_FieldRS_DetBB}    \protect\\  

 
\hdashline 
 
\global\let\CT@@do@color\relax \multirow{2}{*}{\rotatebox[origin=c]{0}{MS \& HS}}   & \global\let\CT@@do@color\oriCT@@do@color Normalized difference Vegetation Index (NDVI or NDVI - NIR or NDVI {0.8}/{0.65})  & $\frac{\rm{NIR} - \rm{Red}}{\rm{NIR} + \rm{Red}}$& \cite{UAV_PrecDetection_BBInf,CrownExt_UASMultispec_MixedForest,EarlyDet_Norway_CentEurope_Sentinel,SpectEvid_Early_Engelmann}   \protect\\ 

\global\let\CT@@do@color\relax  & \global\let\CT@@do@color\oriCT@@do@color Greenness Index (GI) & $\frac{\rm R0.554}{\rm R0.677}$ & \cite{UAV_PrecDetection_BBInf,Comp_FieldRS_DetBB}  \protect\\
 
\hline 

\multicolumn{3}{@{} l}{\scriptsize $\dagger\dagger$ ranges of the DRS values for all spruce pixels in the image } \\ 
\multicolumn{3}{@{} l}{\scriptsize $\ddagger$ the area under the continuum removed (CR) reflectance ({0.65}-{0.72} $\mu m$), normalized by the CR band depth ({0.68} $\mu m$) (RCR($\lambda_{i+1}$) and RCR($\lambda_{i}$) are values of CR reflectance at the $i$ and $i+1$ bands, $\lambda_{i}$ and $\lambda_{i+1}$ are wavelengths of the $j$ and $j+1$ bands, $n$ is number of bands) } 

\end{tabular}
  \label{tab_SVIs}  
} 
\end{table*}

The evaluation of SVIs derived from RGB and NIR images captured by drones highlighted the potential of the Greenness index \cite{UAV_PrecDetection_BBInf}. However, NIR-based SVIs (e.g., SR, NDVI, GNDVI, or GRVI) were less effective. 
In \cite{CrownExt_UASMultispec_MixedForest}, the NDVI and ENDVI discriminated between the healthy and GAtt trees despite their inability to distinguish between pines and YAtt trees. 
The study \cite{Multispectral_Benefits_Spruce_Decline} pointed out the importance of the normalized difference index, which combines red-edge and red wavelengths, in detecting spectral differences in bark beetle symptoms from spring- \& fall-collected images.
In \cite{EarlyDet_Norway_CentEurope_Sentinel}, it was concluded that it is crucial to evaluate seasonal changes in SVIs than rely solely on their static values such that substantial differences appeared in TCW, NDVI {0.8}/{0.65}, and NDVI {0.819}/{1.649} indices employing red and SWIR spectral bands. 
Finally, the analysis of spectral differences over the entire vegetation season in \cite{EarlyDet_EuropeSpruce_NDRS} revealed that SVIs that measured water content were more sensitive to bark beetle attacks than chlorophyll-measured SVIs. Accordingly, the NDRS, NDWI, RDI, and DSWI more reliably detected the effects of GAtt on spectral regions.
\subsubsection{\textbf{Hyperspectral Analysis-based SVIs:}} \label{sec:RS_SVIs_HS}
Useful SVIs to assess the applicability of HS indicators were selected in \cite{Forecast_Vital_Hyperspect} as the MSI, CRI-{1}, GNDVI, ARI-{2}, NWI-{2}, NDVI, NSMI, GOSAVI, NPCI, TCARI, and DI-{1}. However, the performance resulted in a low to moderate differentiation due to a variety of factors, e.g., mixed information (low spatial resolution) or the age of the stands. 
In \cite{SpectEvid_Early_Engelmann}, it was shown that individual HS bands still provided superior performance than SVIs and averaged MS bands. Accordingly, the indices of RGI, WI, and NDVI were helpful indicators based on the differences in foliage between non-infested and GAtt trees.
The study \cite{Map_Gyrocopter_HyperField} analyzed laboratory \& HS indices respectively derived from field and airborne measurements to develop new sensitive SVIs. Among defined indices, the LI-{3} (combined green, blue, and red edge bands associated with photosynthetically active pigments) was the most effective for early detection of bark beetle infestation at the needle level, while its performance was limited on airborne-acquired data. 
In another study \cite{Map_Gyrocopter_HyperField}, the airborne data-based HI-{1} index (using the red-edge region) was successful at the pixel level, transferable to another forest stand, and long-term validated in the presence of mixed spectral signatures. 
Lastly, the sequence and timing of SVIs in distinguishing GAtt trees from healthy ones \cite{Comp_FieldRS_DetBB} showed that REIP, PRI, and ANCB were capable of detecting infestation after {23} days and GI and NDVI after {56} days. These delays were attributed to the consideration of canopy structural changes instead of subtle changes in chlorophyll content, such that the delay was longer for SVIs derived from visible and NIR regions than those using red-edge and green peak wavelengths.
\subsubsection{\textbf{Discussion:}} \label{sec:RS_SVIs_MS}
 {The use of SVIs derived from the spectral measurements of specific wavelength regions can provide valuable insights into the physiological \& biochemical characteristics, health conditions, and stress levels of bark beetle-attacked trees. The results from MS and HS analyses further support the previous findings, emphasizing the crucial role of SWIR, red-edge, and visible bands in the detection of GAtt trees. Although individual bands may often provide more accurate performance, the most promising SVIs were identified as those related to water-related indices, which reflect seasonal changes and have the high spatial resolution to prevent mixed information. However, it is worth noting that the identified SVIs have still limitations in terms of their ability to detect attacks at an early stage, indicating the need for further research and development to improve their effectiveness.}

\section{ {Summary of Current Issues and Future Research Directions} } \label{sec:FutureDirections}
 {While recent advances in RS and ML/DL have shown great potential in improving early detection of bark beetle-attacked trees, these current technologies cannot be considered a complete replacement for traditional methods at present, as the detection still remains a problem. Therefore, this section of our review paper emphasizes the limitations of RS and ML techniques, while also recognizing their ability to provide valuable knowledge and insights that can effectively enhance the efficiency, scalability, and accuracy of future detection methods.
Although many efforts have been made in various types of forest conditions and scenarios, there are still numerous issues and challenges that should be addressed. The major challenge lies in the uncertainties pertaining to various aspects, including (i) the possibility of tree stress caused by other factors (e.g., drought or different species of insect attacks), (ii) non-optimal imagery resolutions, (iii) ineffective combination of imagery systems, (iv) low accuracy due to unbalanced and limited ground-truth samples, (v) ambiguous spectral signatures or signals, (vi) variable tree/forest traits (e.g., physiological, biophysical, biochemical, functional, structural, etc.), (vii) varying attack phase timing, (viii) limited model generalization or robustness, (ix) inadequate assessment of tree/forest damage, (x) inconsistent multi-date imagery, (xi) insufficient study areas, (xii) potential information loss following pre-processing of raw data, (xiii) handcrafted feature/model selection, (xiv) deficient model training, and (xv) inefficacy of methods for detecting trees (or species). The following are the challenges, considerations, and future directions from different perspectives.
Hence, it is imperative to prioritize ongoing research and development in RS and ML/DL techniques, given the limited effectiveness of current methods in this application.}

\subsection{Bark Beetle-Host Tree Interactions} \label{sec:FutDir_BBeetle}
There is still complexity in understanding bark beetle and host interactions when considering interactions among plant hormones, plant primary metabolites (e.g., carbohydrates) \& defense metabolites, environment, as well as when bark beetles are combined with other factors (e.g., biotic agents or drought) \cite{Nadir_Zanganeh_2021}. 
Bark beetle attacks encompass a broad range of underlying causes (e.g., climate change) and effects (e.g., physiological disruption in host trees) and how fast symptoms develop can be influenced by various factors, including thermal condition, tree vigor, and beetle density. The transition among attack phases and spatial dispersion of infestations also depends on various factors (e.g., temperature and precipitation, tree age, distance from the nearest forest edge, number of attacking beetles, tree vigor, wind speed and direction, and soil moisture) that may change from tree to tree even at the same locations. 
 {Early detection of infested trees before the emergence of bark beetle broods is critical for the development of effective pest management activities. For instance, in Western North America, the majority of annual allowable harvesting plus the sanitary felling of bark beetle-infested trees takes place during the winter months when the ground is frozen. These activities would significantly reduce the bark beetle populations as long as early detection is successful. Early detection is particularly important at the fringes of large bark beetle outbreaks where emerging beetles can infest healthy trees over a short distance versus those requiring long-distance dispersal to locate a suitable host tree. These trees are considered relatively easy to detect (due to a higher infestation probability) and should be prioritized in sanitation harvesting.}
Nonetheless, the distribution and timing of bark beetle attacks depend on a variety of factors (e.g., forest structure, bark beetle population density, and environmental conditions) such that the specific timing of attack symptoms may not be feasible. 
For example, spruce trees attacked by spruce beetles (\emph{Dendroctonus rufipennis}) can take between one and three years after they were initially attacked to fade to YAtt. Moreover, in spruce trees attacked by \emph{D. rufipennis}, the transition to RAtt can be equally variable. 
 {Further, it is crucial to detect European spruce bark beetle (\emph{Ips typographus}) attacks as early as possible before they complete their development and emerge from the parental tree}. The first generation of these beetles is typically present from May to July, while the transition from green to yellow stages occurs in August \cite{EarlyDet_EuropeSpruce_NDRS}.

For challenging GAtt detection, various infestation indicators have been studied to quantify subtle variations in forest traits, such as physiological \& biochemical changes and tree structure/stress levels. It is common for these indicators to examine changes in leaf and canopy pigments (e.g., chlorophyll absorption), canopy/cell structure, and water content in attacked trees. 
Despite exploring various characteristics, there is no unanimous agreement regarding which key initial effects are specifically related to bark beetle attacks and excluded other possibilities. For instance, there are complications due to the possibility of misinterpreting beetle-induced changes with those triggered by different agents or conditions (e.g., drought).
 {Hence, field surveys by visual inspection or using sniffer dogs \mbox{\cite{Sniffer_dogs}} remain the most reliable method for identifying recently infested trees. However, they have several limitations, such as being time-consuming, unsuitable for large areas, accessibility, and weather dependent (ineffective after rain or heavy wind)}.
To conclude, early detection of bark beetle attacks is extremely challenging due to variations in area characteristics, beetle species, beetle \& host tree interactions, and other unexplained natural variabilities such as tree genetics in the rate of phase change.
It is, therefore, crucial to integrate advanced knowledge of bark beetle biology and tree physiology to develop automated systems required for the early detection of bark beetle attacks.

\subsection{Remote Sensing Perspective} \label{sec:FutDir_RS}
While RS platforms have proven successful in detecting the late stages of tree mortality caused by bark beetles, the challenge of early detection still remains unresolved. The performance of early detection could be affected by various characteristics of instruments and operations, such as environmental \& data collection conditions.
The selection of RS platforms, sensors, spectral bands, and resolutions (spatial, spectral, \& temporal) can significantly impact the effectiveness of the detection. 
 {Multi-date satellite imagery can provide a desirable tool for characterizing bark beetle outbreaks due to several advantages, such as long-term data coverage, global accessibility, availability of archived image data, and the ability to quantify the spatio-temporal dynamics of a disturbance, including mortality initiation, extension, and forest recovery}.
Although commercial satellite imagery has advanced significantly in recent years, investigations for GAtt detection revealed discrepancies caused by insufficient imagery resolution or effects of clouds \& cloud shadows. 
Meanwhile, small and relatively inexpensive drones can flexibly investigate individual attacked trees based on their structure and composition.
Drone (or aerial) imagery covers more limited areas (i.e., local to regional scale) than satellite imagery, while the mosaicking of the acquired images can form larger synthetic images that are used for analyzing wider areas. Drones also require accurate image acquisition planning and processing, and radiometric inconsistencies of mosaicked images (derived from flight paths) should be eliminated by appropriate calibration approaches to alleviate complex analyses. 

Nowadays, it is evident that early detection of bark beetle attacks requires fine enough spatial resolution with multi-date temporal resolutions. According to the critical role of spatial resolution, fine-resolution imagery (e.g., metric or sub-meter resolution (\textit{ground sample distance} (GSD) $<${1})) is required to investigate individual trees. In addition, the higher frequency of data acquisition (better temporal resolution with time series (e.g., {5}-{10} acquisitions per year)) can facilitate earlier detection of infested trees due to the different change rates depending on bark beetle life cycle dynamics, site conditions, and weather dependency of tree vitality.
Moreover, various analyses have been conducted using low-cost RGB, MS, or HS sensors either individually or in combination. While HS sensors with fine spectral and spatial resolutions are ideally suited to detect subtle changes in response to bark beetle disturbances, they are expensive and time-consuming for analysis. Alternatively, MS and TIR sensors are more readily accessible and provide sufficient coverage to detect changes in large areas. According to the MS \& HS data analyses, the canopy reflectance/temperature/structure, plant functions, and photosynthesis activities significantly impact stress-induced variations following bark beetle attacks in water content (SWIR), surface temperature (thermal bands), and chlorophyll content (visible bands). Although specific stress symptoms and geographic conditions may determine potentially helpful spectral signatures and indices, detecting GAtt trees is still an open problem due to the highly variable spectro-temporal signatures in forests (e.g., uncertainties caused by diverse growing environments and stand properties). For instance, spectral separability may exist between healthy and stressed trees before an attack and does not noticeably increase during the early stage of infestation. 

To sum up, the biological, technical, and logistical constraints associated with the flight activity of bark beetles, the colonization time-frames, and the duration of showing distinct symptoms for infested trees limit the effectiveness of RS-based systems. 
While research has been conducted into combining imagery systems, there is limited evidence that doing so is efficacious for early detection. As well, different instruments, operations, and environmental conditions affect the conclusions drawn about the most suitable imaging systems. Moreover, pre-processing and transformations of the collected data are performed differently and may result in the possible loss of information.
The existing RS-based methods delay detecting infested trees compared to traditional field surveys, partly due to the irregular data acquisition and lack of fine spatial-spectral resolution data. 
Additionally, the spectral resolution for this purpose is unspecified, and there is indecisiveness in finding spectral signatures. Hence, it is still necessary to take into account the progress of RS capabilities and explore advanced systems in order to overcome the challenges of early detection. 

\subsection{ {Machine Learning Perspective}} \label{sec:FutDir_ML}
Early detection of bark beetle attacks is inherently challenging because of the small unbalanced datasets and the limited number of representative samples, leading to low accuracy and robustness of ML/DL models. These scenarios with scarce samples are primarily the result of small study areas, technical difficulties with data collection, and the requirements to annotate data accurately. The manual delineation and labeling of tree crowns represent one of the most arduous and time-consuming tasks that, to date, have not been accomplished to provide a large-scale dataset for GAtt detection. As a result, the methods could not be effectively compared nor applied to other study sites (i.e., limited generalization). 
The preprocessing procedures and data resolution may also significantly affect the performance of ML methods. Prior to applying ML algorithms, it is essential to preprocess the data to enhance its quality and usability for subsequent analyses. This involves various techniques such as data normalization, image enhancement, filtering, dimensionality reduction, and feature extraction. For instance, masking non-tree components or contrast enhancement can effectively improve the accuracy of algorithms according to their impact on all subsequent analyses. Also, dimension reduction can provide a limited number of inputs that facilitates output interpretation. However, these procedures are required careful considerations as they may adversely impact the collected information and the performance of systems. For example, dimension reduction can result in the loss of information, potentially sacrificing fine-grained details and intricate relationships inherent in the raw data. Also,  {high spatial resolution may reduce classification effectiveness due to the increased spectral intra-class variability (i.e., increasing spectral correlations or dependence between different bands) \mbox{\cite{SpatialSpectralVariance1,SpatialSpectralVariance2,EvalPoten_MultiSpect_MultipleStage}}; However, the insufficient image resolution (coarser than the tree crown) will cause shadow effects that dramatically impact the accuracy.}

As discussed in this paper, various ML methods have been used to analyze GAtt detection, exploiting various inputs (e.g., spectral, PAN, pan-sharpened images, time-series, and radar data) and features (e.g., spectral, textural, and topographic). 
These methods are either pixel- or object-based, considering the use of individual pixels or a set of pixels as shapes for analyses. Despite the merits of object-based methods (e.g., facilitating interpretation), pixel-based methods could achieve higher accuracy due to creating more homogeneous areas for independent classes. However, pixel-based methods may suffer from spectral variations (e.g., arising from branches and shadows) within tree crowns of mixed forests \cite{CrownExt_UASMultispec_MixedForest}. 
The selection of features (e.g., spectral bands and SVIs) and their interpretation is another aspect that should be considered when using classical algorithms. These algorithms demand expert knowledge and professional experience to select the best features related to structural and biochemical tree properties and corresponding spectral characteristics. However, the feature selection process is time-consuming and prone to human bias. Hence, these methods generally tend to employ different types of features to provide more reliable results. However, different scenarios and interpretations make selecting the most suitable set of features quite challenging. In addition, classical methods are typically applied following other classical tree detection methodologies (e.g., classical image processing-based methods), which can considerably impact the accuracy and limit the performance to specific scenarios.

The well-known RF is the most commonly used algorithm among classical approaches to detect bark beetle attacks. This algorithm ensembles multiple decision trees and randomly selects training samples and variables. 
The advantages include no assumptions about data distribution and collinearity, no common covariance requirement for classes, sample-independent accuracy, and variable importance measurements. However, it requires careful sampling design, is not generalizable to different study areas, and may become inefficient with a significant number of trees. 
In addition, the presence of a limited number of samples, handcrafted features, and the high variability of spectral information in mixed tree forests can severely impact its accuracy. 
 {Despite the majority of bark beetle attack research focusing on classical algorithms or even traditional analyses (e.g., threshold-based methods), remarkable advancements in DL have emerged in recent years that have yet to be explored for this application. 
DL-based methods offer significant advancements, including: 
i) feature learning (i.e., automatically learning meaningful features from raw data), 
ii) handling big data (i.e., enabling better generalization by learning from large-scale datasets), 
iii) non-linear pattern learning (i.e., capturing intricate non-linear underlying relationships in the data), 
iv) end-to-end learning (i.e., optimizing all model layers simultaneously and learning complex mappings from raw input to output), 
v) transfer learning (i.e., leveraging pre-trained models, enabling the transfer of knowledge and improving performance even with limited labeled data.), 
vi) handling unstructured data (i.e., directly learning from unstructured data modalities without extensive preprocessing), 
vii) time series analysis (i.e., capturing temporal dependencies and handling irregular time intervals), 
viii) better generalization (i.e., learning robust representations of relevant information), 
ix) handling complex data types (i.e., allowing for modeling relationships, e.g., in multimodal data), 
x) few-/zero-shot learning (i.e., quickly adapting models on a small number of data or recognizing/classifying unseen classes), and 
xi) scalable architectures (i.e., allowing scaling up the parameters of neural networks).
Recent DL-based methods using end-to-end trained models have demonstrated superior performance compared to transfer learning of models and classical methods (e.g., RF). However, despite the advancements, challenges persist in detecting subtle changes in spectral responses, which has led to misclassifications of early-attacked trees.
Hence, the effectiveness of these methods still requires further investigations considering the limitations, such as lacking data, types of collected information, or varying definitions of the problem}.
In this regard, it is imperative to consider the most recent advancements in computer vision and ML/DL to overcome the challenges.
\section{Conclusion} \label{sec:Conclusion}
 {This paper provides a comprehensive and systematic review of existing methods used in the early detection of bark beetle-induced tree mortality from three crucial perspectives: bark beetle \& host interactions, RS, and ML/DL. We explore the application of classical ML algorithms and DL-based methods, utilizing data collected from diverse RS systems. This review involves in-depth parsing of the literature using multi- or hyper-spectral analysis and distills valuable knowledge on various aspects, including bark beetle species \& attack phases, host trees, study regions, RS platforms \& sensors, spectral/spatial/temporal resolutions, spectral signatures, helpful SVIs, ML approaches, learning schemes, task categories, models, algorithms, classes/clusters, features, and DL networks \& architectures.
In addition, it compares the outcomes of current approaches to provide a detailed summary of the strengths and weaknesses of the employed systems and methods. The results demonstrate early detection of bark beetle attacks remains an open problem for which recent approaches offer limited success in addressing its challenges. Hence, our review emphasizes the principal challenges and opportunities from three distinct perspectives, facilitating a comprehensive understanding of the current research landscape and providing valuable guidance for future research endeavors.} 
\section{Acknowledgments}
The authors would like to thank Dr.~Guillermo~Castilla from Canadian Forest Service, Natural Resources Canada for reviewing this work and providing valuable comments and suggestions.
\bibliography{0_references.bib}

\begin{thebibliography}{102}
\providecommand{\natexlab}[1]{#1}
\providecommand{\url}[1]{\texttt{#1}}
\expandafter\ifx\csname urlstyle\endcsname\relax
  \providecommand{\doi}[1]{doi: #1}\else
  \providecommand{\doi}{doi: \begingroup \urlstyle{rm}\Url}\fi

\bibitem[Abdollahnejad and Panagiotidis(2020)]{Class_HealthStat_UAS_Multispect}
Azadeh Abdollahnejad and Dimitrios Panagiotidis.
\newblock Tree species classification and health status assessment for a mixed
  broadleaf-conifer forest with uas multispectral imaging.
\newblock \emph{Remote Sensing}, 12\penalty0 (22), 2020.

\bibitem[Abdullah et~al.(2019{\natexlab{a}})Abdullah, Darvishzadeh, Skidmore,
  and Heurich]{Sens_LandSatOLI_TIRS_Foliar}
Haidi Abdullah, Roshanak Darvishzadeh, Andrew~K. Skidmore, and Marco Heurich.
\newblock Sensitivity of {Landsat-8 OLI and TIRS} data to foliar properties of
  early stage bark beetle ({Ips typographus, L.}) infestation.
\newblock \emph{Remote Sensing}, 11\penalty0 (4), 2019{\natexlab{a}}.

\bibitem[Abdullah et~al.(2019{\natexlab{b}})Abdullah, Skidmore, Darvishzadeh,
  and Heurich]{Sentinel2_GA_Landsat8}
Haidi Abdullah, Andrew~K. Skidmore, Roshanak Darvishzadeh, and Marco Heurich.
\newblock {Sentinel-2} accurately maps green-attack stage of {European} spruce
  bark beetle ({Ips typographus, L.}) compared with {Landsat-8}.
\newblock \emph{Remote Sensing in Ecology and Conservation}, 5\penalty0
  (1):\penalty0 87--106, 2019{\natexlab{b}}.

\bibitem[Alvarez-Vanhard et~al.(2021)Alvarez-Vanhard, Corpetti, and
  Houet]{UAV_Sat_Review}
Emilien Alvarez-Vanhard, Thomas Corpetti, and Thomas Houet.
\newblock {UAV} \& satellite synergies for optical remote sensing applications:
  A literature review.
\newblock \emph{Science of Remote Sensing}, 3:\penalty0 100019, 2021.

\bibitem[Bright et~al.(2013)Bright, Hudak, McGaughey, Andersen, and
  Negrón]{LiveDead_Basal_LiDAR_Journal}
Benjamin~C. Bright, Andrew~T. Hudak, Robert McGaughey, Hans-Erik Andersen, and
  José Negrón.
\newblock Predicting live and dead tree basal area of bark beetle affected
  forests from discrete-return {LiDAR}.
\newblock \emph{Canadian Journal of Remote Sensing}, 39\penalty0
  (sup1):\penalty0 S99--S111, 2013.

\bibitem[Bárta et~al.(2021)Bárta, Lukeš, and
  Homolová]{EarlyDet_Norway_CentEurope_Sentinel}
Vojtěch Bárta, Petr Lukeš, and Lucie Homolová.
\newblock Early detection of bark beetle infestation in {Norway} spruce forests
  of {Central Europe using Sentinel-2}.
\newblock \emph{International Journal of Applied Earth Observation and
  Geoinformation}, 100:\penalty0 102335, 2021.

\bibitem[Bárta et~al.(2022)Bárta, Hanuš, Dobrovolný, and
  Homolová]{Comp_FieldRS_DetBB}
Vojtěch Bárta, Jan Hanuš, Lumír Dobrovolný, and Lucie Homolová.
\newblock Comparison of field survey and remote sensing techniques for
  detection of bark beetle-infested trees.
\newblock \emph{Forest Ecology and Management}, 506:\penalty0 119984, 2022.

\bibitem[Cessna et~al.(2021)Cessna, Alonzo, Foster, and
  Cook]{Mapping_Boreal_HealthStatus}
Janice Cessna, Michael~G. Alonzo, Adrianna~C. Foster, and Bruce~D. Cook.
\newblock Mapping boreal forest spruce beetle health status at the individual
  crown scale using fused spectral and structural data.
\newblock \emph{Forests}, 12\penalty0 (9), 2021.

\bibitem[Chiang et~al.(2020)Chiang, Barnes, Angelov, and
  Jiang]{ForestHealth_Aerial_DL}
Chia-Yen Chiang, Chloe Barnes, Plamen Angelov, and Richard Jiang.
\newblock Deep learning-based automated forest health diagnosis from aerial
  images.
\newblock \emph{IEEE Access}, 8:\penalty0 144064--144076, 2020.

\bibitem[Coops et~al.(2006)Coops, Wulder, and
  White]{IntegRS_AncillaryData_CharMPB}
Nicholas~C. Coops, Michael~A. Wulder, and Joanne~C. White.
\newblock Integrating remotely sensed and ancillary data sources to
  characterize a mountain pine beetle infestation.
\newblock \emph{Remote Sensing of Environment}, 105\penalty0 (2):\penalty0
  83--97, 2006.

\bibitem[Deng et~al.(2009)Deng, Dong, Socher, Li, Li, and Fei-Fei]{ImageNet}
Jia Deng, Wei Dong, Richard Socher, Li-Jia Li, Kai Li, and Li~Fei-Fei.
\newblock {ImageNet}: A large-scale hierarchical image database.
\newblock In \emph{Proc. CVPR}, pages 248--255, 2009.

\bibitem[Dennison et~al.(2010)Dennison, Brunelle, and
  Carter]{AssessCanopy_MPB_GeoEye_Sat}
Philip~E. Dennison, Andrea~R. Brunelle, and Vachel~A. Carter.
\newblock Assessing canopy mortality during a mountain pine beetle outbreak
  using {GeoEye-1} high spatial resolution satellite data.
\newblock \emph{Remote Sensing of Environment}, 114\penalty0 (11):\penalty0
  2431--2435, 2010.

\bibitem[Drauschke et~al.(2014)Drauschke, Bartelsen, and
  Reidelstürz]{Toward_UAV_ForestMonitor}
M.~Drauschke, J.~Bartelsen, and P.~Reidelstürz.
\newblock Towards {UAV}-based forest monitoring.
\newblock In \emph{Proc. Workshop on UAV-basaed Remote Sensing Methods for
  Monitoring Vegetation}, pages 21--32, 2014.

\bibitem[Duarte et~al.(2022)Duarte, Borralho, Cabral, and
  Caetano]{Review_Advances_ForestInsect_UAV}
André Duarte, Nuno Borralho, Pedro Cabral, and Mário Caetano.
\newblock Recent advances in forest insect pests and diseases monitoring using
  {UAV}-based data: A systematic review.
\newblock \emph{Forests}, 13\penalty0 (6), 2022.

\bibitem[Duračiová et~al.(2020)Duračiová, Muňko, Barka, Koreň,
  Resnerová, Holuša, Blaženec, Potterf, and
  Jakuš]{Sat_DecisionSupport_TANABBO}
Renata Duračiová, Milan Muňko, Ivan Barka, Milan Koreň, Karolina
  Resnerová, Jaroslav Holuša, Miroslav Blaženec, Mária Potterf, and
  Rastislav Jakuš.
\newblock A bark beetle infestation predictive model based on satellite data in
  the frame of decision support system {TANABBO}.
\newblock \emph{iForest - Biogeosciences and Forestry}, 13\penalty0
  (3):\penalty0 215--223, 2020.

\bibitem[Ecke et~al.(2022)Ecke, Dempewolf, Frey, Schwaller, Endres, Klemmt,
  Tiede, and Seifert]{Review_UAV_ForestHealth}
Simon Ecke, Jan Dempewolf, Julian Frey, Andreas Schwaller, Ewald Endres,
  Hans-Joachim Klemmt, Dirk Tiede, and Thomas Seifert.
\newblock {UAV}-based forest health monitoring: A systematic review.
\newblock \emph{Remote Sensing}, 14\penalty0 (13), 2022.

\bibitem[Edburg et~al.(2012)Edburg, Hicke, Brooks, Pendall, Ewers, Norton,
  Gochis, Gutmann, and Meddens]{BeetleImpacts}
Steven~L Edburg, Jeffrey~A Hicke, Paul~D Brooks, Elise~G Pendall, Brent~E
  Ewers, Urszula Norton, David Gochis, Ethan~D Gutmann, and Arjan~JH Meddens.
\newblock Cascading impacts of bark beetle-caused tree mortality on coupled
  biogeophysical and biogeochemical processes.
\newblock \emph{Frontiers in Ecology and the Environment}, 10\penalty0
  (8):\penalty0 416--424, 2012.

\bibitem[Einzmann et~al.(2021)Einzmann, Atzberger, Pinnel, Glas, Böck, Seitz,
  and Immitzer]{Early_Hyper_Results_Bavaria}
Kathrin Einzmann, Clement Atzberger, Nicole Pinnel, Christina Glas, Sebastian
  Böck, Rudolf Seitz, and Markus Immitzer.
\newblock Early detection of spruce vitality loss with hyperspectral data:
  Results of an experimental study in {Bavaria, Germany}.
\newblock \emph{Remote Sensing of Environment}, 266:\penalty0 112676, 2021.

\bibitem[Erbilgin et~al.(2021)Erbilgin, Zanganeh, Klutsch, Chen, Zhao,
  Ishangulyyeva, Burr, Gaylord, Hofstetter, Keefover-Ring, Raffa, and
  Kolb]{Nadir_Zanganeh_2021}
Nadir Erbilgin, Leila Zanganeh, Jennifer~G. Klutsch, Shih-hsuan Chen, Shiyang
  Zhao, Guncha Ishangulyyeva, Stephen~J. Burr, Monica Gaylord, Richard
  Hofstetter, Ken Keefover-Ring, Kenneth~F. Raffa, and Thomas Kolb.
\newblock Combined drought and bark beetle attacks deplete non-structural
  carbohydrates and promote death of mature pine trees.
\newblock \emph{Plant, Cell \& Environment}, 44\penalty0 (12):\penalty0
  3866--3881, 2021.

\bibitem[Fassnacht et~al.(2012)Fassnacht, Latifi, and
  Koch]{AngularVegIndex_Spectroscopy_Bavarian}
Fabian~Ewald Fassnacht, Hooman Latifi, and Barbara Koch.
\newblock An angular vegetation index for imaging spectroscopy
  data—{P}reliminary results on forest damage detection in the {Bavarian
  National Park, Germany}.
\newblock \emph{International Journal of Applied Earth Observation and
  Geoinformation}, 19:\penalty0 308--321, 2012.

\bibitem[Fassnacht et~al.(2014)Fassnacht, Latifi, Ghosh, Joshi, and
  Koch]{Assess_Hyperspec_Mortality}
Fabian~Ewald Fassnacht, Hooman Latifi, Aniruddha Ghosh, Pawan~Kumar Joshi, and
  Barbara Koch.
\newblock Assessing the potential of hyperspectral imagery to map bark
  beetle-induced tree mortality.
\newblock \emph{Remote Sensing of Environment}, 140:\penalty0 533--548, 2014.

\bibitem[Fernandez-Carrillo et~al.(2020)Fernandez-Carrillo, Patočka,
  Dobrovolný, Franco-Nieto, and Revilla-Romero]{Monitor_CentrEuro_Validate}
Angel Fernandez-Carrillo, Zdeněk Patočka, Lumír Dobrovolný, Antonio
  Franco-Nieto, and Beatriz Revilla-Romero.
\newblock Monitoring bark beetle forest damage in {Central Europe}. {A} remote
  sensing approach validated with field data.
\newblock \emph{Remote Sensing}, 12\penalty0 (21), 2020.

\bibitem[Fettig et~al.(2007)Fettig, Klepzig, Billings, Munson, Nebeker,
  Negrón, and Nowak]{BeetleOutbreak_US}
Christopher~J. Fettig, Kier~D. Klepzig, Ronald~F. Billings, A.~Steven Munson,
  T.~Evan Nebeker, Jose~F. Negrón, and John~T. Nowak.
\newblock The effectiveness of vegetation management practices for prevention
  and control of bark beetle infestations in coniferous forests of the western
  and southern {United States}.
\newblock \emph{Forest Ecology and Management}, 238\penalty0 (1):\penalty0
  24--53, 2007.

\bibitem[Foster et~al.(2017)Foster, Walter, Shugart, Sibold, and
  Negron]{SpectEvid_Early_Engelmann}
Adrianna~C. Foster, Jonathan~A. Walter, Herman~H. Shugart, Jason Sibold, and
  Jose Negron.
\newblock Spectral evidence of early-stage spruce beetle infestation in
  {Engelmann} spruce.
\newblock \emph{Forest Ecology and Management}, 384:\penalty0 347--357, 2017.

\bibitem[Franklin et~al.(2003)Franklin, Wulder, Skakun, and
  Carroll]{MPB_RedAttack_Strat_Landsat_BC}
S.E. Franklin, M.A. Wulder, R.S. Skakun, and A.L. Carroll.
\newblock Mountain pine beetle red-attack forest damage classification using
  stratified {Landsat TM} data in {British Columbia, Canada}.
\newblock \emph{Photogrammetric Engineering and Remote Sensing}, 69\penalty0
  (3):\penalty0 283--288, 2003.

\bibitem[Gandhi and Hofstetter(2021)]{Book_barkbeetle2021}
Kamal~J.K. Gandhi and Richard~W. Hofstetter, editors.
\newblock \emph{Bark Beetle Management, Ecology, and Climate Change}.
\newblock Elsevier, 2021.
\newblock \doi{10.1016/C2019-0-04282-3}.

\bibitem[Garrity et~al.(2013)Garrity, Allen, Brumby, Gangodagamage, McDowell,
  and Cai]{Quanti_MixedSpecies_MultiTempHighRes_Sat}
Steven~R. Garrity, Craig~D. Allen, Steven~P. Brumby, Chandana Gangodagamage,
  Nate~G. McDowell, and D.~Michael Cai.
\newblock Quantifying tree mortality in a mixed species woodland using
  multitemporal high spatial resolution satellite imagery.
\newblock \emph{Remote Sensing of Environment}, 129:\penalty0 54--65, 2013.

\bibitem[Gartner et~al.(2015)Gartner, Veblen, Leyk, and
  Wessman]{Detect_MPB_Ponderosa_HighResAerial}
Meredith~H. Gartner, Thomas~T. Veblen, Stefan Leyk, and Carol~A. Wessman.
\newblock Detection of mountain pine beetle-killed ponderosa pine in a
  heterogeneous landscape using high-resolution aerial imagery.
\newblock \emph{International Journal of Remote Sensing}, 36\penalty0
  (21):\penalty0 5353--5372, 2015.

\bibitem[Goodsman and Weber(2022)]{Devin_MPB_AerialSurvey}
Devin~W. Goodsman and Jim Weber.
\newblock A correction for serial nonindependence in mountain pine beetle
  aerial survey data to reduce overestimation of cumulative damage.
\newblock \emph{Canadian Journal of Forest Research}, 2022.

\bibitem[Government(2014)]{AerialSurvey_manual}
Alberta Government.
\newblock {Alberta Forest health aerial survey manual}, 2014.

\bibitem[GUO(1991)]{BCET_contrast}
LIU~JIAN GUO.
\newblock Balance contrast enhancement technique and its application in image
  colour composition.
\newblock \emph{International Journal of Remote Sensing}, 12\penalty0
  (10):\penalty0 2133--2151, 1991.

\bibitem[Hall et~al.(2016)Hall, Castilla, White, Cooke, and
  Skakun]{Review_Guillermo}
R.J. Hall, G.~Castilla, J.C. White, B.J. Cooke, and R.S. Skakun.
\newblock Remote sensing of forest pest damage: {A} review and lessons learned
  from a canadian perspective.
\newblock \emph{The Canadian Entomologist}, 148\penalty0 (S1):\penalty0
  296–356, 2016.

\bibitem[Hart and Veblen(2015)]{Spruce_HighMedium_Remote}
Sarah~J. Hart and Thomas~T. Veblen.
\newblock Detection of spruce beetle-induced tree mortality using high- and
  medium-resolution remotely sensed imagery.
\newblock \emph{Remote Sensing of Environment}, 168:\penalty0 134--145, 2015.

\bibitem[He et~al.(2016)He, Zhang, Ren, and Sun]{ResNet}
Kaiming He, Xiangyu Zhang, Shaoqing Ren, and Jian Sun.
\newblock {Deep residual learning for image recognition}.
\newblock In \emph{Proc. IEEE CVPR}, pages 770--778, 2016.

\bibitem[Hellwig et~al.(2021)Hellwig, Stelmaszczuk-Górska, Dubois, Wolsza,
  Truckenbrodt, Sagichewski, Chmara, Bannehr, Lausch, and
  Schmullius]{Map_Gyrocopter_HyperField}
Florian~M. Hellwig, Martyna~A. Stelmaszczuk-Górska, Clémence Dubois, Marco
  Wolsza, Sina~C. Truckenbrodt, Herbert Sagichewski, Sergej Chmara, Lutz
  Bannehr, Angela Lausch, and Christiane Schmullius.
\newblock Mapping european spruce bark beetle infestation at its early phase
  using gyrocopter-mounted hyperspectral data and field measurements.
\newblock \emph{Remote Sensing}, 13\penalty0 (22), 2021.

\bibitem[Hicke and Logan(2009)]{Map_WhiteBark_MPB_Sat}
Jeffrey~A. Hicke and Jesse Logan.
\newblock Mapping whitebark pine mortality caused by a mountain pine beetle
  outbreak with high spatial resolution satellite imagery.
\newblock \emph{International Journal of Remote Sensing}, 30\penalty0
  (17):\penalty0 4427--4441, 2009.

\bibitem[Hl{\'a}sny et~al.(2021)Hl{\'a}sny, K{\"o}nig, Krokene, Lindner,
  Montagn{\'e}-Huck, M{\"u}ller, Qin, Raffa, Schelhaas, Svoboda, Viiri, and
  Seidl]{Living_Bark_Beetles}
Tom{\'a}{\v{s}} Hl{\'a}sny, Louis K{\"o}nig, Paal Krokene, Marcus Lindner,
  Claire Montagn{\'e}-Huck, J{\"o}rg M{\"u}ller, Hua Qin, Kenneth~F. Raffa,
  Mart-Jan Schelhaas, Miroslav Svoboda, Heli Viiri, and Rupert Seidl.
\newblock Bark beetle outbreaks in {Europe}: State of knowledge and ways
  forward for management.
\newblock \emph{Current Forestry Reports}, 7\penalty0 (3):\penalty0 138--165,
  2021.

\bibitem[Honkavaara et~al.(2020)Honkavaara, N\"asi, Oliveira, Viljanen,
  Suomalainen, Khoramshahi, Hakala, Nevalainen, Markelin, Vuorinen,
  Kankaanhuhta, Lyytik\"ainen-Saarenmaa, and
  Haataja]{MultiTemp_HyperMultispect_NorwaySpruce}
E.~Honkavaara, R.~N\"asi, R.~Oliveira, N.~Viljanen, J.~Suomalainen,
  E.~Khoramshahi, T.~Hakala, O.~Nevalainen, L.~Markelin, M.~Vuorinen,
  V.~Kankaanhuhta, P.~Lyytik\"ainen-Saarenmaa, and L.~Haataja.
\newblock Using multitemporal hyper- and multispectral {UAV} imaging for
  detecting bark beetle infestation on norway spruce.
\newblock \emph{The International Archives of the Photogrammetry, Remote
  Sensing and Spatial Information Sciences}, XLIII-B3-2020:\penalty0 429--434,
  2020.

\bibitem[Hsieh et~al.(2001)Hsieh, Lee, and Chen]{SpatialSpectralVariance2}
Pi-Fuei Hsieh, L.C. Lee, and Nai-Yu Chen.
\newblock Effect of spatial resolution on classification errors of pure and
  mixed pixels in remote sensing.
\newblock \emph{IEEE Trans. Geoscience and Remote Sensing}, 39\penalty0
  (12):\penalty0 2657--2663, 2001.

\bibitem[{Huang} et~al.(2017){Huang}, {Liu}, v.~d. {Maaten}, and
  {Weinberger}]{DenseNet}
G.~{Huang}, Z.~{Liu}, L.~v.~d. {Maaten}, and K.~Q. {Weinberger}.
\newblock Densely connected convolutional networks.
\newblock In \emph{Proc. IEEE CVPR}, pages 2261--2269, 2017.

\bibitem[Huo et~al.(2021)Huo, Persson, and
  Lindberg]{EarlyDet_EuropeSpruce_NDRS}
Langning Huo, Henrik~Jan Persson, and Eva Lindberg.
\newblock Early detection of forest stress from european spruce bark beetle
  attack, and a new vegetation index: Normalized distance red \& {SWIR}
  ({NDRS}).
\newblock \emph{Remote Sensing of Environment}, 255:\penalty0 112240, 2021.

\bibitem[Immitzer and Atzberger(2014)]{EarlyDet_NorwaySpruce_WorldView2}
Markus Immitzer and Clement Atzberger.
\newblock Early detection of bark beetle infestation in norway spruce ({Picea
  abies, L.}) using {WorldView-2} data.
\newblock \emph{Journal of Photogrammetry, Remote Sensing and Geoinformation
  Science}, 5:\penalty0 351--367, 2014.

\bibitem[Jiang et~al.(2019)Jiang, Yao, and Heurich]{DeadWood_FCNDenseNet}
Shenlu Jiang, Wei Yao, and Marco Heurich.
\newblock Dead wood detection based on semantic segmentation of {VHR} aerial
  {CIR} imagery using optimized {FCN-Densenet}.
\newblock In \emph{Proc. Joint ISPRS Conference on Photogrammetric Image
  Analysis and Munich Remote Sensing Symposium}, pages 127--133, 2019.

\bibitem[Junttila et~al.(2022)Junttila, Näsi, Koivumäki, Imangholiloo,
  Saarinen, Raisio, Holopainen, Hyyppä, Hyyppä, Lyytikäinen-Saarenmaa,
  Vastaranta, and Honkavaara]{Multispectral_Benefits_Spruce_Decline}
Samuli Junttila, Roope Näsi, Niko Koivumäki, Mohammad Imangholiloo, Ninni
  Saarinen, Juha Raisio, Markus Holopainen, Hannu Hyyppä, Juha Hyyppä, Päivi
  Lyytikäinen-Saarenmaa, Mikko Vastaranta, and Eija Honkavaara.
\newblock Multispectral imagery provides benefits for mapping spruce tree
  decline due to bark beetle infestation when acquired late in the season.
\newblock \emph{Remote Sensing}, 14\penalty0 (4), 2022.

\bibitem[Kamińska et~al.(2020)Kamińska, Lisiewicz, Kraszewski, and
  Stereńczak]{Habitat_Dynamics_NorwayDieback}
Agnieszka Kamińska, Maciej Lisiewicz, Bartłomiej Kraszewski, and Krzysztof
  Stereńczak.
\newblock Habitat and stand factors related to spatial dynamics of {Norway}
  spruce dieback driven by {Ips typographus (L.)} in the {Białowieża} forest
  district.
\newblock \emph{Forest Ecology and Management}, 476:\penalty0 118432, 2020.

\bibitem[Kapil et~al.(2022)Kapil, Marvasti-Zadeh, Goodsman, Ray, and
  Erbilgin]{Ours_DL_BarkBeetle}
Rudraksh Kapil, Seyed~Mojtaba Marvasti-Zadeh, Devin Goodsman, Nilanjan Ray, and
  Nadir Erbilgin.
\newblock Classification of bark beetle-induced forest tree mortality using
  deep learning.
\newblock In \emph{Proc. ICPR Workshop}, 2022.

\bibitem[Kislov et~al.(2021)Kislov, Korznikov, Altman, Vozmishcheva, and
  Krestov]{DistSeg_Sat_DL}
Dmitry~E. Kislov, Kirill~A. Korznikov, Jan Altman, Anna~S. Vozmishcheva, and
  Pavel~V. Krestov.
\newblock Extending deep learning approaches for forest disturbance
  segmentation on very high-resolution satellite images.
\newblock \emph{Remote Sensing in Ecology and Conservation}, 7\penalty0
  (3):\penalty0 355--368, 2021.

\bibitem[Klouček et~al.(2019)Klouček, Komárek, Surový, Hrach, Janata, and
  Vašíček]{UAV_PrecDetection_BBInf}
Tomáš Klouček, Jan Komárek, Peter Surový, Karel Hrach, Přemysl Janata,
  and Bedřich Vašíček.
\newblock The use of {UAV} mounted sensors for precise detection of bark beetle
  infestation.
\newblock \emph{Remote Sensing}, 11\penalty0 (13), 2019.

\bibitem[Koontz et~al.(2021)Koontz, Latimer, Mortenson, Fettig, and
  North]{CrosInteract_Mortality}
Michael~J. Koontz, Andrew~M. Latimer, Leif~A. Mortenson, Christopher~J. Fettig,
  and Malcolm~P. North.
\newblock Cross-scale interaction of host tree size and climatic water deficit
  governs bark beetle-induced tree mortality.
\newblock \emph{Nat Commun}, 12\penalty0 (129), 2021.

\bibitem[Koreň et~al.(2021)Koreň, Jakuš, Zápotocký, Barka, Holuša,
  Ďuračiová, and Blaženec]{ML_SpatialDist_BBI}
Milan Koreň, Rastislav Jakuš, Martin Zápotocký, Ivan Barka, Jaroslav
  Holuša, Renata Ďuračiová, and Miroslav Blaženec.
\newblock Assessment of machine learning algorithms for modeling the spatial
  distribution of bark beetle infestation.
\newblock \emph{Forests}, 12\penalty0 (4), 2021.

\bibitem[Kurz et~al.(2008)Kurz, Dymond, Stinson, Rampley, Neilson, Carroll,
  Ebata, and Safranyik]{Kurz_etal2008}
W.~A. Kurz, C.~C. Dymond, G.~Stinson, G.~J. Rampley, E.~T. Neilson, A.~L.
  Carroll, T.~Ebata, and L.~Safranyik.
\newblock Mountain pine beetle and forest carbon feedback to climate change.
\newblock \emph{Nature}, 452\penalty0 (7190):\penalty0 987--990, 2008.

\bibitem[Latifi et~al.(2014{\natexlab{a}})Latifi, Fassnacht, Schumann, and
  Dech]{ObjectExtract_LandsatSPOT}
Hooman Latifi, Fabian~E. Fassnacht, Bastian Schumann, and Stefan Dech.
\newblock Object-based extraction of bark beetle ({Ips typographus L.})
  infestations using multi-date {LANDSAT and SPOT} satellite imagery.
\newblock \emph{Progress in Physical Geography: Earth and Environment},
  38\penalty0 (6):\penalty0 755--785, 2014{\natexlab{a}}.

\bibitem[Latifi et~al.(2014{\natexlab{b}})Latifi, Schumann, Kautz, and
  Dech]{Spatial_Multidate_SPOT_Landsat}
Hooman Latifi, Bastian Schumann, Markus Kautz, and Stefan Dech.
\newblock Spatial characterization of bark beetle infestations by a multidate
  synergy of {SPOT and Landsat} imagery.
\newblock \emph{Environmental Monitoring and Assessment}, 186:\penalty0
  441–456, 2014{\natexlab{b}}.

\bibitem[Latifi et~al.(2018)Latifi, Dahms, Beudert, Heurich, Kübert, and
  Dech]{RapidEye_Detect_AreaSpruce}
Hooman Latifi, Thorsten Dahms, Burkhard Beudert, Marco Heurich, Carina Kübert,
  and Stefan Dech.
\newblock Synthetic {RapidEye} data used for the detection of area-based spruce
  tree mortality induced by bark beetles.
\newblock \emph{GIScience \& Remote Sensing}, 55\penalty0 (6):\penalty0
  839--859, 2018.

\bibitem[Lausch et~al.(2013)Lausch, Heurich, Gordalla, Dobner,
  Gwillym-Margianto, and Salbach]{Forecast_Vital_Hyperspect}
A.~Lausch, M.~Heurich, D.~Gordalla, H.-J. Dobner, S.~Gwillym-Margianto, and
  C.~Salbach.
\newblock Forecasting potential bark beetle outbreaks based on spruce forest
  vitality using hyperspectral remote-sensing techniques at different scales.
\newblock \emph{Forest Ecology and Management}, 308:\penalty0 76--89, 2013.

\bibitem[Lechner et~al.(2020)Lechner, Foody, and
  Boyd]{RS_ForestEcology_Management}
Alex~M. Lechner, Giles~M. Foody, and Doreen~S. Boyd.
\newblock Applications in remote sensing to forest ecology and management.
\newblock \emph{One Earth}, 2\penalty0 (5):\penalty0 405--412, 2020.

\bibitem[Li et~al.(2012)Li, Xu, and Liang]{RandomFrog}
Hong-Dong Li, Qing-Song Xu, and Yi-Zeng Liang.
\newblock Random frog: An efficient reversible jump {Markov Chain Monte
  Carlo-like} approach for variable selection with applications to gene
  selection and disease classification.
\newblock \emph{Analytica Chimica Acta}, 740:\penalty0 20--26, 2012.

\bibitem[Liang et~al.(2014)Liang, Chen, Hawbaker, Zhu, and
  Gong]{MPB_GrowthTrend_Landsat}
Lu~Liang, Yanlei Chen, Todd~J. Hawbaker, Zhiliang Zhu, and Peng Gong.
\newblock Mapping mountain pine beetle mortality through growth trend analysis
  of time-series landsat data.
\newblock \emph{Remote Sensing}, 6\penalty0 (6):\penalty0 5696--5716, 2014.

\bibitem[Lieutier et~al.(2009)Lieutier, Yart, and
  Salle]{Stimulation_tree_defenses}
Fran{\c c}ois Lieutier, Annie Yart, and Aur{\'e}lien Salle.
\newblock {Stimulation of tree defenses by Ophiostomatoid fungi can explain
  attack success of bark beetles on conifers}.
\newblock \emph{{Annals of Forest Science}}, 66\penalty0 (8), 2009.

\bibitem[Lin et~al.(2019)Lin, Huang, Wang, Huang, and
  Liu]{PineShootBeetle_Hyperspect_LiDAR}
Qinan Lin, Huaguo Huang, Jingxu Wang, Kan Huang, and Yangyang Liu.
\newblock Detection of pine shoot beetle ({PSB}) stress on pine forests at
  individual tree level using {UAV}-based hyperspectral imagery and {LiDAR}.
\newblock \emph{Remote Sensing}, 11\penalty0 (21), 2019.

\bibitem[Liu et~al.(2021)Liu, Frey, Denter, Zielewska-Büttner, Still, and
  Koch]{DeadTree_PixObj_WorldView}
Xiang Liu, Julian Frey, Martin Denter, Katarzyna Zielewska-Büttner, Nicole
  Still, and Barbara Koch.
\newblock Mapping standing dead trees in temperate montane forests using a
  pixel- and object-based image fusion method and stereo {WorldView-3} imagery.
\newblock \emph{Ecological Indicators}, 133:\penalty0 108438, 2021.

\bibitem[Long and Lawrence(2016)]{PercentMortality_MPB_Damage}
John~A. Long and Rick~L. Lawrence.
\newblock Mapping percent tree mortality due to mountain pine beetle damage.
\newblock \emph{Forest Science}, 62\penalty0 (4):\penalty0 392--402, 04 2016.

\bibitem[Lopatin et~al.(2019)Lopatin, Dolos, Kattenborn, and
  Fassnacht]{SpatialSpectralVariance1}
Javier Lopatin, Klara Dolos, Teja Kattenborn, and Fabian~E. Fassnacht.
\newblock How canopy shadow affects invasive plant species classification in
  high spatial resolution remote sensing.
\newblock \emph{Remote Sensing in Ecology and Conservation}, 5\penalty0
  (4):\penalty0 302--317, 2019.

\bibitem[Marvasti-Zadeh et~al.(2022)Marvasti-Zadeh, Cheng, Ghanei-Yakhdan, and
  Kasaei]{DL_Tracking}
Seyed~Mojtaba Marvasti-Zadeh, Li~Cheng, Hossein Ghanei-Yakhdan, and Shohreh
  Kasaei.
\newblock Deep learning for visual tracking: A comprehensive survey.
\newblock \emph{IEEE Trans. Intelligent Transportation Systems}, 23\penalty0
  (5):\penalty0 3943--3968, 2022.

\bibitem[Meddens et~al.(2011)Meddens, Hicke, and
  Vierling]{EvalPoten_MultiSpect_MultipleStage}
Arjan~J.H. Meddens, Jeffrey~A. Hicke, and Lee~A. Vierling.
\newblock Evaluating the potential of multispectral imagery to map multiple
  stages of tree mortality.
\newblock \emph{Remote Sensing of Environment}, 115\penalty0 (7):\penalty0
  1632--1642, 2011.

\bibitem[Meddens et~al.(2013)Meddens, Hicke, Vierling, and
  Hudak]{Eval_DetectBB_SingMultiData_Landsat}
Arjan~J.H. Meddens, Jeffrey~A. Hicke, Lee~A. Vierling, and Andrew~T. Hudak.
\newblock Evaluating methods to detect bark beetle-caused tree mortality using
  single-date and multi-date {Landsat} imagery.
\newblock \emph{Remote Sensing of Environment}, 132:\penalty0 49--58, 2013.

\bibitem[Meng et~al.(2022)Meng, Gao, Zhao, Huang, Sun, Lv, and
  Huang]{Landsat_SouthPineBeet_Severity}
Ran Meng, Renjie Gao, Feng Zhao, Chengquan Huang, Rui Sun, Zhengang Lv, and
  Zehua Huang.
\newblock Landsat-based monitoring of southern pine beetle infestation severity
  and severity change in a temperate mixed forest.
\newblock \emph{Remote Sensing of Environment}, 269:\penalty0 112847, 2022.

\bibitem[Minařík et~al.(2020)Minařík, Langhammer, and
  Lendzioch]{CrownExt_UASMultispec_MixedForest}
Robert Minařík, Jakub Langhammer, and Theodora Lendzioch.
\newblock Automatic tree crown extraction from uas multispectral imagery for
  the detection of bark beetle disturbance in mixed forests.
\newblock \emph{Remote Sensing}, 12\penalty0 (24), 2020.

\bibitem[Minařík et~al.(2021)Minařík, Langhammer, and
  Lendzioch]{UAS_Multispect_DL}
Robert Minařík, Jakub Langhammer, and Theodora Lendzioch.
\newblock Detection of bark beetle disturbance at tree level using {UAS}
  multispectral imagery and deep learning.
\newblock \emph{Remote Sensing}, 13\penalty0 (23), 2021.

\bibitem[Mullen et~al.(2018)Mullen, Yuan, and
  Mitchell]{MPB_BlackHills_SouthDakota}
Kyle Mullen, Fei Yuan, and Martin Mitchell.
\newblock The mountain pine beetle epidemic in the {Black Hills, South Dakota}:
  The consequences of long term fire policy, climate change and the use of
  remote sensing to enhance mitigation.
\newblock \emph{Journal of Geography and Geology}, 10\penalty0 (1):\penalty0
  69--82, 2018.

\bibitem[Nguyen et~al.(2021)Nguyen, Lopez~Caceres, Moritake, Kentsch, Shu, and
  Diez]{SickFir_UAV_DL}
Ha~Trang Nguyen, Maximo~Larry Lopez~Caceres, Koma Moritake, Sarah Kentsch, Hase
  Shu, and Yago Diez.
\newblock Individual sick fir tree (abies mariesii) identification in insect
  infested forests by means of uav images and deep learning.
\newblock \emph{Remote Sensing}, 13\penalty0 (2), 2021.

\bibitem[Niemann et~al.(2015)Niemann, Quinn, Stephen, Visintini, and
  Parton]{HyperRS_MPB_Emphasis_Previsual}
K.~Olaf Niemann, G.~Quinn, R.~Stephen, F.~Visintini, and D.~Parton.
\newblock Hyperspectral remote sensing of mountain pine beetle with an emphasis
  on previsual assessment.
\newblock \emph{Canadian Journal of Remote Sensing}, 41\penalty0 (3):\penalty0
  191--202, 2015.

\bibitem[Näsi et~al.(2015)Näsi, Honkavaara, Lyytikäinen-Saarenmaa,
  Blomqvist, Litkey, Hakala, Viljanen, Kantola, Tanhuanpää, and
  Holopainen]{Photogram_Hyperspectral_TreeLevel}
Roope Näsi, Eija Honkavaara, Päivi Lyytikäinen-Saarenmaa, Minna Blomqvist,
  Paula Litkey, Teemu Hakala, Niko Viljanen, Tuula Kantola, Topi Tanhuanpää,
  and Markus Holopainen.
\newblock Using {UAV}-based photogrammetry and hyperspectral imaging for
  mapping bark beetle damage at tree-level.
\newblock \emph{Remote Sensing}, 7\penalty0 (11):\penalty0 15467--15493, 2015.

\bibitem[Näsi et~al.(2018)Näsi, Honkavaara, Blomqvist,
  Lyytikäinen-Saarenmaa, Hakala, Viljanen, Kantola, and
  Holopainen]{UrbanForest_Hyperspec_UAV_Aircraft}
Roope Näsi, Eija Honkavaara, Minna Blomqvist, Päivi Lyytikäinen-Saarenmaa,
  Teemu Hakala, Niko Viljanen, Tuula Kantola, and Markus Holopainen.
\newblock Remote sensing of bark beetle damage in urban forests at individual
  tree level using a novel hyperspectral camera from {UAV} and aircraft.
\newblock \emph{Urban Forestry \& Urban Greening}, 30:\penalty0 72--83, 2018.

\bibitem[of~Alberta(2016)]{Book_MPB_Alberta_Gov}
Government of~Alberta.
\newblock \emph{Mountain Pine Beetle Detection and Management in Alberta}.
\newblock 2016.
\newblock [Online; accessed 10-Feb-2022].

\bibitem[Ortiz et~al.(2013)Ortiz, Breidenbach, and
  Kändler]{EarlyDet_TerraSarX_RapidEye}
Sonia~M. Ortiz, Johannes Breidenbach, and Gerald Kändler.
\newblock Early detection of bark beetle green attack using {TerraSAR-X} and
  {RapidEye} data.
\newblock \emph{Remote Sensing}, 5\penalty0 (4):\penalty0 1912--1931, 2013.

\bibitem[Pontius et~al.(2020)Pontius, Schaberg, and
  Hanavan]{Review_Book_EarlyForestDisturbance}
Jennifer Pontius, Paul Schaberg, and Ryan Hanavan.
\newblock \emph{Remote sensing for early, detailed, and accurate detection of
  forest disturbance and decline for protection of biodiversity}, pages
  121--154.
\newblock Springer International Publishing, 2020.

\bibitem[Redmon and Farhadi(2017)]{YOLO_v2}
Joseph Redmon and Ali Farhadi.
\newblock {YOLO9000}: Better, faster, stronger.
\newblock In \emph{Proc. CVPR}, July 2017.

\bibitem[Redmon and Farhadi(2018)]{YOLO_v3}
Joseph Redmon and Ali Farhadi.
\newblock {YOLOv3}: An incremental improvement.
\newblock \emph{CoRR}, abs/1804.02767, 2018.

\bibitem[Safonova et~al.(2019)Safonova, Tabik, Alcaraz-Segura, Rubtsov,
  Maglinets, and Herrera]{BB_Fir_DL}
Anastasiia Safonova, Siham Tabik, Domingo Alcaraz-Segura, Alexey Rubtsov, Yuriy
  Maglinets, and Francisco Herrera.
\newblock Detection of fir trees ({Abies sibirica}) damaged by the bark beetle
  in unmanned aerial vehicle images with deep learning.
\newblock \emph{Remote Sensing}, 11\penalty0 (6), 2019.

\bibitem[Safonova et~al.(2022)Safonova, Hamad, Alekhina, and
  Kaplun]{Detection_UAV_YOLOs}
Anastasiia Safonova, Yousif Hamad, Anna Alekhina, and Dmitry Kaplun.
\newblock Detection of norway spruce trees ({Picea Abies}) infested by bark
  beetle in {UAV} images using {YOLOs} architectures.
\newblock \emph{IEEE Access}, 10:\penalty0 10384--10392, 2022.

\bibitem[Schaeffer et~al.(2021)Schaeffer, Jiménez-Lizárraga,
  Rodriguez-Sanchez, Cuellar-Rodríguez, Aguirre-Calderón, Reyna-González,
  and Escobar]{DetectBB_Threshold_CellAutomata}
S.~Elisa Schaeffer, Manuel Jiménez-Lizárraga, Sara~V. Rodriguez-Sanchez,
  Gerardo Cuellar-Rodríguez, Oscar~A. Aguirre-Calderón, Angel~M.
  Reyna-González, and Alan Escobar.
\newblock {Detection of bark beetle infestation in drone imagery via
  thresholding cellular automata}.
\newblock \emph{Journal of Applied Remote Sensing}, 15\penalty0 (1):\penalty0 1
  -- 20, 2021.

\bibitem[Senf et~al.(2015)Senf, Pflugmacher, Wulder, and
  Hostert]{SpectTemporal_Defoliat_Landsat}
Cornelius Senf, Dirk Pflugmacher, Michael~A. Wulder, and Patrick Hostert.
\newblock Characterizing spectral–temporal patterns of defoliator and bark
  beetle disturbances using {Landsat} time series.
\newblock \emph{Remote Sensing of Environment}, 170:\penalty0 166--177, 2015.

\bibitem[Senf et~al.(2017)Senf, Seidl, and Hostert]{Review_InsectDisturbances}
Cornelius Senf, Rupert Seidl, and Patrick Hostert.
\newblock Remote sensing of forest insect disturbances: Current state and
  future directions.
\newblock \emph{International Journal of Applied Earth Observation and
  Geoinformation}, 60:\penalty0 49--60, 2017.

\bibitem[Stych et~al.(2019)Stych, Jerabkova, Lastovicka, Riedl, and
  Paluba]{Comp_WorldViewLandsat_SVMNeuralNet}
Premysl Stych, Barbora Jerabkova, Josef Lastovicka, Martin Riedl, and Daniel
  Paluba.
\newblock A comparison of {WorldView-2 and Landsat-8} images for the
  classification of forests affected by bark beetle outbreaks using a support
  vector machine and a neural network: A case study in the {Sumava} mountains.
\newblock \emph{Geosciences}, 9\penalty0 (9), 2019.

\bibitem[Tane et~al.(2018)Tane, Roberts, Koltunov, Sweeney, and
  Ramirez]{Framework_Ecoregion_Spaceborne}
Zachary Tane, Dar Roberts, Alexander Koltunov, Stuart Sweeney, and Carlos
  Ramirez.
\newblock A framework for detecting conifer mortality across an ecoregion using
  high spatial resolution spaceborne imaging spectroscopy.
\newblock \emph{Remote Sensing of Environment}, 209:\penalty0 195--210, 2018.

\bibitem[Toth and Jóźków(2016)]{RS_review_platform_sensor}
Charles Toth and Grzegorz Jóźków.
\newblock Remote sensing platforms and sensors: A survey.
\newblock \emph{ISPRS Journal of Photogrammetry and Remote Sensing},
  115:\penalty0 22--36, 2016.

\bibitem[Trubin et~al.(2022)Trubin, Mezei, Zabihi, Surový, and
  Jakuš]{Northwenmost_Model_Climate}
Aleksei Trubin, Pavel Mezei, Khodabakhsh Zabihi, Peter Surový, and Rastislav
  Jakuš.
\newblock Northernmost european spruce bark beetle {Ips typographus} outbreak:
  Modelling tree mortality using remote sensing and climate data.
\newblock \emph{Forest Ecology and Management}, 505:\penalty0 119829, 2022.

\bibitem[Ustin and Middleton(2021)]{RS_Advances}
Susan~L. Ustin and Elizabeth~M. Middleton.
\newblock Current and near-term advances in {Earth} observation for ecological
  applications.
\newblock \emph{Ecological Processes}, 10\penalty0 (1), 2021.

\bibitem[Vega and Hofstetter(2014)]{Book_BarkBeetles_Biology_Ecology_Vega}
Fernando~E Vega and Richard~W Hofstetter.
\newblock \emph{Bark beetles: Biology and ecology of native and invasive
  species}.
\newblock Academic Press, 2014.

\bibitem[Vošvrdová et~al.(2023)Vošvrdová, Johansson, Turčáni, Jakuš,
  Tyšer, Schlyter, and Modlinger]{Sniffer_dogs}
N.~Vošvrdová, A.~Johansson, M.~Turčáni, R.~Jakuš, D.~Tyšer, F.~Schlyter,
  and R.~Modlinger.
\newblock Dogs trained to recognise a bark beetle pheromone locate recently
  attacked spruces better than human experts.
\newblock \emph{Forest Ecology and Management}, 528:\penalty0 120626, 2023.

\bibitem[Westfall et~al.(2019)Westfall, Ebata, and Inc]{AerialSurvey_BC}
Joan Westfall, Tim Ebata, and HR~GISolutions Inc.
\newblock {Forest health arial overview survey standards for British Columbia},
  2019.

\bibitem[White et~al.(2005)White, Wulder, Brooks, Reich, and
  Wheate]{RedAttack_MPB_HSR_Sat}
Joanne~C. White, Michael~A. Wulder, Darin Brooks, Richard Reich, and Roger~D.
  Wheate.
\newblock Detection of red attack stage mountain pine beetle infestation with
  high spatial resolution satellite imagery.
\newblock \emph{Remote Sensing of Environment}, 96\penalty0 (3):\penalty0
  340--351, 2005.

\bibitem[Wich and Koh(2018)]{Book_Conservation_Drones}
Serge~A. Wich and Lian~Pin Koh.
\newblock \emph{Conservation drones: Mapping and monitoring biodiversity}.
\newblock Oxford University Press, 2018.

\bibitem[Wulder et~al.(2006{\natexlab{a}})Wulder, White, Bentz, Alvarez, and
  Coops]{EstProb_MPB_RedAttack}
M.A. Wulder, J.C. White, B.~Bentz, M.F. Alvarez, and N.C. Coops.
\newblock Estimating the probability of mountain pine beetle red-attack damage.
\newblock \emph{Remote Sensing of Environment}, 101\penalty0 (2):\penalty0
  150--166, 2006{\natexlab{a}}.

\bibitem[Wulder et~al.(2006{\natexlab{b}})Wulder, Dymond, White, Leckie, and
  Carroll]{Review_Wulder}
Michael~A. Wulder, Caren~C. Dymond, Joanne~C. White, Donald~G. Leckie, and
  Allan~L. Carroll.
\newblock Surveying mountain pine beetle damage of forests: {A} review of
  remote sensing opportunities.
\newblock \emph{Forest Ecology and Management}, 221\penalty0 (1):\penalty0
  27--41, 2006{\natexlab{b}}.

\bibitem[Wulder et~al.(2009)Wulder, White, Carroll, and
  Coops]{Review_Wulder_Challenges}
Michael~A Wulder, Joanne~C White, Allan~L Carroll, and Nicholas~C Coops.
\newblock Challenges for the operational detection of mountain pine beetle
  green attack with remote sensing.
\newblock \emph{The Forestry Chronicle}, 85\penalty0 (1):\penalty0 32--38,
  2009.

\bibitem[Xie et~al.(2008)Xie, Sha, and Yu]{RS_VegetationMapping_Review}
Yichun Xie, Zongyao Sha, and Mei Yu.
\newblock {Remote sensing imagery in vegetation mapping: a review}.
\newblock \emph{Journal of Plant Ecology}, 1\penalty0 (1):\penalty0 9--23,
  2008.

\bibitem[Ye et~al.(2021)Ye, Rogan, Zhu, Hawbaker, Hart, Andrus, Meddens, Hicke,
  Eastman, and Kulakowski]{SubtleChange_Landsat_MPBSpruce}
Su~Ye, John Rogan, Zhe Zhu, Todd~J. Hawbaker, Sarah~J. Hart, Robert~A. Andrus,
  Arjan~J.H. Meddens, Jeffrey~A. Hicke, J.~Ronald Eastman, and Dominik
  Kulakowski.
\newblock Detecting subtle change from dense landsat time series: Case studies
  of mountain pine beetle and spruce beetle disturbance.
\newblock \emph{Remote Sensing of Environment}, 263:\penalty0 112560, 2021.

\bibitem[Zabihi et~al.(2021)Zabihi, Surovy, Trubin, Singh, and
  Jakuš]{Review_Zabihi}
Khodabakhsh Zabihi, Peter Surovy, Aleksei Trubin, Vivek~Vikram Singh, and
  Rastislav Jakuš.
\newblock A review of major factors influencing the accuracy of mapping
  green-attack stage of bark beetle infestations using satellite imagery:
  Prospects to avoid data redundancy.
\newblock \emph{Remote Sensing Applications: Society and Environment},
  24:\penalty0 100638, 2021.

\bibitem[Zhan et~al.(2020)Zhan, Yu, Li, Ren, Gao, Wang, and
  Luo]{GF2Sentinel2_Turpentine_China}
Zhongyi Zhan, Linfeng Yu, Zhe Li, Lili Ren, Bingtao Gao, Lixia Wang, and
  Youqing Luo.
\newblock Combining {GF-2 and Sentinel-2} images to detect tree mortality
  caused by red turpentine beetle during the early outbreak stage in north
  {China}.
\newblock \emph{Forests}, 11\penalty0 (2), 2020.

\bibitem[Zhao et~al.(2019)Zhao, Zheng, Xu, and Wu]{DL_Detection}
Zhong-Qiu Zhao, Peng Zheng, Shou-Tao Xu, and Xindong Wu.
\newblock Object detection with deep learning: A review.
\newblock \emph{IEEE Trans. Neural Networks and Learning Systems}, 30\penalty0
  (11):\penalty0 3212--3232, 2019.

\end{thebibliography}
\end{document}